\documentclass[twoside,11pt]{article}

\usepackage{amsmath}
\usepackage{amssymb}
\usepackage{amsthm}
\usepackage{indentfirst}
\usepackage[hidelinks]{hyperref}
\usepackage{paralist}
\usepackage{geometry}
\usepackage{setspace,lipsum}
\usepackage[title]{appendix}
\usepackage{color}
\usepackage{multimedia}
\usepackage{multicol}
\usepackage{bbm}
\usepackage{graphics}
\usepackage{graphicx}
\usepackage{booktabs}
\usepackage{romannum}
\usepackage{grffile}
\usepackage{float}
\usepackage{url}
\usepackage{mathabx}
\usepackage{supertabular}
\usepackage{rotating}
\usepackage{pdflscape}
\usepackage{verbatim}
\usepackage{listings}
\usepackage{xcolor}
\usepackage{multicol}
\usepackage[shortlabels]{enumitem} 
\usepackage{caption,subcaption}
\usepackage{epstopdf}
\usepackage{algorithm}
\usepackage{algorithmic}
\usepackage[round]{natbib}
\usepackage{authblk}
\usepackage{etoolbox}
\graphicspath{{figure/}}
\newcommand{\fracpartial}[2]{\frac{\partial #1}{\partial  #2}}

\newcommand{\bm}[1]{\boldsymbol{#1}}
\newcommand{\dd}{\mathrm{d}}
\newcommand{\one}{\mathbbm{1}}
\newcommand{\E}{\mathbb{E}}
\newcommand{\p}{\mathbb{P}}

\newcommand{\g}{\mathcal{G}}
\newcommand{\f}{\mathcal{F}}

\newtheorem{assumption}{Assumption}

\newtheorem{theorem}{Theorem}
\newtheorem{lemma}{Lemma}
\newtheorem{definition}{Definition}

\begin{document}
\pagenumbering{arabic}
\providecommand{\keywords}[1]
{
\small	
\textbf{\textit{Keywords:}} #1
}
\providecommand{\jel}[1]
{
\small	
\textbf{\textit{JEL Classification:}} #1
}

\title{Policy Gradient and Actor--Critic Learning in Continuous Time and Space: Theory and Algorithms}

\author{Yanwei Jia\thanks{Department of Industrial Engineering and Operations Research, Columbia University, New York, NY 10027, USA. Email: yj2650@columbia.edu.} ~ ~ ~ Xun Yu Zhou\thanks{Department of Industrial Engineering and Operations Research \& Data Science Institute, Columbia University, New York, NY 10027, USA. Email: xz2574@columbia.edu.}}

\maketitle
\begin{abstract}
\singlespacing
We study policy gradient (PG) for reinforcement learning in continuous time and space
under the regularized exploratory formulation developed by \cite{wang2020reinforcement}. We represent the gradient of the value function with respect to a given parameterized stochastic policy as the expected integration of an auxiliary running reward function that can be evaluated using
samples and the current value function. This representation effectively turns PG into
a policy evaluation (PE) problem, enabling us to apply the martingale approach recently developed
by \cite{jia2021policy} for PE to solve our PG problem. Based on this analysis, we propose two types of actor-critic algorithms for RL, where we learn and update value functions and policies simultaneously and alternatingly. The first type is based directly on the aforementioned representation, which involves future trajectories and is offline. The second type, designed for online learning, employs the first-order condition of the policy gradient and turns
it into martingale orthogonality conditions. These conditions are then incorporated using
stochastic approximation when updating policies. Finally,
we demonstrate the algorithms by simulations in two concrete examples.
\end{abstract}


\section{Introduction}
The essence of reinforcement learning (RL) is ``trial and error": repeatedly trying a policy for actions, receiving and evaluating reward signals, and improving the policy.
This manifests three key components of RL: 1) {\bf exploration} with {\it stochastic} policies - to broaden search space via randomization; 2) {\bf policy evaluation} - to evaluate the value function of a current policy; and 3) {\bf policy improvement} - to improve the current policy. Numerous algorithms have been proposed in the RL literature, generally categorized into three types: critic-only, actor-only, and actor--critic. Here, an {\it actor} refers to a policy that governs the actions, and a {\it critic} refers to the value function that evaluates the performance of a policy. The {\it critic-only} approach learns a value function to compare the estimated outcomes of different actions and selects the best one following the current value function. 
The {\it
	actor-only}  approach acts directly without learning the expected outcomes of different policies. The {\it actor--critic} approach uses
an actor simultaneously to improve the policy for generating actions given the current state of
the environment and a critic to judge the selected policy and guide improving the actor. See \cite{sutton2011reinforcement} and the references therein for extensive discussions on these methods.

All these algorithms and indeed the general RL study have been hitherto predominantly limited to  {\it discrete-time} Markov decision processes (MDPs). From a practical point of view, however, the study on continuous-time RL with possibly continuous state and action spaces is more important. The world is inherently continuous-time, and a discrete-time dynamic is just an approximation of the reality by taking a sequence of snapshots of the world over time. As a result,  in real life,  examples abound in which an agent can or actually {\it needs} to interact with a random environment at an ultra-high frequency or outright continuously, e.g., high-frequency stock trading, autonomous driving, and robot navigation. Solving these problems in the discrete-time setting has a notorious drawback: the resulting algorithms are highly sensitive to time discretization; see \citet{tallec2019making,yildiz2021continuous} and the references therein.

Theoretically, it remains a largely uncharted territory to study RL in continuous time and spaces. The few existing papers on RL in the continuous setting are mostly restricted to {\it deterministic} systems; see, for example  \citet{baird1993advantage,doya2000reinforcement,munos2006policy,
	vamvoudakis2010online,fremaux2013reinforcement,lee2021policy,yildiz2021continuous,kim2021hamilton}  where there are no environmental noises. \cite{munos1997reinforcement} introduce RL for diffusion-based stochastic control problems without proposing a data-driven solution. Model-based methods such as those in \cite{basei2020logarithmic,szpruch2021exploration} aim to estimate model coefficients by assuming their known and simple functional forms, which are still prone to model misspecification errors. The RL research for continuous-time diffusion processes with 
data/sample-driven solutions started only recently.
\citet{wang2020reinforcement} propose an entropy-regularized stochastic relaxed control framework to study RL in continuous time and space and derive Boltzmann distributions as the generally optimal stochastic policies  for exploring the environment and generating actions. In particular, when the problem is linear--quadratic (LQ), namely, when the dynamic is linear and the reward is quadratic in state and action, the optimal policy specializes to Gaussian distributions.
Extensions and applications of this work include \citet{wang2020continuous,DDJ2020, GuoXZ2020, GXZ2020}.

While \citet{wang2020reinforcement} address the first component of RL -- exploration - for the continuous setting,  \cite{jia2021policy} investigate the second component, namely policy evaluation (PE), aiming at  establishing a theoretical foundation for PE in continuous time and space. They show that
PE is theoretically equivalent to maintaining the martingale condition of a specifically defined stochastic process, based on which they propose several online and offline PE algorithms.
These algorithms have discrete-time counterparts, such as gradient Monte Carlo, TD($\lambda$),  and GTD, that scatter around in the MDP RL literature. Therefore, through the ``martingale lens", \cite{jia2021policy} not only devise new PE algorithms for the continuous case but also interpret and unify many classical algorithms initially designed for MDPs.

The current paper is a continuation of \citet{wang2020reinforcement} and  \cite{jia2021policy}, dealing with the third component of RL -- policy improvement -- in the continuous setting under stochastic policies and, thereby, completing the whole procedure for typical RL tasks. Note that \citet{wang2020continuous} put forth a policy improvement theorem for the special case of a continuous-time mean--variance portfolio selection problem.  Furthermore, they show that defining a new policy by maximizing the Hamiltonian of the currently learned value function is proved to achieve a better objective value than the current policy. However, this method, akin to Q-learning for MDPs, has a drawback in requiring the functional form of the Hamiltonian, which in turn involves the knowledge of the environment.\footnote{In a more recent working paper \cite{jia2022q}, we develop a (little) q-learning theory to learn essentially the Hamiltonian from samples only, including a general policy improvement theorem.}  Moreover, even if the Hamiltonian is known, maximizing a potentially complex function in high dimensions is computationally demanding or daunting.

In this paper, we take a different
approach -- that of  {\it policy gradient} (PG) -- which
optimizes the value function  over a parameterized family of policies.
This approach has at least two advantages. First, selecting actions does not involve maximization, and actions are sampled from a known parametric distribution. Second, approximating policies directly facilitates more efficient learning if one has prior knowledge or intuition about the classes of potentially optimal  policies  (e.g., Gaussian distributions), leading to fewer parameters of the parametric family to be learned.  

PG as a general sub-method of RL has a long history that can be traced back to \cite{aleksandrov1968stochastic,glynn1990likelihood,williams1992simple,barto1983neuronlike};
see also \cite{bhatnagar2009natural} for more literature review and references therein. PG theorems specifically  for MDPs are established in \cite{sutton2000policy} and \cite{marbach2001simulation}. Deterministic policy gradient algorithms for semi-MDPs (with discrete time and continuous action space)  are  developed  in \cite{silver2014deterministic} and later extended to incorporate  deep neural networks in \cite{lillicrap2015continuous}. Empirically, however, such algorithms tend to be unstable \citep{duan2016benchmarking}. 
Recent studies have focused on stochastic policies with possible entropy regularizers, also known as the {\it softmax} method; see for example 
\cite{mnih2016asynchronous, schulman2017equivalence,schulman2017proximal, haarnoja2018soft}. 

PG updates and improves policies along the gradient ascent direction, and is often carried out simultaneously and alternatingly  with PE. 
The resulting algorithms for RL, therefore, are essentially actor--critic (AC) ones. Such methods have been successful in many real-world applications, notably AlphaGo \citep{silver2017mastering} and dexterous hand manipulation \citep{haarnoja2018soft}.
But then, again, most PG and AC algorithms have been developed for discrete-time MDPs, and many of them in heuristic and {\it ad hoc} manners. Existing works on PG and AC in continuous time either focus on deterministic systems \citep{fremaux2013reinforcement, kim2021hamilton} or study specific models such as linear--quadratic ones \citep{wang2021global}. There are a few papers on applications of specifically designed continuous-time PG- and/or AC-based algorithms. Toy examples include the cart-pole swing-up problem in \cite{doya2000reinforcement} and Half-Cheetah in \cite{wawrzynski2007learning}, both involving physical laws of motion. Real-world applications include portfolio selections \citep{wang2020continuous}, traffic control \citep{aragon2020traffic}, autonomous driving \citep{kiran2021deep}, and biological neural networks \citep{fremaux2013reinforcement,zambrano2015continuous}.

In sum, it remains a significant open question to develop general continuous-time PG and AC algorithms and, more importantly, to lay an overarching theoretical underpinning for them.
This paper aims to answer these questions by studying PG for a general problem in continuous time and space. Based on this, we develop model-free, data-driven AC algorithms for RL, covering both episodic and continuing, and both online and offline tasks.
As its predecessors \citet{wang2020reinforcement} and \cite{jia2021policy}, we
develop theory in continuous time and discretize time only at the final algorithmic implementation stage, instead of discretizing time upfront and applying the existing MDP results.
Specifically, we conduct our analysis in the stochastic relaxed control framework of  \citet{wang2020reinforcement} involving distribution-valued stochastic policies.
As such, it is necessary to first extend the PE theory of \cite{jia2021policy}, including the martingale characterization and the resulting methods of the martingale loss function and the martingale orthogonality conditions,  from deterministic policies to stochastic ones. This extension is technically non-trivial.
Our main contributions, however, are a thorough analysis of the PG and the resulting AC algorithms. More precisely, we deduce the representation of the gradient of the current value function with respect to a parameterized (stochastic) policy. This representation turns out to have the {\it same form} as the value function
in the PE step, effectively turning PG into an auxiliary  PE problem.
However, a subtle difficulty is that the corresponding ``auxiliary" reward depends on  the Hamiltonian and hence on the functional forms of the system dynamics. We solve this difficulty by integration by parts and It\^o's formula, transforming the representation into the  expected integration of functions that can be evaluated using samples along with the current value function approximator.

The aforementioned representation is forward-looking. Namely, it is the conditional expectation of a term involving {\it future} states. Hence it is suitable for offline learning only. For online learning, we employ  the first-order condition of the policy gradient and turn it into martingale orthogonality conditions.
These conditions are then incorporated using stochastic approximation when updating policies. Finally, combining the newly developed PG methods in this paper and the PE methods in \cite{jia2021policy}, we propose several AC algorithms for episodic and continuing/ergodic tasks. 

Within the continuous-time stochastic relaxed control framework, there are several studies involving updating policies. For example, \citet{wang2020continuous} consider mean--variance portfolio selection  and update policies by maximizing the Hamiltonian. As mentioned earlier, this requires knowledge about the market and hence the method is essentially model-based. \citet{DDJ2020} address the time-inconsistency issue and focus on learning equilibrium policies. \citet{GuoXZ2020} study multi-agent RL by solving an LQ mean-field game. These two papers rely on differentiating with respect to the policy hence are both model-based methods. In contrast, the present paper provides  general model-free (up to the underlying dynamics being diffusion processes) AC algorithms that can be applied in all the above problems. In particular, we apply our algorithms to the mean--variance portfolio selection problem in \citet{wang2020continuous} and show they outperform significantly.

The rest of the paper proceeds as follows. In Section \ref{sec:setup}, we review \citet{wang2020reinforcement}'s  entropy-regularized, exploratory  formulation for RL in continuous time and space, and put forth an equivalent formulation convenient for the subsequent analysis. In Section \ref{sec:foundation}, we develop a theory for PG, based on which we present general AC algorithms. Section \ref{sec:extension} is devoted to an extension to ergodic tasks. We demonstrate our algorithms by simulation with two concrete examples  in Section \ref{sec:applications}. Finally, Section \ref{sec:conclusion} concludes. In Appendix, we discuss the connection of our results with their discrete-time counterparts, present some theoretical results used in the simulation studies, and supply proofs of the results stated in the main text.
\section{Problem Formulation and Preliminaries}
\label{sec:setup}

Throughout this paper, by convention all vectors are {\it column} vectors unless otherwise specified, and $\mathbb{R}^k$ is the space of all $k$-dimensional  vectors (hence $k\times 1$ matrices). Let $A$ and $B$ be two matrices of the same size.
We denote by $A \circ B$ the inner product between $A$ and $B$, by $|A|$ the Eculidean/Frobenius norm of $A$, and write $A^2: = AA^\top$, where $A^\top$ is  $A$'s transpose. For a positive semidefinite matrix $A$, we write $\sqrt{A} = UD^{1/2}V^\top$, where $A = UDV^\top$ is its singular value decomposition with $U,V$ two orthogonal matrices and $D$ a diagonal matrix, and $D^{1/2}$ is the diagonal matrix whose entries are the square root of those of $D$.
We use $f=f(\cdot)$ to denote the {\it function} $f$, and $f(x)$ to denote
the {\it function value} of $f$ at $x$. 
For any stochastic process $X=\{X_{s},$ $s\geq 0\}$, we denote by $\{\f^X_s\}_{s\geq 0}$ the natural filtration generated by $X$. Finally, for any filtration $\g=\{\g_s\}_{s\geq 0}$ and any semi-martingale $Y=\{Y_{s},$ $s\geq 0\}$, we denote
\[\begin{aligned}
L^2_{\g}([0,T];Y) = \bigg\{ & \kappa = \{\kappa_t, 0\leq t\leq T \}: \\
& \mbox{ $\kappa$ is $\g_t$-progressively measurable and } \E\int_0^T |\kappa_t|^2 \dd \langle Y \rangle_t < \infty \bigg\},
\end{aligned}
\]
which is a Hilbert space with the $L^2$-norm $||\kappa||_{L^2} = \left(\E\int_0^T \kappa_t^2 \dd \langle Y \rangle_t\right)^{\frac{1}{2}}$, where  $\langle \cdot \rangle$ is the quadratic variation of a given process.

Let $d,n$ be given positive integers, $T>0$, 
and $b: [0,T]\times \mathbb{R}^d\times \mathcal{A} \mapsto \mathbb{R}^d$ and $\sigma:
[0,T]\times \mathbb{R}^d\times \mathcal{A}\mapsto \mathbb{R}^{d\times n}$ be given
functions, where $\mathcal{A}$ is the action set.
The classical stochastic control problem is to control the {\it state} (or {\it feature}) dynamics governed by  a stochastic differential equation (SDE), defined
on a filtered probability space $\left( \Omega ,\mathcal{F},\mathbb{P}^W; \{\mathcal{F}_s^W\}_{s\geq0}\right) $ along with a standard  
$n$-dimensional Brownian motion  $W=\{W_{s},$ $s\geq 0\}$:
\begin{equation}
\label{eq:model classical}
\dd X_s^{{a}} = b(s,X_s^{{a}},a_s)\dd s + \sigma(s,X_s^{{a}},a_s) \dd W_s,\
s\in [0,T],
\end{equation}
where $a_s$ stands for the agent's action (control) at time $s$. 
The goal of  stochastic control is, for each initial time-state pair $(t,x)$ of \eqref{eq:model classical}, to find the optimal $\{\mathcal{F}_s^W\}_{s\geq0}$-progressively measurable (continuous) sequence of actions $a = \{a_s,t\leq s \leq T\}$  -- also called the optimal strategy -- that maximizes the expected total reward:
\begin{equation}
\label{eq:objective}
\E^{\p^W}\left[\int_t^T e^{-\beta(s-t)} r(s,X_s^{a},a_s)\dd s + e^{-\beta (T-t)} h(X_T^{a})\Big|X_t^{a}= x\right],
\end{equation}
where $r$ is the (expected) running reward function,  $h$ is the (expected) lump-sum reward function applied at the end of the planning period $T$, and $\beta\geq 0$ is a discount factor that measures the time-value of the payoff or the impatience level of the agent.
Note in the above the state process $X^a = \{X^a_s, t\leq s \leq T\}$ also depends on $(t,x)$. However, to ease notation, here (and similarly in the sequel)  we use $X^a$ instead of $X^{t,x,a}= \{X^{t,x,a}_s, t\leq s \leq T\}$ to denote the solution to SDE \eqref{eq:model classical} with initial condition $X_t^{a} = x$ whenever no ambiguity may arise.

Let  $\mathcal{L}^a$ be the {\it infinitesimal generator} associated with the diffusion process governed by \eqref{eq:model classical}:
\[ \mathcal{L}^a \varphi (t,x): = \frac{\partial \varphi}{\partial t}(t,x) + b\big( t,x, a\big) \circ \frac{\partial \varphi}{\partial x}(t,x) + \frac{1}{2}\sigma^2\big( t,x,a \big) \circ \frac{\partial^2 \varphi}{\partial x^2}(t,x),\;\;a\in \mathcal{A}, \]
where $\frac{\partial \varphi}{\partial x} \in \mathbb{R}^d$ is the gradient, and $\frac{\partial^2 \varphi}{\partial x^2}\in \mathbb{R}^{d\times d}$ is the Hessian. We  make the following  assumption to ensure theoretically the well-posedness of the stochastic control problem \eqref{eq:model classical}--\eqref{eq:objective}.
\begin{assumption}
\label{ass:dynamic}
The following conditions for the state dynamics and reward functions hold true:

\begin{enumerate}[(i)]
	\item $b,\sigma,r,h$ are all continuous functions in their respective arguments;
	\item $b,\sigma$ are uniformly Lipschitz continuous in $x$, i.e., for $\varphi \in\{ b, \sigma\}$, there exists a constant $C>0$ such that
	\[ |\varphi(t,x,a) - \varphi(t,x',a)| \leq C|x-x'|,\;\;\forall (t,a)\in [0,T] \times \mathcal{A},\;\forall x,x'\in \mathbb{R}^d; \]
	\item $b,\sigma$ have linear growth in $x$, i.e., for $\varphi \in\{ b, \sigma\}$, there exists a constant $C>0$ such that
	\[|\varphi(t,x,a)| \leq C(1+|x|) ,\;\;\forall (t,x,a)\in [0,T] \times \mathbb{R}^d\times \mathcal{A};\]
	\item $r$ and $h$ have polynomial growth  in $(x,a)$ and $x$ respectively,  i.e., 
	there exists a constant $C>0$ and $\mu\geq 1$ such that
	\[
	|r(t,x,a)| \leq C(1+|x|^{\mu} + |a|^{\mu}) ,\;\;|h(x)| \leq C(1+|x|^{\mu}),\; \forall (t,x,a)\in [0,T] \times \mathbb{R}^d \times \mathcal{A}.\]
\end{enumerate}
\end{assumption}

Classical model-based stochastic control theory has been well developed (e.g., \citealp{fleming2006controlled} and \citealp{YZbook}) to solve the above problem, under the premise that
the functional forms of $b,\sigma,r,h$ are all given and known.
In the RL setting, however, the agent does not have this knowledge of the environment. Instead, what she can do is ``trial and error" -- to try a sequence of actions $a = \{a_s,t\leq s \leq T\}$, observe the corresponding state process $X^a = \{X_s^a,t\leq s \leq T\}$ and collect both a stream of discounted running rewards $\{e^{-\beta(s-t)} r(s,X_s^{a},a_s),t\leq s \leq T\}$ and a discounted, end-of-period lump-sum reward $e^{-\beta (T-t)}h(X_T^{a})$ where $\beta$ is a given, known discount factor. In the offline setting, the agent can repeatedly try different sequences of actions over the same time period $[0,T]$ and record the corresponding state processes and payoffs. In the online setting, the agent updates the actions as she goes, based on all the up-to-date historical observations.

A critical question is how to {\it generate} these trial-and-error sequences of actions. The idea is {\it randomization}, namely,
the agent  employs a {\it stochastic policy}, which is a probability distribution on the action space, to produce actions according to the current time--state pair. It is important to note that this randomization itself is {\it independent} of the underlying Brownian motion $W$, the random source of the original control problem that stands for the environmental noise.   \cite{wang2020reinforcement} formulate an RL problem in continuous time and space, incorporating distribution-valued stochastic policies with an entropy regularizer to account for  the tradeoff between exploration and exploitation.
Specifically,  assume the probability space is rich enough to
support a random variable $Z$ that is uniformly distributed on $[0, 1]$ and independent
of $W$. We then expand the original filtered probability space to $\left( \Omega ,\mathcal{F},\mathbb{P}; \{\mathcal{F}_s\}_{s\geq0}\right) $ where
$\mathcal{F}_s=\mathcal{F}_s^W \vee \sigma(Z)$ and $\mathbb{P}$ is now the probability measure on $\mathcal{F}_T$.\footnote{Note that a single uniform random variable $Z$ can produce many {\it independent} random variables having density functions. No dynamics are needed for these random
variables and they are all independent of each other and of the Brownian motion.
The independence means it makes no difference if these variables are given all
at once at time 0 or are revealed as time evolves. We opt for the (mathematically
speaking) easier construction where these are all defined using one
single uniform $Z$. Meanwhile, $\p$ is the product extension from $\p^W$; the two probability measures coincide when restricted to $\f^W_T$.}
Let $\bm{\pi}:(t,x)\in [0,T] \times \mathbb{R}^d \mapsto \bm{\pi}(\cdot|t,x)\in \mathcal{P}(\mathcal{A})$ be a given (feedback) policy, where $\mathcal{P}(\mathcal{A})$ is a suitable collection of probability density functions (pdfs).\footnote{Here we assume that the action space $\mathcal{A}$ is continuous and randomization is restricted to those distributions that have density functions. The analysis and results of this paper can be easily extended to the cases of discrete action spaces and/or randomization with probability mass functions.}
At each time $s$, an action $a_s$ is generated or sampled from the distribution $\bm{\pi}(\cdot|s,X_s)$. 

Given a stochastic policy $\bm{\pi}$, an initial time--state pair $(t,x)$, and an $\{\mathcal{F}_s\}_{s\geq0}$-progressively measurable action process $a^{\bm{\pi}} = \{a_s^{\bm{\pi}},t\leq s \leq T\}$  generated from $\bm{\pi}$,
the corresponding state process
$X^{\bm{\pi}} = \{X_s^{\bm{\pi}},t\leq s \leq T\}$ follows
\begin{equation}
\label{eq:model pi}
\dd X_s^{\bm{\pi}} = b(s,X_s^{\bm{\pi}},a_s^{\bm{\pi}})\dd s + \sigma(s,X_s^{\bm{\pi}},a_s^{\bm{\pi}}) \dd W_s,\
s\in [t,T]; \;\;X_{t}^{\bm{\pi}} = x
\end{equation}
defined on $\left( \Omega ,\mathcal{F},\mathbb{P}; \{\mathcal{F}_s\}_{s\geq0}\right) $. Moreover, following \cite{wang2020reinforcement}, we add a regularizer to the reward function to encourage  exploration (represented by the stochastic policy), leading to
\begin{equation}
\label{eq:objective relaxed action}
\begin{aligned}
	J(t,x;\bm{\pi}) = &	\E^{\p}\bigg[\int_t^T e^{-\beta(s-t)}\left[ r(s,X_s^{\bm{\pi}},a_s^{\bm{\pi}}) + \gamma p\big(s,X_s^{\bm{\pi}},a_s^{\bm{\pi}},\bm{\pi}(\cdot|s,X_s^{\bm{\pi}}) \big) \right]\dd s \\
	& + e^{-\beta (T-t)}h(X_T^{\bm{\pi}})\Big|X_t^{\bm{\pi}} = x \bigg],
\end{aligned}
\end{equation}
where $\E^{\p}$ is the expectation with respect to (w.r.t.) both the Brownian motion and the action randomization. In the above, $p:[0,T]\times \mathbb{R}^d \times \mathcal{A} \times \mathcal{P}(\mathcal{A}) \mapsto \mathbb{R}$ is the regularizer  and $\gamma \geq 0$ a weighting parameter on exploration, also  known as the {\it temperature} parameter. \cite{wang2020reinforcement} take the differential entropy as the regularizer, which corresponds to
\[ p\big(t,x,a,\pi(\cdot)\big) = - \log\pi(a) . \]



	Through a law of large number argument, \cite{wang2020reinforcement} show that $\{{X}_s^{\bm{\pi}}, t\leq s\leq T\}$ has the same distribution as the solution to the following SDE, denoted by $\{\tilde{X}_s^{\bm{\pi}}, t\leq s\leq T\}$:
	\begin{equation}
		\label{eq:model relaxed}
		\dd X_s = \tilde{b}\big( s,X_s,\bm{\pi}(\cdot|s, X_s) \big)\dd t + \tilde{\sigma}\big( s,X_s,\bm{\pi}(\cdot|s, X_s) \big) \dd W_s,\;s\in[t,T];\;\;\ X_{t} = x,
	\end{equation}
	where
	\[ \tilde{b}\big(s,x,\pi(\cdot)\big) = \int_{\mathcal{A}} b(s,x,a) \pi(a)\dd a,\  \tilde{\sigma}\big(s,x,\pi(\cdot)\big) = \sqrt{\int_{\mathcal{A}} \sigma^2(s,x,a) \pi(a)\dd a} .\]
	Moreover, the reward function  \eqref{eq:objective relaxed action} is identical to
	\begin{equation}
		\label{eq:objective relaxed}
		\begin{aligned}
			J(t,x;\bm{\pi}) = & \E^{\p^W}\bigg[\int_t^T e^{-\beta(s-t)}\int_{\mathcal{A}} [ r(s,\tilde{X}_s^{\bm{\pi}},a) + \gamma p\big(s,\tilde{X}_s^{\bm{\pi}},a,\bm{\pi}(\cdot|s,\tilde{X}_s^{\bm{\pi}}) \big) ]\bm{\pi}(a|s,\tilde{X}_{s}^{\bm{\pi}})\dd a \dd s\\
			& + e^{-\beta (T-t)}h(\tilde{X}_T^{\bm{\pi}})\Big|\tilde{X}_t^{\bm{\pi}} = x \bigg].
		\end{aligned}
	\end{equation}
	
	Mathematically,  \eqref{eq:model relaxed} and \eqref{eq:objective relaxed} together form a so-called \textit{relaxed stochastic control} problem where the effect of individually sampled  actions has been averaged out (over the randomization/exploration) and, hence,  one can focus on how a policy $\bm{\pi}$ impacts the {\it distribution} of the ``averaged" state $\tilde{X}$; see \cite{wang2020reinforcement}.
	
	Here, $J(t,x;\bm{\pi})$ is called the \textit{value function} of the  policy $\bm{\pi}$, and the task of RL is to find
	\begin{equation}
		\label{eq:maximize objective}
		J^*(t,x) = \max_{\bm{\pi}\in \bm{\Pi}}J(t,x;\bm{\pi}),
	\end{equation}
	where $\bm{\Pi}$ stands for the set of admissible policies. 
	The following gives the precise definition of admissible (feedback) policies.
	
	\begin{definition}
		\label{ass:admissible}
		A policy $\bm{\pi}=\bm{\pi}(\cdot|\cdot,\cdot)$ is called {\it admissible} if
		\begin{enumerate}[(i)]
			\item $\bm{\pi}(\cdot|t,x)\in \mathcal{P}(\mathcal{A})$, and $\bm{\pi}(a|t,x):(t,x,a)\in [0,T] \times \mathbb{R}^d\times \mathcal{A}\mapsto
			\mathbb{R}$ is measurable;
			\item the SDE \eqref{eq:model relaxed} admits a unique weak solution (in the sense of distribution) for any initial $(t,x)\in [0,T] \times \mathbb{R}^d$;
			\item 
			$\int_{\mathcal{A}} |r(t,x,a) + \gamma p\big(t,x,a,\bm{\pi}(\cdot|t,x) \big)| \bm{\pi}(a|t,x)\dd a \leq C(1+|x|^{\mu})$,  $\forall (t,x)$ where $C>0$ and $\mu\geq 1$ are constants;
			\item $\bm{\pi}(a|t,x)$ is continuous in $(t,x)$  and uniformly Lipschitz continuous in $x$ in the total variation distance, i.e., for each fixed $a$, $ \int_{\mathcal{A}} |\bm{\pi}(a|t,x) - \bm{\pi}(a|t',x')|\dd a \to 0$ as $(t',x')\to (t,x)$, and
			there is a constant $C>0$ independent of $(t,a)$ such that
			\[ \int_{\mathcal{A}} |\bm{\pi}(a|t,x) - \bm{\pi}(a|t,x')|\dd a \leq C|x-x'|,\;\;\forall x,x'\in \mathbb{R}^d. \]
		\end{enumerate}
	\end{definition}
	
	
	The conditions required in the above definition, while not necessarily the
	weakest ones,  are to theoretically guarantee the well-posedness of the control problem \eqref{eq:model relaxed}--\eqref{eq:objective relaxed}. This is implied by the following result.

	\begin{lemma}
		\label{lemma:relaxed sde solution}
		Let Assumptions \ref{ass:dynamic} hold and $\bm{\pi}$ be a given admissible policy. Then the SDE \eqref{eq:model relaxed} admits a unique strong solution. Moreover, for any $\mu\geq 2$, the solution  satisfies the growth condition $\E^{\p^W}\bigg[\max_{t\leq s \leq T}|\tilde{X}_s^{\bm{\pi}}|^{\mu}\Big|\tilde{X}_t^{\bm{\pi}}=x\bigg] \leq C(1 + |x|^{\mu})$ for some constant $C=C(\mu)$. Finally,  the expected payoff \eqref{eq:objective relaxed} is finite.
	\end{lemma}

	We stress that the solution to \eqref{eq:model relaxed}, $\{\tilde{X}_s^{\bm{\pi}}, t\leq s\leq T\}$, is the {\it average} of the sample trajectories over infinitely many randomized actions and is in itself {\it not} a sample trajectory  {\it nor} observable. The stochastic relaxed control problem \eqref{eq:model relaxed}-- \eqref{eq:objective relaxed}, introduced in \cite{wang2020reinforcement},  just provides a framework for {\it theoretical} analysis. In contrast, the solution to \eqref{eq:model pi}, $\{{X}_s^{\bm{\pi}}, t\leq s\leq T\}$, is a sample trajectory
	under a realization of action sequence, $\{a_s^{\bm{\pi}}, t\leq s\leq T\}$, generated from the policy $\bm{\pi}$, and can indeed be observed.
	Meanwhile, the difference between \eqref{eq:model pi} and \eqref{eq:model classical} is that   actions in the former are randomized: $a^{\bm{\pi}}$ is also driven by the randomization  and hence is {\it not} $\f_t^W$-adapted. By taking the expectation w.r.t. the action randomization, the expectation in \eqref{eq:objective relaxed action} reduces to the expectation in \eqref{eq:objective relaxed}. In other words, the problem  \eqref{eq:model relaxed}--\eqref{eq:objective relaxed} is mathematically equivalent to the problem \eqref{eq:model pi}--\eqref{eq:objective relaxed action}; yet they serve different purposes in our study: the former provides a framework for theoretical analysis of the value function while the latter  directly involves  observable samples.

	Unlike most RL problems that are formulated in an infinite planning horizon (known as {\it continuing tasks}), the current paper mainly focuses on a finite horizon setting (known as {\it episodic tasks}). Finite horizons reflect limited lifespans of real-life tasks, 
	e.g., a trader sells a financial contract with a maturity date, a robot finishes a task before a deadline, and a game player strives to pass a checkpoint given a time limit. If we let $T\to \infty$, under suitable regularity conditions (e.g., when $\beta$ is large enough) our formulation covers the discounted formulation of the continuing tasks. In addition, later we will consider an ergodic setting as an alternative formulation for continuing tasks in Section \ref{sec:extension}.


	\section{Theoretical Foundation of Actor--Critic Algorithms}
	\label{sec:foundation}
	An actor-critic (AC) algorithm consists of two parts: to estimate the value function of a given policy and to update (improve) the policy. In this section, we  provide the theoretical analysis to guide  devising such an algorithm through policy evaluation (PE) and policy gradient (PG).
	
	\subsection{Policy Evaluation}
	\label{sec:pe}
	
	\cite{jia2021policy} take a martingale perspective  to characterize PE as well as its link to solving a linear partial differential equation (PDE) numerically.
	However, they consider only {\it deterministic} policies (i.e. no randomization/exploration), without  explicitly involving  actions sampled  from a stochastic policy. The extension to the case of stochastic policies is non-trivial and specific statements of the corresponding results are important for the subsequent PG and AC algorithm design; so we present and prove them here.

	For a given stochastic policy $\bm{\pi}$, $J(\cdot,\cdot;\bm{\pi})$ can be characterized by a PDE based on the celebrated Feynman--Kac formula (cf. \citealp{karatzas2014brownian}), which also holds true for the relaxed control setting.
	\begin{lemma}
		\label{lemma:f-k pde}
		Assume there is a unique viscosity solution $v\in C\big([0,T]\times \mathbb{R}^d \big)$ to the following PDE:
		\begin{equation}
			\label{eq:pde characterization}
			\int_{\mathcal{A}} \big[ \mathcal{L}^{a} v(t,x) + r\big(t,x,a\big) + \gamma p\big(t,x,a,\bm{\pi}(\cdot|t,x) \big) - \beta v(t,x) \big] \bm{\pi}(a|t,x)\dd a = 0,\;(t,x)\in [0,T) \times \mathbb{R}^d,
		\end{equation}
		with the terminal condition $v(T,x) = h(x), \;x\in  \mathbb{R}^d $, which  satisfies $|v(t,x)|\leq C(1 + |x|^{\mu})$ for a constant $C>0$ and $\mu\geq 1$. Then $v$ is the value function, that is, $v(t,x) = J(t,x;\bm{\pi})$ for all $(t,x)\in [0,T)\times \mathbb{R}^d $.
	\end{lemma}
	
	To avoid unduly technicalities, we assume throughout this paper that the value function $J\in C^{1,2}\big([0,T)\times \mathbb{R}^d \big) \cap C\big([0,T]\times \mathbb{R}^d \big)$. There is a rich literature on conditions ensuring the unique existence and regularity of the viscosity solution to the type of equations like \eqref{eq:pde characterization}; but see \cite{tang2021exploratory} for some latest results.

	
	
	The following is the main theoretical result underpinning PE, extended from the setting of deterministic feedback policies in \cite{jia2021policy} to that of stochastic policies.
	
	\begin{theorem}
		\label{prop:pe martingale}
		A function $J(\cdot,\cdot;\bm{\pi})$ is the value function associated with the policy $\bm{\pi}$ if and only if it satisfies terminal condition $J(T,x;\bm{\pi}) = h(x)$,  and for any initial $(t,x)\in[0,T)\times \mathbb{R}^d$:
		\[ e^{-\beta s} J(s,\tilde{X}_s^{\bm{\pi}};\bm{\pi}) + \int_t^s e^{-\beta s'} \int_{\mathcal{A}} [r(s',\tilde{X}_{s'}^{\bm{\pi}},a) + \gamma p\big(s',\tilde{X}_{s'}^{\bm{\pi}},a,\bm{\pi}(\cdot|s',\tilde{X}_{s'}^{\bm{\pi}}) \big)]\bm{\pi}(a|s',\tilde{X}_{s'}^{\bm{\pi}})\dd a\dd s'  \]
		is an $(\f^{\tilde{X}^{\bm{\pi}}}, \p^W)$-martingale on $[t,T]$.
		Moreover, it is also equivalent to the martingale orthogonality condition:
		\begin{equation}
			\label{eq:martingale orthogonal}
			\E^{\p}\int_0^T \xi_t \bigg[\dd J(t,X_t^{\bm{\pi}};\bm{\pi}) + r(t,X_t^{\bm{\pi}},a_t^{\bm{\pi}})\dd t + \gamma p\big(t,X_t^{\bm{\pi}},a_t^{\bm{\pi}},\bm{\pi}(\cdot|t,X_t^{\bm{\pi}}) \big)\dd t - \beta J(t,X_t^{\bm{\pi}};\bm{\pi}) \dd t\bigg]  = 0,
		\end{equation}
		for any $\xi \in L^2_{\f^{X^{\bm{\pi}}}}\big( [0,T]; J(\cdot,X_{\cdot}^{\bm{\pi}};\bm{\pi}) \big)$. 
	\end{theorem}
	
	In the above theorem,  $\xi$ is called a {\it test function} by convention, although in general  it is actually a stochastic process.
	
	In RL, one typically employs  function approximation for learning functions of interest.
	Specifically, for PE, one uses a family of parameterized functions $J^{\theta}\equiv J^\theta(\cdot,\cdot;\bm{\pi})$ on $[0,T]\times \mathbb{R}^d$ to approximate $J$, where $\theta\in \Theta \subseteq \mathbb{R}^{L_{\theta}}$, and the problem is reduced to finding the ``best" (in some sense) $\theta$. We make the following assumption on these function approximators to be used. (Henceforth we may drop $\bm{\pi}$
	from $J^\theta(\cdot,\cdot;\bm{\pi})$ whenever no ambiguity arises.)
	\begin{assumption}
		\label{ass:value function parameterization}
		For all $\theta\in \Theta$, $J^{\theta}\in C^{1,2}\big([0,T)\times \mathbb{R}^d \big) \cap C\big([0,T]\times \mathbb{R}^d \big)$ and satisfies the polynomial growth condition in $x$. Moreover, $J^{\theta}(t,x)$ is  a smooth function in $\theta$ with  $\frac{\partial J^{\theta}}{\partial \theta}, \frac{\partial^2 J^{\theta}}{\partial \theta^2} \in C^{1,2}\big([0,T)\times \mathbb{R}^d \big) \cap C\big([0,T]\times \mathbb{R}^d \big)$ satisfying the polynomial growth condition in $x$.
	\end{assumption}
	
	Thanks to the martingale characterization in Theorem \ref{prop:pe martingale}, the PE algorithms developed  in  \cite{jia2021policy} can be adapted to the current setting in a straightforward manner. We now
	summarize them.
	\begin{enumerate}[(i)]
		\item Minimize the martingale loss function (offline):
		\[ \begin{aligned}
			\min_{\theta \in \Theta}\E^{\p}\bigg[ \int_0^T & \bigg( e^{-\beta T}h(X_T^{\bm{\pi}}) - e^{-\beta t} J^{\theta}(t,X_t^{\bm{\pi}}) \\
			& + \int_t^T e^{-\beta s} [r(s,X_s^{\bm{\pi}},a_s^{\bm{\pi}}) + \gamma p\big(s,X_s^{\bm{\pi}},a_s^{\bm{\pi}},\bm{\pi}(\cdot|s,X_s^{\bm{\pi}})\big)]\dd s  \bigg)^2 \dd t \bigg] .
		\end{aligned} \]
		This objective corresponds to the gradient Monte-Carlo algorithm  for discrete MDPs \citep{sutton2011reinforcement}.
		\item Solve the martingale orthogonality condition (online/offline):
		\[ \begin{aligned}
			\E^{\p}\bigg\{ \int_0^T \xi_t \big[ & \dd J^{\theta}(t,X_t^{\bm{\pi}}) + r(t,X_t^{\bm{\pi}},a_t^{\bm{\pi}})\dd t \\
			& + \gamma p\big(t,X_t^{\bm{\pi}},a_t^{\bm{\pi}},\bm{\pi}(\cdot|t,X_t^{\bm{\pi}}) \big)\dd t - \beta J^{\theta}(t,X_t^{\bm{\pi}}) \dd t\big] \bigg\} = 0 .
		\end{aligned}	 \]
		This objective corresponds to various (semi-gradient) TD algorithms and their variants  for  MDPs \citep{sutton1988learning,bradtke1996linear}, depending on the choices of
		the test function $\xi$.
		\item Minimize a quadratic form of the martingale orthogonality condition (online/offline):
		\[\begin{aligned}
			\min_{\theta \in \Theta}  \E^{\p}\bigg\{  \int_0^T & \xi_t \big[\dd J^{\theta}(t,X_t^{\bm{\pi}}) + r(t,X_t^{\bm{\pi}},a_t^{\bm{\pi}})\dd t + \gamma p\big(t,X_t^{\bm{\pi}},a_t^{\bm{\pi}},\bm{\pi}(\cdot|t,X_t^{\bm{\pi}}) \big)\dd t \\
			& - \beta J^{\theta}(t,X_t^{\bm{\pi}}) \dd t\big] \bigg\}^\top  A \E^{\p}\bigg\{ \int_0^T \xi_t \big[\dd J^{\theta}(t,X_t^{\bm{\pi}}) + r(t,X_t^{\bm{\pi}},a_t^{\bm{\pi}})\dd t \\
			& + \gamma p\big(t,X_t^{\bm{\pi}},a_t^{\bm{\pi}},\bm{\pi}(\cdot|t,X_t^{\bm{\pi}}) \big)\dd t - \beta J^{\theta}(t,X_t^{\bm{\pi}}) \dd t\big] \bigg\} ,
		\end{aligned}  \]
		where $A$ is a positive definite matrix of a suitable size. Typical choices are $A = I$ or $A = \big(\E^{\p}[ \int_0^T \xi_t \xi_t^\top\dd t ]\big)^{-1} $. This objective corresponds to the gradient TD algorithms and their variants  for MDPs \citep{sutton2008convergent,sutton2009fast,maei2009convergent}.
	\end{enumerate}
	
	In the above, the choice of  the parametric family $J^{\theta}$ may be guided by exploiting some special structure of the underlying problem; see \cite{wang2020continuous} for an example.  More general choices include linear combinations of some basis functions or  neural networks. On the other hand, common
	choices of the test functions are $\xi_t = \frac{\partial J^{\theta}}{\partial \theta}(t,X_t^{\bm{\pi}})$ or $\xi_t = \int_0^t \lambda^{s-t}\frac{\partial J^{\theta}}{\partial \theta}(s,X_s^{\bm{\pi}})\dd s$. Refer  to the aforementioned references for details, and in particular to \cite{jia2021policy} for the continuous setting.  Finally, when implementing these algorithms we need to discretize time, and the convergence when the mesh size goes to zero is established in \cite{jia2021policy}, which can be readily extended to the current setting.

	\subsection{Policy Gradient}
	\label{sec:policy gradient derive}
	Given an admissible policy, suppose we have carried out the PE step and obtained  an estimate of the corresponding value function. The next step is PG, namely, to estimate the gradient of the (learned) value function w.r.t. the policy.
	Specifically, let  $\bm{\pi}^{\phi}$ be a parametric family of policies with the parameter $\phi\in \Phi\subset \mathbb{R}^{L_{\phi}}$. We aim to compute the policy gradient $g(t,x;\phi): = \frac{\partial }{\partial \phi} J(t,x;\bm{\pi}^{\phi}) \in \mathbb{R}^{L_{\phi}}$ at the current time--state pair $(t,x)$. Here and throughout we always assume $\bm{\pi}^{\phi}$
	is an admissible policy.
	
	Based on the PDE characterization \eqref{eq:pde characterization} of the value function, we take the derivative in $\phi$ on both sides of \eqref{eq:pde characterization}, with $v(t,x)$ replaced by $J(t,x;\bm{\pi}^{\phi})$, to get  a new PDE satisfied by $g(t,x;\phi)$:
	\begin{equation}
		\label{eq:relaxed control policy gradient pde}
		\left\{
		\begin{aligned}
			\int_{\mathcal{A}} & \bigg\{ \big[  \mathcal{L}^a g(t,x;\phi) - \beta g(t,x;\phi) + \gamma q(t,x,a,\phi) \big]\bm{\pi}^{\phi}(a|t,x) \\
			& + \big[ \mathcal{L}^a J(t,x;\bm{\pi}^{\phi}) + r(t,x,a) + \gamma p\big( t,x,a,\bm{\pi}^{\phi}(\cdot|t,x) \big) - \beta J(t,x;\bm{\pi}^{\phi}) \big]\frac{\partial \bm{\pi}^{\phi}}{\partial \phi} (a|t,x)\bigg\} \dd a = 0,\\
			g( & T,x;\phi) = 0,
		\end{aligned}\right.
	\end{equation}
	where $q(t,x,a,\phi) = \frac{\partial }{\partial \phi}p\big( t,x,a,\bm{\pi}^{\phi}(\cdot|t,x) \big)$ that maps $[0,T]\times \mathbb{R}^d \times \mathcal{A} \times \Phi$ to $\mathbb{R}^{L_{\phi}}$. Note that \eqref{eq:relaxed control policy gradient pde} is a {\it system} of $L_{\phi}$ equations, and  $\mathcal{L}^a g$ denotes applying the operator $\mathcal{L}^a$ to each component of the $\mathbb{R}^{L_{\phi}}$-valued function $g(\cdot,\cdot;\phi)$.
	
	Define
	\[\begin{aligned}
		\check{r}(t,x,a;\phi) = & \bigg[  \mathcal{L}^a J(t,x;\bm{\pi}^{\phi}) + r(t,x,a) + \gamma p\big( t,x,a,\bm{\pi}^{\phi}(\cdot|t,x) \big) - \beta J(t,x;\bm{\pi}^{\phi}) \bigg]\frac{\frac{\partial \bm{\pi}^{\phi}}{\partial \phi} (a|t,x)}{\bm{\pi}^{\phi}(a|t,x)} \\
		& + \gamma q(t,x,a,\phi)\\
		= & \bigg[  \mathcal{L}^a J(t,x;\bm{\pi}^{\phi}) + r(t,x,a)+ \gamma p\big( t,x,a,\bm{\pi}^{\phi}(\cdot|t,x) \big)  - \beta J(t,x;\bm{\pi}^{\phi}) \bigg] \frac{\partial}{\partial \phi}\log \bm{\pi}^{\phi}(a|t,x) \\
		& + \gamma q(t,x,a,\phi),
	\end{aligned}\]
	which is again a function that maps $[0,T]\times \mathbb{R}^d \times \mathcal{A} \times \Phi$ to $\mathbb{R}^{L_{\phi}}$. Then \eqref{eq:relaxed control policy gradient pde} can be written as
	\begin{equation}
		\label{eq:relaxed control policy gradient pde 2}
		\int_{\mathcal{A}}   \big[ \mathcal{L}^a g(t,x;\phi) - \beta g(t,x;\phi) + \check{r}(t,x,a;\phi) \big]\bm{\pi}^{\phi}(a|t,x) \dd a = 0,\ g(T,x;\phi) = 0.
	\end{equation}
	Observe that \eqref{eq:relaxed control policy gradient pde 2} has the similar form to \eqref{eq:pde characterization}. Thus
	a Feynman--Kac formula (similar to Lemma \ref{lemma:f-k pde}) represents $g$ as
	\begin{equation}
		\label{eq:gradient representation expectation integrated a}
		\begin{aligned}
			g(t,x;\phi) = & \E^{\p}\left[\int_t^T e^{-\beta (s-t)} \check{r}(s,X_s^{\bm{\pi}^{\phi}},a_s^{\bm{\pi}^{\phi}};\phi)  \dd s\Big|X_t^{\bm{\pi}^{\phi}}= x\right] \\
			= & \E^{\p^W}\left[\int_t^T e^{-\beta (s-t)} \int_{\mathcal{A}} \check{r}(s,\tilde{X}_s^{\bm{\pi}^{\phi}},a;\phi)\bm{\pi}^{\phi}(a|s,\tilde{X}_s^{\bm{\pi}^{\phi}})\dd a  \dd s\Big|\tilde{X}_t^{\bm{\pi}^{\phi}}= x\right].
		\end{aligned}
	\end{equation}
	Therefore, computing PG boils down mathematically to a PE problem with a {\it different} reward function. Indeed, the task here is much easier because we only need to compute the function {\it value},
	$g(t,x;\phi)$, via \eqref{eq:gradient representation expectation integrated a} at  {\it some} $(t,x)$
	along a sample trajectory, instead of learning the {\it entire} function $g(\cdot,\cdot;\phi)$ as in PE.
	However, unlike a normal PE problem, the new reward function $\check{r}$ involves
	the operator $\mathcal{L}^a$ applied to $J$ which can not be observed nor computed without the knowledge of the environment.
	
	The remedy to overcome this difficulty rests with  It\^o's lemma and martingality. We now provide an informal argument for explanation before presenting the formal result. Suppose at time $t$, an action $a$ is generated from $\bm{\pi}^{\phi}(\cdot|t,X_t)$ and applied to the system within a small time window $[t,t+\Delta t]$. Apply It\^o's lemma to obtain
	\[ J(t+\Delta t,X_{t+\Delta t}^{a};\bm{\pi}^{\phi}) - J(t,X_{t};\bm{\pi}^{\phi}) = \int_t^{t+\Delta t} \mathcal{L}^{a} J(s,X_{s}^{a};\bm{\pi}^{\phi})\dd s + \frac{\partial J}{\partial x}(s,X_{s}^a;\bm{\pi}^{\phi})^\top\sigma_{s}\dd W_{s} .\]
	Therefore,
	\begin{equation}
		\label{eq:r check approximation}
		\begin{aligned}
			& \check{r}(s,X_s^{a},a;\phi) \dd s \\
			\equiv  & \left[ \mathcal{L}^a J(s,X_s^{a};\bm{\pi}^{\phi}) + r(s,X_s^{a},a) +  \gamma p\big( s,X_s^a,a,\bm{\pi}^{\phi}(\cdot|s,X_s^a) \big) - \beta J(s,X_s^{a};\bm{\pi}^{\phi}) \right]\\
			& \times \frac{\partial}{\partial \phi}\log \bm{\pi}^{\phi}(a|s,X_s^{a})\dd s + \gamma q( s,X_s^a,a,\phi)\dd s  \\
			\approx &  \frac{\partial}{\partial \phi}\log \bm{\pi}^{\phi}(a|s,X_s^{a})  \bigg\{ \dd J(s, X_s^{a};\bm{\pi}^{\phi})  +  \big[  r(s,X_s^{a},a)+  \gamma p\big( s,X_s^a,a,\bm{\pi}^{\phi}(\cdot|s,X_s^a) \big) \\
			& - \beta J(s,X_s^{a};\bm{\pi}^{\phi})  \big]      \dd s  - \frac{\partial J}{\partial x}(s,X_s^a;\bm{\pi})^\top\sigma_s \dd W_s \bigg\} + \gamma q( s,X_s^a,a,\phi)\dd s .
		\end{aligned}
	\end{equation}
	Since the stochastic integral w.r.t. the $\dd W$ term  above  is a martingale (under suitable regularity conditions), such a term, even if unknown, does not contribute to the expectation and thus can be ignored. As a result, $\check{r}$ can be incrementally estimated based on observations of samples and the learned value function.
	
	Before stating the main result of this paper, we
	impose the following technical conditions on the policy approximators.
	\begin{assumption}
		\label{ass:log likelihood pg}
		$\bm{\pi}^{\phi}(a|t,x)$ is smooth in $\phi\in \Phi$ for all $(t,x,a)$. Moreover,
		\[\int_{\mathcal{A}} |\check{r}(t,x,a;\phi)|\bm{\pi}^{\phi}(a|t,x)\dd a \leq C(1+|x|^{\mu})\]
		for all $(t,x,\phi)$, where $C>0,\mu\geq 1$ are constants. Furthermore, $\int_{\mathcal{A}} |\frac{\partial }{\partial \phi}\log\bm{\pi}^{\phi}(a|t,x)|^2\bm{\pi}^{\phi}(a|t,x)\dd a$ is  continuous in $(t,x)$ for all $\phi\in \Phi$.
	\end{assumption}
	
	
	\begin{theorem}
		\label{thm:pg}
		Given an admissible parameterized policy $\bm{\pi}^{\phi}$, its policy gradient $g(t,x;\phi) = \frac{\partial }{\partial \phi} J(t,x;\bm{\pi}^{\phi})$ admits the following representation:
		\begin{equation}
			\label{eq:policy gradient expectation}
			\begin{aligned}
				g(t,x;\phi) = & \E^{\p}\Bigg[\int_t^T e^{-\beta (s-t)} \bigg\{  \frac{\partial}{\partial \phi}\log \bm{\pi}^{\phi}(a_s^{\bm{\pi}^{\phi}}|s,X_s^{\bm{\pi}^{\phi}})   \bigg(  \dd J(s, X_s^{\bm{\pi}^{\phi}};\bm{\pi}^{\phi})  \\
				& +  [ r(s,X_s^{\bm{\pi}^{\phi}},a_s^{\bm{\pi}^{\phi}}) + \gamma p\big( s,X_s^{\bm{\pi}},a_s^{\bm{\pi}^{\phi}},\bm{\pi}^{\phi}(\cdot|s, X_s^{\bm{\pi}^{\phi}}) \big) - \beta J(s,X_s^{\bm{\pi}^{\phi}};\bm{\pi}^{\phi})  ]      \dd s  \bigg) \\
				&  +\gamma q(s, X_s^{\bm{\pi}^{\phi}},a_s^{\bm{\pi}^{\phi}},\phi) \dd s\bigg\} \Big| X_t^{\bm{\pi}^{\phi}} = x  \Bigg] ,\;\;(t,x)\in [0,T]\times \mathbb{R}^d.
			\end{aligned}
		\end{equation}
	\end{theorem}
	
	Once again, all the terms inside the expectation above are all computable given samples (including  action trajectories and the corresponding state trajectories) on $[t,T]$, together with an estimated value function $J$ (obtained in the previous PE step). Note that the expectation \eqref{eq:policy gradient expectation} gives the gradient of the value function w.r.t. any policy, which is not 0 in general.
	
	Observing \eqref{eq:policy gradient expectation} more closely, we can write
	$g(t,x;\phi)=g_1(t,x;\phi)+g_2(t,x;\phi)$ where
	{\small \[ \begin{array}{rl}
			g_1(t,x;\phi) = &\E^{\p}\Bigg[\int_t^T e^{-\beta (s-t)} \frac{\partial}{\partial \phi}\log \bm{\pi}^{\phi}(a_s^{\bm{\pi}^{\phi}}|s,X_s^{\bm{\pi}^{\phi}})   \bigg(  \dd J(s, X_s^{\bm{\pi}^{\phi}};\bm{\pi}^{\phi})\\
			&\;\;		 +  [ r(s,X_s^{\bm{\pi}^{\phi}},a_s^{\bm{\pi}^{\phi}}) + \gamma p\big( s,X_s^{\bm{\pi}},a_s^{\bm{\pi}^{\phi}},\bm{\pi}^{\phi}(\cdot|s, X_s^{\bm{\pi}^{\phi}}) \big) - \beta J(s,X_s^{\bm{\pi}^{\phi}};\bm{\pi}^{\phi})  ]      \dd s  \bigg)\Big| X_t^{\bm{\pi}^{\phi}} = x  \Bigg]
		\end{array}
		\]}
	and
	\[ g_2(t,x;\phi) = \E^{\p}\bigg[\int_t^T e^{-\beta (s-t)}
	\gamma q(s, X_s^{\bm{\pi}^{\phi}},a_s^{\bm{\pi}^{\phi}},\phi) \dd s\Big| X_t^{\bm{\pi}^{\phi}} = x  \bigg].
	\]
	The integrand in the expression of  $g_1$ is the discounted derivative of the log-likelihood (log-pdf) that determines the direction, multiplied by a scalar term. This scalar term is actually the TD error in the continuous setting \citep{jia2021policy} that also appears  in the martingale orthogonality condition \eqref{eq:martingale orthogonal}. Note that $g_1(t,x;\phi)\neq 0$ in general, because $\frac{\partial}{\partial \phi}\log \bm{\pi}^{\phi}(a_s^{\bm{\pi}^{\phi}}|s,X_s^{\bm{\pi}^{\phi}})$ depends on the realization of $a_s^{\bm{\pi}^{\phi}}$, and hence is not $\f_s^{X^{\bm{\pi}^{\phi}}}$-measurable and does not qualify as a test function $\xi$ in Theorem \ref{prop:pe martingale}. On the other hand, $g_2$ comes entirely from the regularizer and vanishes should the latter be absent.
	
	There are two equivalent forms  of the representation \eqref{eq:policy gradient expectation}, which can be used to  add more flexibilities in designing PG algorithms and to optimize their performance. The first one is to add a ``baseline'' action-independent function $B(t,x)$ to the integrand in \eqref{eq:policy gradient expectation}. Precisely, it follows from  $a_s^{\bm{\pi}^{\phi}}\sim\bm{\pi}^{\phi}(\cdot|s,X_s^{\bm{\pi}^{\phi}}) $  that \[\begin{aligned}
		& \E^{\p}\left[ \frac{\partial}{\partial \phi}\log \bm{\pi}^{\phi}(a_s^{\bm{\pi}^{\phi}}|s,X_s^{\bm{\pi}^{\phi}})B(s,X_s^{\bm{\pi}^{\phi}}) \Big| X_s^{\bm{\pi}^{\phi}}\right] \\
		= & B(s,X_s^{\bm{\pi}^{\phi}}) \int_{\mathcal A} \left[\frac{\partial}{\partial \phi}\log \bm{\pi}^{\phi}(a|s,X_s^{\bm{\pi}^{\phi}})\right] \bm\pi^{\phi}(a|s,X_s^{\bm{\pi}^{\phi}})\dd a \\
		= & B(s,X_s^{\bm{\pi}^{\phi}}) \int_{\mathcal A} \frac{\partial \bm{\pi}^{\phi}(a|s,X_s^{\bm{\pi}^{\phi}})}{\partial \phi}  \dd a = B(s,X_s^{\bm{\pi}^{\phi}}) \frac{\partial}{\partial \phi} \int_{\mathcal A} \bm\pi^{\phi}(a|s,X_s^{\bm{\pi}^{\phi}})  \dd a =  0.
	\end{aligned}\]
	Hence, an alternative representation of \eqref{eq:policy gradient expectation} is
	\begin{equation}
		\label{eq:policy gradient expectation baseline}
		\begin{aligned}
			& g(t,x;\phi) \\
			= & \E^{\p}\Bigg[\int_t^T e^{-\beta (s-t)} \bigg\{  \big[ \frac{\partial}{\partial \phi}\log \bm{\pi}^{\phi}(a_s^{\bm{\pi}^{\phi}}|s,X_s^{\bm{\pi}^{\phi}}) \big]  \big[  \dd J(s, X_s^{\bm{\pi}^{\phi}};\bm{\pi}^{\phi})  \\
			& +  [ r(s,X_s^{\bm{\pi}^{\phi}},a_s^{\bm{\pi}^{\phi}}) + \gamma p\big( s,X_s^{\bm{\pi}},a_s^{\bm{\pi}^{\phi}},\bm{\pi}^{\phi}(\cdot|s, X_s^{\bm{\pi}^{\phi}}) \big) - \beta J(s,X_s^{\bm{\pi}^{\phi}};\bm{\pi}^{\phi}) - B(s,X_s^{\bm{\pi}^{\phi}})  ]      \dd s  \big] \\
			&  +\gamma q(s, X_s^{\bm{\pi}^{\phi}},a_s^{\bm{\pi}^{\phi}},\phi) \dd s\bigg\} \Big| X_t^{\bm{\pi}^{\phi}} = x  \Bigg] .
		\end{aligned}
	\end{equation}
	Including such a baseline function in the representation of PG goes back at least to \cite{williams1992simple}. \cite{sutton2000comparing} and \cite{zhao2011analysis} find that adding an appropriate baseline function  can reduce the variance of the learning process.
	In particular, a common choice of baseline function, though not theoretically optimal, is the current value function, which leads to the so-called advantage AC algorithms \citep{degris2012model,mnih2016asynchronous}. Interestingly,
	without including any exogenous baseline function,  the PG algorithms out of \eqref{eq:policy gradient expectation} are exactly the continuous-time versions of the advantage AC algorithms. As such, we do not add other baseline functions for designing our  algorithms below. More connections to the representation of policy gradient in discrete-time and detailed discussions of the baseline function can be found in Appendix A.
	%
	%
	
	The second alternative form  of \eqref{eq:policy gradient expectation} is to add
	an admissible test function to the derivative of the log-likelihood. Specifically, suppose $ \zeta \in L^2_{\f^{X^{\bm{\pi}^{\phi}}}}\big( [0,T]; J(\cdot,X_{\cdot}^{\bm{\pi}^{\phi}};\bm{\pi}^{\phi}) \big)$
	is an $\mathbb{R}^{L_{\phi}}$-valued process. Then based on Theorem \ref{prop:pe martingale}, the policy gradient can also be represented by
	\begin{equation}
		\label{eq:policy gradient expectation baseline in loglikelihood}
		\begin{aligned}
			g(t,x;\phi) = & \E^{\p}\Bigg[\int_t^T e^{-\beta (s-t)} \bigg\{ \big[ \frac{\partial}{\partial \phi}\log \bm{\pi}^{\phi}(a_s^{\bm{\pi}^{\phi}}|s,X_s^{\bm{\pi}^{\phi}})  + \zeta_s \big] \bigg(  \dd J(s, X_s^{\bm{\pi}^{\phi}};\bm{\pi}^{\phi})  \\
			& +  [ r(s,X_s^{\bm{\pi}^{\phi}},a_s^{\bm{\pi}^{\phi}}) + \gamma p\big( s,X_s^{\bm{\pi}},a_s^{\bm{\pi}^{\phi}},\bm{\pi}^{\phi}(\cdot|s, X_s^{\bm{\pi}^{\phi}}) \big) - \beta J(s,X_s^{\bm{\pi}^{\phi}};\bm{\pi}^{\phi})  ]      \dd s  \bigg) \\
			&  +\gamma q(s, X_s^{\bm{\pi}^{\phi}},a_s^{\bm{\pi}^{\phi}},\phi) \dd s\bigg\} \Big| X_t^{\bm{\pi}^{\phi}} = x  \Bigg] ,\;\;(t,x)\in [0,T]\times \mathbb{R}^d.
		\end{aligned}
	\end{equation}

	As discussed before, we do not use \eqref{eq:policy gradient expectation} to approximate the function
	$g(\cdot,\cdot;\phi)$. Rather, at any current time--state $(t,x)$, \eqref{eq:policy gradient expectation} gives the gradient of $J(t,x;\phi)$ in $\phi$ so that we can update $\phi$
	in the most promising  direction  (based on the gradient ascent algorithm) to improve the value of $J$. However, the right hand side of \eqref{eq:policy gradient expectation} involves only the {\it future} trajectories from $t$; so Theorem \ref{thm:pg} works only for the offline setting.
	
	\smallskip
	
	To treat the online case, assume that $\phi^*$ is the optimal point of $J(t,x;\bm{\pi}^{\phi})$
	for any $(t,x)$ and  that the first-order condition holds (e.g., when $\phi^*$ is an interior point).\footnote{A theoretically optimal
		policy $\bm{\pi}^{*}$ indeed maximizes $J(t,x;\bm{\pi})$ for {\it any} $(t,x)$, based on the verification theorem; see \cite{YZbook}.} Then $g(t,x;\phi^*) = 0$. It thus follows from \eqref{eq:relaxed control policy gradient pde} that
	\begin{equation}
		\label{eq:optimal value function foc}
		\begin{aligned}
			0 = &\int_{\mathcal{A}} \bigg\{ \big[ \mathcal{L}^a J(t,x;\bm{\pi}^{\phi^*}) + r(t,x,a)+ \gamma p\big( t,x,a,\bm{\pi}^{\phi^*}(\cdot|t,x) \big) - \beta J(t,x;\bm{\pi}^{\phi^*}) \big]\frac{\partial \bm{\pi}^{\phi^*}}{\partial \phi} (a|t,x)\\
			&  +  \gamma q( t,x,a,\phi^*)\bm{\pi}^{\phi^*}(a|t,x)  \bigg\}\dd a \\
			= & \int_{\mathcal{A}} \check{r}(t,x,a;\phi^*)  \bm{\pi}^{\phi^*}(a|t,x) \dd a.
		\end{aligned}
	\end{equation}
	This is the same type of equation as \eqref{eq:pde characterization} involved in the Feynman--Kac formula. In the same way as \eqref{eq:pde characterization} leading to Theorem \ref{prop:pe martingale}, we can prove the following conclusion.
	\begin{theorem}
		\label{prop:pg optimal policy condition}	
		If there exists an interior optimal point $\phi^*$ that maximizes $J(0,x;\bm{\pi}^{\phi})$ for any $x\in\mathbb{R}^d$, then
		\begin{equation}
			\label{eq:optimal policy condition}
			\begin{aligned}
				0 = & \E^{\p}\Bigg[\int_0^T \eta_s \bigg\{  \big[ \frac{\partial}{\partial \phi}\log \bm{\pi}^{\phi^*}(a_s^{\bm{\pi}^{\phi^*}}|s,X_s^{\bm{\pi}^{\phi^*}}) + \zeta_s \big]  \big[  \dd J(s, X_s^{\bm{\pi}};\bm{\pi}^{\phi^*})  \\
				& +  [ r(s,X_s^{\bm{\pi}^{\phi^*}},a_s^{\bm{\pi}^{\phi^*}}) + \gamma p\big( s,X_s^{\bm{\pi}^{\phi^*}},a_s^{\bm{\pi}^{\phi^*}},\bm{\pi}^{\phi^*}(\cdot|s, X_s^{\bm{\pi}^{\phi^*}}) \big) - \beta J(s,X_s^{\bm{\pi}^{\phi^*}};\bm{\pi}^{\phi^*})  ]      \dd s  \big] \\
				&  +\gamma q(s, X_s^{\bm{\pi}^{\phi^*}},a_s^{\bm{\pi}^{\phi^*}},\phi^*) \dd s\bigg\} \Big| X_0^{\bm{\pi}^{\phi^*}} = x  \Bigg]
			\end{aligned}
		\end{equation}
		for any $\eta,\zeta\in L^2_{\f^{X^{\bm{\pi}^{\phi^*}}}}\big( [0,T];  J(\cdot,X^{\bm{\pi}^{\phi^*}}_{\cdot};\bm{\pi}^{\phi^*} ) \big)$.
	\end{theorem}
	
	If we take  $\eta_s = e^{-\beta s}$, then the right hand side of \eqref{eq:optimal policy condition} coincides with $g(0,x,\phi^*)$. However, though only a necessary condition, \eqref{eq:optimal policy condition} contains infinitely many equations with different test functions $\eta$. More importantly, besides the flexibility of choosing  different sets of  test functions, \eqref{eq:optimal policy condition} provides a way to derive a system of  equations based on only {\it past} observations and, hence, enables online learning. For example, by taking $\eta_s = 0$ on $[t,T]$, \eqref{eq:optimal policy condition} involves sample trajectories up to only the present time $t$. Thus, learning the optimal policy either offline or online boils down  to solving a system of equations (with suitably chosen test functions) via stochastic approximation to find $\phi^*$.
	
	In sum, Theorems \ref{thm:pg} and \ref{prop:pg optimal policy condition} foreshadow two different types of algorithms which we will develop in the next subsection.
	
	\subsection{Actor--Critic Algorithms}
	We now design  actor--critic (AC) algorithms  by combining the PE and the PG steps. For the former, \cite{jia2021policy} develop two methods, those of martingale loss function and martingale orthogonality conditions, to devise several online/offline PE algorithms  for the continuous setting. As discussed in Subsection \ref{sec:pe}, one can adopt any of these algorithms that is suitable for the given  learning context and computational resource to estimate the value function of any given policy. Here we focus on how to update the policy based on  our previous theoretical analysis on PG.
	
	First, in the offline setting where full state trajectories under any given policy can be repeatedly sampled and observed, the gradient of the value function w.r.t. the policy is given by \eqref{eq:policy gradient expectation}, which can be estimated using future samples from any {\it current} time--state $(t,x)$. That is, $g(t,x;\phi)$ is the gradient direction that would maximally improve the total reward at $(t,x)$. 
	
	For online learning, as explained earlier,  \eqref{eq:policy gradient expectation} is no longer implementable.
	Instead of computing gradients, we turn to \eqref{eq:optimal policy condition} for directly solving the optimal policy. Specifically, at any current time $t$, we  choose $\eta_s = 0$ for $s\in [t,T]$ so that the integral in \eqref{eq:optimal policy condition} only utilizes past observations up to $t$, and hence is computable. Therefore, in the online setting one applies stochastic approximation to solve the optimal condition \eqref{eq:optimal policy condition} in order to search for the optimal policy $\phi^*$.
	
	Recall that $J^{\theta}\equiv J^{\theta}(\cdot,\cdot)$, where $J^{\theta}(t,x)\in \mathbb{R}$, is a family of scalar functions on $(t,x)\in [0,T]\times \mathbb{R}^d$ parameterized by  $\theta\in \Theta \subseteq \mathbb{R}^{L_{\theta}}$, and $\bm{\pi}^{\phi}\equiv \bm{\pi}^{\phi}(\cdot|\cdot,\cdot)$, where  $\bm{\pi}^{\phi}(\cdot|t,x) \in \mathcal{P}(\mathcal{A})$, is a family of pdf-valued policy functions on $(t,x)\in [0,T]\times \mathbb{R}^d$ parameterized by $\phi\in \Phi \subseteq \mathbb{R}^{L_{\phi}}$. The aim of an AC algorithm is to find the optimal
	$(\theta,\phi)$ jointly, by updating the two parameters  alternatingly. Note that, although our problem is continuous in time, the final algorithmic implementation requires discretizing time. For simplicity, we use equally spaced mesh grid $t_k = k \Delta t$, with $k =0,\cdots, K = \lfloor{\frac{T}{\Delta t }}\rfloor$.
	
	We now present the following pseudo codes in Algorithms \ref{algo:offline episodic} and \ref{algo:online incremental}. Algorithm \ref{algo:offline episodic} is for offline-episodic learning, where  full trajectories are sampled  and observed repeatedly during different episodes and $(\theta,\phi)$ are updated after one whole episode. Algorithm \ref{algo:online incremental}  is for online incremental learning, where only the past sample  trajectory is available and $(\theta,\phi)$ are updated in real-time incrementally.
	
	\begin{algorithm}[htbp]
		\caption{Offline--Episodic Actor--Critic Algorithm}
		\textbf{Inputs}: initial state $x_0$,  horizon $T$, time step $\Delta t$, number of episodes $N$, number of mesh grids $K$, initial learning rates $\alpha_{\theta},\alpha_{\phi}$ and a learning rate schedule function $l(\cdot)$ (a function of the number of episodes), functional form of the value function $J^{\theta}(\cdot,\cdot)$, functional form of the policy $\bm{\pi}^{\phi}(\cdot|\cdot,\cdot)$, functional form of the regularizer $p\big(t,x,a,\pi(\cdot) \big)$, functional forms of the test functions $\bm{\xi}(t,x_{\cdot \wedge t})$, $\bm{\zeta}(t,x_{\cdot \wedge t})$, and temperature parameter $\gamma$.

		\textbf{Required program}: an environment simulator $(x',r) = \textit{Environment}_{\Delta t}(t,x,a)$ that takes current time-state pair $(t,x)$ and action $a$ as inputs and generates state $x'$ at time $t+\Delta t$ and the instantaneous reward $r$ at time $t$.

		\textbf{Learning procedure}:
		\begin{algorithmic}
			\STATE Initialize $\theta,\phi$.
			\FOR{episode $j=1$ \TO $N$} \STATE{Initialize $k = 0$. Observe the initial state $x_0$ and store $x_{t_k} \leftarrow  x_0$.
				\WHILE{$k < K$} \STATE{
					Compute and store the test function $\xi_{t_k} = \bm{\xi}(t_k, x_{t_0},\cdots, x_{t_k})$, $\zeta_{t_k} = \bm{\zeta}(t_k, x_{t_0},\cdots, x_{t_k})$.	
					
					Generate action $a_{t_k}\sim \bm{\pi}^{\phi}(\cdot|t_k,x_{t_k})$.
					
					Apply $a_{t_k}$ to the environment simulator $(x,r) = Environment_{\Delta t}(t_k, x_{t_k}, a_{t_k})$, and observe the output new state $x$ and reward $r$. Store $x_{t_{k+1}} \leftarrow x$ and $r_{t_k} \leftarrow r$.
					
					Update $k \leftarrow k + 1$.
				}
				\ENDWHILE	
				
				Compute
				\[\begin{aligned}
					\Delta \theta = \sum_{i=0}^{K-1} \xi_{t_i} \big[ & J^{\theta}(t_{i+1},x_{t_{i+1}}) - J^{\theta}(t_{i},x_{t_{i}}) + r_{t_i}\Delta t  \\
					&+ \gamma p\big(t_i,x_{t_i},a_{t_i},\bm{\pi}^{\phi}(\cdot|t_i,x_{t_i}) \big)\Delta t - \beta J^{\theta}(t_{i},x_{t_{i}}) \Delta t  \big],
				\end{aligned}  \]
				\[\begin{aligned}
					\Delta \phi =  \sum_{i=0}^{K-1}e^{-\beta t_i} \bigg\{  & \big[ \frac{\partial}{\partial \phi}\log\bm{\pi}^{\phi}(a_{t_i}|t_i,x_{t_i}) + \zeta_{t_i}\big]\big[ J^{\theta}(t_{i+1},x_{t_{i+1}}) - J^{\theta}(t_{i},x_{t_{i}}) + r_{t_i}\Delta t\\
					& + \gamma p\big(t_i,x_{t_i},a_{t_i},\bm{\pi}^{\phi}(\cdot|t_i,x_{t_i}) \big)\Delta t - \beta J^{\theta}(t_{i},x_{t_{i}}) \Delta t  \big] \\
					& + \gamma \frac{\partial p}{\partial \phi}\big(t_i,x_{t_i},a_{t_i},\bm{\pi}^{\phi}(\cdot|t_i,x_{t_i}) \big)\Delta t   \bigg\}.
				\end{aligned}  \]
				
				Update $\theta$ (policy evaluation) by
				\[ \theta \leftarrow \theta + l(j)\alpha_{\theta} \Delta \theta .\]
				
				Update $\phi$ (policy gradient) by
				\[ \phi \leftarrow \phi + l(j)\alpha_{\phi} \Delta \phi .  \]
				
			}
			\ENDFOR
		\end{algorithmic}
		\label{algo:offline episodic}
	\end{algorithm}

	\begin{algorithm}[htbp]
		\caption{Online-Incremental Actor--Critic Algorithm}
		\textbf{Inputs}: initial state $x_0$, horizon $T$, time step $\Delta t$, number of mesh grids $K$, initial learning rates $\alpha_{\theta},\alpha_{\phi}$ and learning rate schedule function $l(\cdot)$ (a function of the number of episodes), functional form of the value function $J^{\theta}(\cdot,\cdot)$, functional form of the policy $\bm{\pi}^{\phi}(\cdot|\cdot,\cdot)$, functional form of  the regularizer $p\big(t,x,a,\pi(\cdot) \big)$, functional forms of the test functions $\bm{\xi}(t,x_{\cdot \wedge t}),\bm{\eta}(t,x_{\cdot \wedge t})$, $\bm{\zeta}(t,x_{\cdot \wedge t})$, and temperature parameter $\gamma$.
		
		\textbf{Required program}: an environment simulator $(x',r) = \textit{Environment}_{\Delta t}(t,x,a)$ that takes current time-state pair $(t,x)$ and action $a$ as inputs and generates state $x'$ at time $t+\Delta t$ and the instantaneous reward $r$ at time $t$.

		\textbf{Learning procedure}:
		\begin{algorithmic}
			\STATE Initialize $\theta,\phi$.
			\FOR{episode $j=1$ \TO $\infty$} \STATE{Initialize $k = 0$. Observe the initial state $x_0$ and store $x_{t_k} \leftarrow  x_0$.
				\WHILE{$k < K$} \STATE{
					Compute test function $\xi_{t_k} = \bm{\xi}(t_k,x_{t_0},\cdots, x_{t_k})$, $\eta_{t_k} = \bm{\eta}(t_k,x_{t_0},\cdots, x_{t_k})$, and $\zeta_{t_k} = \bm{\zeta}(t_k,x_{t_0},\cdots, x_{t_k})$.	
					
					Generate action $a_{t_k}\sim \bm{\pi}^{\phi}(\cdot|t_k,x_{t_k})$.
					
					Apply $a_{t_k}$ to the environment simulator $(x,r) = Environment_{\Delta t}(t_k, x_{t_k}, a_{t_k})$, and observe the output new state $x$ and reward $r$. Store $x_{t_{k+1}} \leftarrow x$ and $r_{t_k} \leftarrow r$.
					
					Compute
					\[ \begin{aligned}
						& \delta = J^{\theta}(t_{k+1},x_{t_{k+1}}) - J^{\theta}(t_{k},x_{t_{k}}) + r_{t_k}\Delta t \\
						& + \gamma p\big(t_k,x_{t_k},a_{t_k},\bm{\pi}^{\phi}(\cdot|t_k,x_{t_k}) \big)\Delta t - \beta J^{\theta}(t_{k},x_{t_{k}}) \Delta t ,\\
						& \Delta \theta = \xi_{t_k} \delta, \\
						& \Delta \phi = \eta_{t_k} \bigg\{ \big[ \frac{\partial}{\partial \phi}\log\bm{\pi}^{\phi}(a_{t_k}|t_k,x_{t_k}) + \zeta_{t_k}\big]\delta + \gamma \frac{\partial p}{\partial \phi}\big(t_k,x_{t_k},a_{t_k},\bm{\pi}^{\phi}(\cdot|t_k,x_{t_k}) \big) \Delta t \bigg\}.
					\end{aligned} \]

					Update $\theta$ (policy evaluation) by
					\[ \theta \leftarrow \theta + l(j)\alpha_{\theta} \Delta \theta .\]
					
					Update $\phi$ (policy gradient) by
					\[ \phi \leftarrow \phi + l(j)\alpha_{\phi} \Delta \phi .  \]
					
					Update $k \leftarrow k + 1$
				}
				\ENDWHILE	
				
			}
			\ENDFOR
		\end{algorithmic}
		\label{algo:online incremental}
	\end{algorithm}
	
	Note that Algorithms \ref{algo:offline episodic} and \ref{algo:online incremental} presented here are just for illustrative  purpose; there is ample flexibility to devise their variants depending on the specific problems concerned. In particular, the choice of test functions dictates in which sense we approximate the value function and policy.\footnote{See \cite{jia2021policy} for detailed discussions on this point  for the PE part. Also, to save computational and memory cost of  algorithms, we usually choose test functions that can be computed incrementally. For example, in a TD($\lambda$) algorithm, $\xi_{t_k} = \int_0^{t_k} \lambda^{t_k-s}\frac{\partial J^{\theta}}{\partial \theta}(s,X_s)\dd s \approx \lambda^{\Delta t}\xi_{t_{k-1}} + \frac{\partial J^{\theta}}{\partial \theta}(t_k,X_{t_k})\Delta t$, and $\zeta_{t_k} \approx \lambda^{\Delta t}\zeta_{t_{k-1}} + \frac{\partial }{\partial \phi}\log\bm{\pi}^{\phi}(a^{\bm{\pi}^{\phi}}_{t_{k-1}}|t_{k-1},X_{t_{k-1}})  \Delta t$, which can be calculated recursively.}
	For example, if  we take the test functions $\xi_t = \frac{\partial J^{\theta}}{\partial \theta}(t,X_t)$, and $\eta_t = e^{-\beta t} $, then we have  essentially TD(0) AC algorithms. If we take $\xi_t = \int_0^t \lambda^{t-s}\frac{\partial J^{\theta}}{\partial \theta}(s,X_s)\dd s$, $\zeta_t = \int_0^{t-\Delta t} \lambda^{t-s}\frac{\partial }{\partial \phi}\log\bm{\pi}^{\phi}(a_s^{\bm{\pi}^{\phi}}|s,X_s)  \dd s $, then we end up with TD($\lambda$) algorithms \citep{sutton2011reinforcement}.
	Moreover, in the PE part of the algorithms
	we can also use other methods (online or offline) as summarized in Subsection \ref{sec:pe}.

	Finally, we reiterate that the main purpose of this paper is to provide a theoretical foundation to guide designing AC algorithms, instead of comparing which algorithm performs better. As such,  we only present the TD-type algorithms for illustration, acknowledging that there are multiple ways to combine PE and the newly developed PG methods  to design new learning algorithms.
	

	\section{Extension to Ergodic Tasks}
	\label{sec:extension}
	
	In this section we extend our results and algorithms to ergodic (long-term average) tasks, which are also  commonly studied in the RL literature. The ergodic objective is one possible formulation of continuing tasks, in which a learning algorithm is based on only one single trajectory.

		Consider a regularized  ergodic objective function
		\[\begin{aligned}
			& \liminf_{T\to \infty}\frac{1}{T}\E^{\p^W}\bigg[ \int_t^T \int_{\mathcal{A}} [ r(\tilde{X}_s^{\bm{\pi}},a) + \gamma p\big(\tilde{X}_s^{\bm{\pi}},a,\bm{\pi}(\cdot|\tilde{X}_s^{\bm{\pi}}) \big) ]\bm{\pi}(a|\tilde{X}_{s}^{\bm{\pi}})\dd a  \dd s \Big| \tilde{X}_t^{\bm{\pi}} = x\bigg] \\
			= & \liminf_{T\to \infty}\frac{1}{T}\E^{\p}\bigg[ \int_t^T [ r(X_s^{\bm{\pi}},a_s^{\bm{\pi}}) + \gamma p\big(X_s^{\bm{\pi}},a_s^{\bm{\pi}},\bm{\pi}(\cdot|X_s^{\bm{\pi}}) \big) ] \dd s \Big| X_t^{\bm{\pi}} = x\bigg],
		\end{aligned}  \]
		where $p$ is the regularizer and $\gamma \geq 0$ is the temperature parameter. Note that now the running reward, the regularizer and the policy do not depended on  time explicitly due to the stationary nature of ergodic tasks.
		
		One way to study an ergodic task is to connect it to a discounted, infinite horizon problem:
		\[\begin{aligned}
			& \E^{\p^W}\bigg[ \int_t^{\infty}e^{-\beta(s- t)} \int_{\mathcal{A}} [ r(\tilde{X}_s^{\bm{\pi}},a) + \gamma p\big(\tilde{X}_s^{\bm{\pi}},a,\bm{\pi}(\cdot|\tilde{X}_s^{\bm{\pi}}) \big) ]\bm{\pi}(a|\tilde{X}_{s}^{\bm{\pi}})\dd a  \dd s \Big| \tilde{X}_t^{\bm{\pi}} = x\bigg]  \\
			= &  \E^{\p}\bigg[ \int_t^{\infty}e^{-\beta(s- t)} [ r(X_s^{\bm{\pi}},a_s^{\bm{\pi}}) + \gamma p\big(X_s^{\bm{\pi}},a_s^{\bm{\pi}},\bm{\pi}(\cdot|X_s^{a}) \big) ]  \dd s \Big| X_t^{\bm{\pi}} = x\bigg] .
		\end{aligned}\]
		It has been shown that, under suitable conditions, the optimal value function of the discounted infinite horizon problem converges to the optimal ergodic reward as the discount factor $\beta \to 0$; see,  e.g., \cite{borkar1988ergodic,borkar1990ergodic, bensoussan1992bellman}.
		
		Here, we opt for a direct treatment of ergodic problems.
		According to \citet[page 249]{sutton2011reinforcement},
		ergodic tasks are actually better behaved than continuing tasks with discounting. For a systematic account of  classical ergodic control theory in continuous time, see \citet{arapostathis2012ergodic} and the references therein.

		We first present  the ergodic version of the Feynman--Kac formula.
		\begin{lemma}
			\label{lemma:ergodic feynmann-kac}
			Let $\bm{\pi}=\bm{\pi}(\cdot|\cdot)$ be a given (time-invariant) policy.
			Suppose there is a function $J(\cdot;\bm{\pi})\in C^2(\mathbb{R}^d)$ and a scalar $V(\bm{\pi}) \in \mathbb{R}$ satisfying
			\begin{equation}
				\label{eq:relaxed control f-k formula ergodic}
				\int_{\mathcal{A}} \big[ \mathcal{L}^a J(x;\bm{\pi}) + r(x,a) +\gamma p\big(x,a,\bm{\pi}(\cdot|x)\big) \big]\bm{\pi}(a|x) \dd a - V(\bm{\pi})= 0,\;\;x\in \mathbb{R}^d.
			\end{equation}
			Then for any $t\geq0$, 
			\begin{equation}
				\label{con1}
				\begin{array}{rl}
					V(\bm{\pi}) = & \liminf_{T\to \infty}\frac{1}{T}\E^{\p^W}\bigg[ \int_t^T \int_{\mathcal{A}} [ r(\tilde{X}_s^{\bm{\pi}},a) + \gamma p\big(\tilde{X}_s^{\bm{\pi}},a,\bm{\pi}(\cdot|\tilde{X}_s^{\bm{\pi}}) \big) ]\bm{\pi}(a|\tilde{X}_{s}^{\bm{\pi}})\dd a  \dd s \Big| \tilde{X}_t^{\bm{\pi}} = x\bigg] \\
					= & \liminf_{T\to \infty}\frac{1}{T}\E^{\p}\bigg[ \int_t^T [ r(X_s^{\bm{\pi}},a_s^{\bm{\pi}}) + \gamma p\big(X_s^{\bm{\pi}},a_s^{\bm{\pi}},\bm{\pi}(\cdot|X_s^{\bm{\pi}}) \big) ] \dd s \Big| X_t^{\bm{\pi}} = x\bigg].
				\end{array}
			\end{equation}
			Moreover, $J(X_t^{\bm{\pi}};\bm{\pi}) + \int_0^t [ r(X_s^{\bm{\pi}}, a_s^{\bm{\pi}}) + \gamma p\big(X_s^{\bm{\pi}}, a_s^{\bm{\pi}}, \bm{\pi}(\cdot|X_s^{\bm{\pi}}) \big) - V(\bm{\pi}) ]  \dd s$
			is an $(\f^{X^{\bm{\pi}}},\p)$-martingale.
			
		\end{lemma}

		We emphasize that the solution to \eqref{eq:relaxed control f-k formula ergodic} is a {\it pair} of $(J, V)$, where $J(\cdot;\bm\pi)$ is a function of the state and $V(\bm\pi)\in \mathbb{R}$ is a scalar. The long term average of the payoff does not depend on the initial state $x$ nor the initial time $t$ due to the ergodicity, and hence remains a constant as \eqref{con1} implies. The function $J$, on the other hand, only represents the first-order approximation of long-run average and is not unique. Indeed, for any constant $c$, $(J+c, V)$ is also a solution to \eqref{eq:relaxed control f-k formula ergodic}. We  refer to $V$ as the ``value''. Lastly, since the value  does not depend on the  initial time,  we will fix the latter  as 0 in the following discussions and applications of ergodic tasks.
		
		For a given policy $\bm{\pi}$, the PE problem is now to find a function $J(\cdot;\bm{\pi})$ and a value $V \in \mathbb{R}$, such that
		\[ J(X_t^{\bm{\pi}};\bm{\pi}) + \int_0^t [ r(X_s^{\bm{\pi}}, a_s^{\bm{\pi}}) + \gamma p\big(X_s^{\bm{\pi}}, a_s^{\bm{\pi}}, \bm{\pi}(\cdot|X_s^{\bm{\pi}}) \big) - V(\bm{\pi}) ]  \dd s
		\]
		is a martingale. Following \cite{jia2021policy}, we can then design online PE algorithms based on the following martingale orthogonality conditions:
		\begin{equation}
			\label{eq:pe martingale ergodic}
			\E^{\p} \int_0^{T} \xi_t \Big\{ \dd J(X_t^{\bm{\pi}};\bm{\pi}) + \big[r(X_t^{\bm{\pi}}, a_t^{\bm{\pi}}) + \gamma p\big(X_t^{\bm{\pi}}, a_t^{\bm{\pi}}, \bm{\pi}(\cdot|X_t^{\bm{\pi}}) \big) - V \big]\dd t\Big\} 
			=0,
		\end{equation}
		for any $T>0$, any initial state $x$, and any test function $\xi\in L^2_{\f^{X^{\bm{\pi}}}}\big([0,T];  J(X_{\cdot}^{\bm{\pi}};\bm{\pi}) \big)$.

		We now focus on PG. Suppose we parameterize the policy by $\bm{\pi}^{\phi}$, we aim to  estimate $\frac{\partial V(\bm{\pi}^{\phi})}{\partial \phi}$.
		Taking the derivative in $\phi$ in \eqref{eq:relaxed control f-k formula ergodic}, we obtain
		\[\begin{aligned}
			\frac{\partial V(\bm{\pi}^{\phi})}{\partial \phi} = & \int_{\mathcal{A}}  \big[ \mathcal{L}^a J(x;\bm{\pi}^{\phi}) + r(x,a) +\gamma p\big(x,a,\bm{\pi}^{\phi}(\cdot|x)\big)  \big] \frac{\partial \bm{\pi}^{\phi}(a|x)}{\partial \phi} \dd a \\
			& +  \int_{\mathcal{A}}  \mathcal{L}^a \frac{\partial J(x;\bm{\pi}^{\phi})}{\partial \phi} \bm{\pi}^{\phi}(a|x) \dd a  + \gamma\int_{\mathcal{A}}  \frac{\partial p\big(x,a,\bm{\pi}^{\phi}(\cdot|x)\big)}{\partial \phi} \bm{\pi}^{\phi}(a|x)\dd a.
		\end{aligned}  \]
		Denote $q(x,a,\phi): =  \frac{\partial }{\partial \phi}p\big(x,a,\bm{\pi}^{\phi}(\cdot|x)\big)$,\[\check{r}(x,a;\phi): =  \big[ \mathcal{L}^a J(x;\bm{\pi}^{\phi}) + r(x,a) +\gamma p\big(x,a,\bm{\pi}(\cdot|x)\big)   \big] \frac{\frac{\partial \bm{\pi}^{\phi}(a|x)}{\partial \phi}}{\bm{\pi}^{\phi}(a|x)}  + \gamma q(x,a,\phi),\] and $g(x;\phi): = \frac{\partial }{\partial \phi}J(x;\bm{\pi}^{\phi})$. Then
		\[ \int_{\mathcal{A}}  [\mathcal{L}^a g(x;\phi) + \check{r}(x,a;\phi)] \bm{\pi}^{\phi}(a|x) \dd a -\frac{\partial V(\bm{\pi}^{\phi})}{\partial \phi} =0.  \]
		Therefore, analogous to the case of episodic tasks, $\frac{\partial V(\bm{\pi}^{\phi})}{\partial \phi}$ is the value corresponding to the long-term average of a different running reward, according to the ergodic Feynman--Kac formula (Lemma \ref{lemma:ergodic feynmann-kac}); that is
		\begin{equation}\label{erg}
			\begin{array}{rl}
				\frac{\partial V(\bm{\pi}^{\phi})}{\partial \phi} = & \liminf_{T\to \infty} \frac{1}{T}\E^{\p}\bigg[\int_0^T  \check{r}(X_t^{\bm{\pi}^{\phi}}, a_t^{\bm{\pi}^{\phi}};\phi) \dd t \Big| X_0^{\bm{\pi}^{\phi}} = x \bigg] \\
				= & \liminf_{T\to \infty} \frac{1}{T}\E^{\p^W}\bigg[\int_0^T \int_{\mathcal{A}} \check{r}(\tilde{X}_t^{\bm{\pi}^{\phi}}, a;\phi) \bm{\pi}^{\phi}(a|\tilde{X}_t^{\bm{\pi}^{\phi}})\dd a \dd t \Big|\tilde{X}_0^{\bm{\pi}^{\phi}} = x \bigg] \\
				= & \liminf_{T\to \infty} \frac{1}{T}\E^{\p^W}\bigg[\int_0^T \int_{\mathcal{A}} \big\{ \big[ \mathcal{L}^a J(\tilde{X}_t^{\bm{\pi}^{\phi}};\bm{\pi}^{\phi}) + r(\tilde{X}_t^{\bm{\pi}^{\phi}},a) + \gamma p\big(\tilde{X}_t^{\bm{\pi}^{\phi}},a,\bm{\pi}^{\phi}(\cdot|\tilde{X}_t^{\bm{\pi}^{\phi}}) \big)  \big] \\
				& \times \frac{\partial }{\partial \phi}\log \bm{\pi}^{\phi}(a|\tilde{X}_t^{\bm{\pi}^{\phi}}) + \gamma q(\tilde{X}_t^{\bm{\pi}^{\phi}}, a,\phi)\big\} \bm{\pi}^{\phi}(a|\tilde{X}_t^{\bm{\pi}^{\phi}}) \dd a \dd t \Big|\tilde{X}_0^{\bm{\pi}^{\phi}} = x \bigg] \\
				= & \liminf_{T\to \infty} \frac{1}{T}\E^{\p}\bigg[\int_0^T  \big\{ \frac{\partial }{\partial \phi}\log \bm{\pi}^{\phi}(a_t^{\bm{\pi}^{\phi}}|X_t^{\bm{\pi}^{\phi}}) \big[ \dd J(X_t^{\bm{\pi}^{\phi}};\bm{\pi}^{\phi}) + r(X_t^{\bm{\pi}^{\phi}}, a_t^{\bm{\pi}^{\phi}})\dd t \\
				& + \gamma p\big( X_t^{\bm{\pi}^{\phi}},a_t^{\bm{\pi}^{\phi}},\bm{\pi}^{\phi}(\cdot|X_t^{\bm{\pi}^{\phi}}) \big) \dd t\big]  + \gamma q(X_t^{\bm{\pi}^{\phi}},a_t^{\bm{\pi}^{\phi}},\phi) \dd t \big\} \Big|X_0^{\bm{\pi}^{\phi}} = x\bigg] \\
				= & \liminf_{T\to \infty} \frac{1}{T}\E^{\p}\bigg[\int_0^T  \big\{ \frac{\partial }{\partial \phi}\log \bm{\pi}^{\phi}(a_t^{\bm{\pi}^{\phi}}|X_t^{\bm{\pi}^{\phi}}) \big[ \dd J(X_t^{\bm{\pi}^{\phi}};\bm{\pi}^{\phi}) + r(X_t^{\bm{\pi}^{\phi}}, a_t^{\bm{\pi}^{\phi}})\dd t \\
				& + \gamma p\big( X_t^{\bm{\pi}^{\phi}},a_t^{\bm{\pi}^{\phi}},\bm{\pi}^{\phi}(\cdot|X_t^{\bm{\pi}^{\phi}}) \big) \dd t - V \dd t \big] + \gamma q(X_t^{\bm{\pi}^{\phi}},a_t^{\bm{\pi}^{\phi}},\phi) \dd t \big\} \Big|X_0^{\bm{\pi}^{\phi}} = x\bigg]  ,
		\end{array}  \end{equation}
		where the last equality is due to
		\[\begin{aligned}
			& \E^{\p}\bigg[\int_0^T V \frac{\partial }{\partial \phi}\log \bm{\pi}^{\phi}(a_t^{\bm{\pi}^{\phi}}|X_t^{\bm{\pi}^{\phi}})\dd t \Big|X_0^{\bm{\pi}^{\phi}} = x \bigg] \\
			= & V \E^{\p^W}\bigg[\int_0^T \dd t \int_{\mathcal{A}} \bm{\pi}^{\phi}(a|\tilde{X}_t^{\bm{\pi}^{\phi}}) \frac{\partial }{\partial \phi}\log \bm{\pi}^{\phi}(a|\tilde{X}_t^{\bm{\pi}^{\phi}})\dd a \Big|\tilde{X}_0^{\bm{\pi}^{\phi}} = x \bigg] \\
			= & V \E^{\p^W}\bigg[\int_0^T \dd t \frac{\partial }{\partial \phi}\int_{\mathcal{A}} \bm{\pi}^{\phi}(a|\tilde{X}_t^{\bm{\pi}^{\phi}})\dd a \Big|\tilde{X}_0^{\bm{\pi}^{\phi}} = x \bigg]  = 0.
		\end{aligned}  \]
		
		An ergodic task is a continuing task so we are naturally interested in  online algorithms only.
		We can design two algorithms based on the analysis above. The first one follows directly from the representation  (\ref{erg}), 
		in which the policy gradient is the expectation of a long-run average and hence can be estimated online incrementally by
		\[\begin{aligned}
			& \frac{\partial }{\partial \phi}\log \bm{\pi}^{\phi}(a_t^{\bm{\pi}^{\phi}}|X_t^{\bm{\pi}^{\phi}}) \Big[ \dd J(X_t^{\bm{\pi}^{\phi}};\bm{\pi}^{\phi}) + [r(X_t^{\bm{\pi}^{\phi}}, a_t^{\bm{\pi}^{\phi}}) \\
			& + \gamma p\big(X_t^{\bm{\pi}^{\phi}},a_t^{\bm{\pi}^{\phi}},\bm{\pi}^{\phi}(\cdot|X_t^{\bm{\pi}^{\phi}})\big) - V]\dd t \Big]  + \gamma q(X_t^{\bm{\pi}^{\phi}},a_t^{\bm{\pi}^{\phi}},\phi)\dd t
		\end{aligned}  \]
		since it will converge to its stationary distribution as $t\to \infty$.\footnote{To be more specific, the reason why an infinitesimal increment of the (inner) integral can be used as an estimate for the gradient is due to the ergodicity of the state process. The expression of the gradient \eqref{erg} is the long-time average of the integrand of the inner integral, which converges to its expectation with respect to the stationary measure. On the other hand, the distribution of the integrand itself also converges to its stationary measure. Therefore, the integrand itself becomes an asymptotically unbiased estimate for the gradient as time tends to infinity. For a brief summary of the ergodicity properties, see \cite{sandric2017note}. More details can be found in Part III of \cite{meyn2012markov}. }

		Moreover, due to the martingale orthogonality condition \eqref{eq:pe martingale ergodic}, we can also add a test function $\zeta$ as we did in \eqref{eq:policy gradient expectation baseline in loglikelihood}. Consequently, the algorithm  updates $\phi$ by gradient ascent:
		\[\begin{aligned}
			\phi \leftarrow \phi + \alpha_{\phi}\bigg\{ & \big[ \frac{\partial }{\partial \phi}\log \bm{\pi}^{\phi}(a_t^{\bm{\pi}^{\phi}}|X_t^{\bm{\pi}^{\phi}}) + \zeta_t\big] \Big[ \dd J(X_t^{\bm{\pi}^{\phi}};\bm{\pi}^{\phi}) \\
			&+ [r(X_t^{\bm{\pi}^{\phi}}, a_t^{\bm{\pi}^{\phi}}) + \gamma p\big(X_t^{\bm{\pi}^{\phi}},a_t^{\bm{\pi}^{\phi}},\bm{\pi}^{\phi}(\cdot|X_t^{a})\big) - V]\dd t \Big]  + \gamma q(X_t^{\bm{\pi}^{\phi}},a_t^{\bm{\pi}^{\phi}},\phi)\dd t \bigg\} .
		\end{aligned}\]
		
		The second algorithm applies a test function $\eta$ and  stochastic approximation to solve the optimality condition  as in Theorem \ref{prop:pg optimal policy condition}, by updating
		\[\begin{aligned}
			\phi \leftarrow \phi + \alpha_{\phi}\eta_t\bigg\{ & \big[ \frac{\partial }{\partial \phi}\log \bm{\pi}^{\phi}(a_t^{\bm{\pi}^{\phi}}|X_t^{\bm{\pi}^{\phi}})+\zeta_t \big] \Big[ \dd J(X_t^{\bm{\pi}^{\phi}};\bm{\pi}^{\phi}) \\
			& + [r(X_t^{\bm{\pi}^{\phi}}, a_t^{\bm{\pi}^{\phi}}) + \gamma p\big(X_t^{\bm{\pi}^{\phi}},a_t^{\bm{\pi}^{\phi}},\bm{\pi}^{\phi}(\cdot|X_t^{\bm{\pi}^{\phi}})\big) - V]\dd t \Big] + \gamma q(X_t^{\bm{\pi}^{\phi}},a_t^{\bm{\pi}^{\phi}},\phi)\dd t \bigg\} .
		\end{aligned}\]
		
		Observe the two algorithms above differ by only the presence of the test function $\eta$.
		To illustrate, we describe the second one in Algorithm \ref{algo:ergodic incremental}.
		

		\begin{algorithm}[htbp]
			\caption{Actor--Critic Algorithm for Ergodic Tasks}
			\textbf{Inputs}: initial state $x_0$, time step $\Delta t$, initial learning rates $\alpha_{\theta},\alpha_{\phi},\alpha_{V}$ and learning rate schedule function $l(\cdot)$ (a function of time), functional form of the value function $J^{\theta}(\cdot)$, functional form of the policy $\bm{\pi}^{\phi}(\cdot|\cdot)$, functional form of the regularizer $p\big(x,a,\pi(\cdot) \big)$, functional forms of  test functions $\bm{\xi}(x_{\cdot \wedge t}),\bm{\eta}(x_{\cdot \wedge t}),\bm{\zeta}(x_{\cdot \wedge t})$, and temperature parameter $\gamma$.
			
			\textbf{Required program}: an environment simulator $(x',r) = \textit{Environment}_{\Delta t}(x,a)$ that takes initial state $x$ and action $a$ as inputs and generates a new state $x'$ (at $\Delta t$) and an instantaneous reward $r$.

			\textbf{Learning procedure}:
			\begin{algorithmic}
				\STATE Initialize $\theta,\phi,V$. Initialize $k = 0$. Observe the initial state $x_0$ and store $x_{t_k} \leftarrow  x_0$.
				
				\LOOP \STATE{
					Compute test function $\xi_{t_k} = \bm{\xi}(x_{t_0},\cdots, x_{t_k})$, $\eta_{t_k} = \bm{\eta}(x_{t_0},\cdots, x_{t_k})$ and $\zeta_{t_k} = \bm{\eta}(x_{t_0},\cdots, x_{t_k})$.	
					
					Generate action $a\sim \bm{\pi}^{\phi}(\cdot|x)$.
					
					Apply $a$ to the environment simulator $(x',r) = Environment_{\Delta t}(x, a)$, and observe the output new state $x'$ and reward $r$. Store $x_{t_{k+1}} \leftarrow x'$.
					
					Compute
					\[\begin{aligned}
						& \delta = J^{\theta}(x') - J^{\theta}(x) + r\Delta t + \gamma p\big(x,a,\bm{\pi}^{\phi}(\cdot|x)\big)\Delta t - V \Delta t, \\
						& \Delta \theta = \xi_{t_k} \delta, \\
						& \Delta V = \delta, \\
						& \Delta \phi =  \eta_{t_k} \bigg\{ \big[ \frac{\partial}{\partial \phi}\log\bm{\pi}^{\phi}(a|x)   + \zeta_{t_k}\big]\delta + \gamma  \frac{\partial p}{\partial \phi}\big( x,a,\bm{\pi}^{\phi}(\cdot|x)  \big)\Delta t \bigg\}.
					\end{aligned}  \]
					
					Update $\theta$ and $V$ (policy evaluation) by
					\[ \theta \leftarrow \theta + l(k\Delta t)\alpha_{\theta} \Delta \theta,\]
					\[ V \leftarrow V + l(k\Delta t)\alpha_{V} \Delta V.\]
					Update $\phi$ (policy gradient) by
					\[ \phi \leftarrow \phi + l(k\Delta t)\alpha_{\phi}   \Delta \phi.  \]
					
					Update $x\leftarrow x'$ and $k \leftarrow k + 1$.
				}
				\ENDLOOP	
				
			\end{algorithmic}
			\label{algo:ergodic incremental}
		\end{algorithm}
		
		\section{Applications}
		\label{sec:applications}
		In this section we report simulation experiments on our  algorithms in  two applications. The first one is mean--variance portfolio selection in a finite time horizon with multiple episodes of simulated stock price data. The second application is ergodic linear--quadratic control with a single sample trajectory.
		\subsection{Mean--Variance Portfolio Selection}
		\label{sec:application mv}
		We first review the formulation of the exploratory mean--variance portfolio selection problem proposed by \citet{wang2020continuous}. The investment universe consists of one risky asset (e.g. a stock index) and one risk-free asset (e.g. a saving account)  whose risk-free interest rate is $r$. The price of the risky asset is governed by a geometric Brownian motion with mean $\mu$ and volatility $\sigma>0$ on a filtered probability space $(\Omega,\f,\p^W;\{\f_t^W\}_{0\leq t\leq T})$:
		\begin{equation}
			\label{eq:model stock}
			\frac{\dd S_t}{S_t}=\mu\dd t+\sigma\dd W_t.
		\end{equation}
		Denote by $\rho = \frac{\mu-r}{\sigma}$ the Sharpe ratio of the risky asset.
		
		An agent has a fixed investment horizon $0<T<\infty$ and an initial endowment $x_0$.
		A self-financing portfolio is represented by the real-valued adapted process $a = \{a_t, 0\leq t \leq T\}$, where $a_t$ is the discounted dollar value invested in the risky asset at time $t$. Then the discounted value of this portfolio satisfies the wealth equation
		\begin{equation}
			\label{eq:wealth process}
			\dd x_t^{a} = a_t[(\mu-r) \dd t + \sigma\dd W_t] = a_t \frac{\dd (e^{-rt}S_t)}{e^{-rt}S_t}, \ x_0^{a} = x_0,
		\end{equation}
		where $e^{-rt}S_t$ is the discounted stock price. 
		We stress that the model on the stock price \eqref{eq:model stock} is mainly for theoretical analysis and for generating samples in our simulation; we do not assume that the agent knows its parameters. 
		
		The agent has the mean--variance preference, namely, she aims to minimize the variance of the discounted value of the portfolio at $T$ while achieving  a given level of expected return:
		\begin{equation}
			\label{eq:mv} \min_{a} \text{Var}(x_T^{a}),\ \mbox{subject to }\ \E[x_T^{a}] = z,
		\end{equation}
		where $z$ is the target value, and the variance and expectation are w.r.t. the probability measure $\p^W$. 
		
		This problem is not a standard stochastic control  problem and cannot be solved directly by the dynamic programming (DP) principle, or any DP-based reinforcement learning algorithms such as Q-learning. This is because  the variance term causes {\it time-inconsistency} which violates the assumptions of DP. \citet{strotz1955myopia} discusses three types of agents when facing time-inconsistency. Here, we consider  one of them -- the  so-called {\it{pre-committed}} agent who solves the problem  at time 0 and sticks to it afterwards.\footnote{The other two types are the {\it na\"ive} one who re-optimizes at any given time and the {\it sophisticated} one who seeks  subgame perfect Nash equilibria among her-selves at different times.
			The latter has been well studied in the continuous-time setting in recent years; see
			e.g. \citet{ekeland2006being,bjork2014mean,basak2010dynamic,dai2021dynamic}. The RL counterpart  is studied in \cite{DDJ2020}.}
		For this type of agent,
		to overcome the difficulty of DP not being directly applicable, \citet{zhou2000continuous} extend the embedding method, initially introduced by \citet{li2000optimal} for the discrete-time mean--variance problem, to transform \eqref{eq:mv} into an equivalent, unconstrained, and expectation-only problem:
		\[\min_{a} \E[(x_T^{a})^2] -z^2 - 2w(\E[x_T^{a}] - z) = \min_{a}\E[(x_T^{a} - w)^2] - (w-z)^2,  \]
		where $w$ is the Lagrange multiplier associated with the constraint  $\E[x^{a}_T] = z$. This new problem is time-consistent and therefore can be solved by DP. Once the optimal $a^*$ is derived, $w$ can  be obtained by the equation $\E[x^{a^*}_T] = z$.

		In a reinforcement learning framework, \citet{wang2020continuous} allow randomized actions to incorporate exploration. A stochastic policy  is denoted by
		$\bm{\pi}=\bm{\pi}(\cdot|t, x)$, namely, at any current  time--wealth pair  $(t,x)$, the total amount of discounted wealth invested in the stock is a random draw from the distribution with the density function $\bm{\pi}(\cdot|t, x)$. Under such a policy, we denote by $\tilde{X}^{\bm{\pi}}=\{\tilde{X}^{\bm{\pi}}_s:t\leq s\leq T\}$  the solution to the following SDE
		\[ \dd \tilde{X}^{\bm{\pi}}_s = (\mu-r)\int_{\mathbb{R}} a \bm{\pi}(a|s, \tilde{X}^{\bm{\pi}}_s)  \dd a \dd s + \sigma \sqrt{\int_{\mathbb{R}} a^2 \bm{\pi}(a|s, \tilde{X}^{\bm{\pi}}_s)  \dd a } \dd W_s;\;\tilde{X}_t^{\bm{\pi}} = x, \]
		which is \eqref{eq:model relaxed} specializing to the current case.

		Moreover, an entropy regularizer is added to incentivize exploration. Mathematically, the entropy-regularized mean--variance portfolio choice problem is to solve
		\begin{equation}
			\label{eq:entropy mv objective function}
			V(t,x;w) = \min_{\bm{\pi}} \E\bigg[(\tilde{X}_T^{\bm{\pi}} - w)^2 - \gamma \int_t^T \mathcal{H}(\pi_s)\dd s\Big| \tilde{X}_t^{\bm{\pi}} = x \bigg] - (w-z)^2,
		\end{equation}
		where $z$ is the target expected terminal wealth,
		$\pi_s=\bm{\pi}(\cdot|s, \tilde{X}_s^{\bm{\pi}}), \;t\leq s\leq T$,
		$\mathcal{H}$ is the differential entropy $\mathcal{H}(\pi)=-\int_{{\cal A}}\pi(a)\log \pi(a)\dd a$, $\gamma$ is  the temperature parameter, and  $w$ is the Lagrange multiplier similar to that introduced earlier. 

		We follow \citet{wang2020continuous} to parameterize the value function by
		\[ J^{\theta}(t,x;w) = (x - w)^2 e^{-\theta_3(T-t)} + \theta_2(t^2 - T^2) + \theta_1(t-T) - (w-z)^2,  \]
		and parameterize the policy by
		\[ \bm{\pi}^{\phi}(\cdot|t,x;w) = \mathcal{N}(\cdot|-\phi_1(x-w), e^{\phi_2 + \phi_3(T-t)}) , \]
		where $\mathcal{N}(\cdot|\alpha,\delta^2)$ is the pdf of the normal distribution with mean $\alpha$ and variance $\delta^2$.
		These function approximators are derived in \citet{wang2020continuous} by exploiting the special structure of the underlying problem; see also Appendix B1.
		
		There is no running  reward from the actions except  the regularizer
		\[ 
		\mathcal{H}(\bm{\pi}^{\phi}(\cdot|t,x;w))=-\frac{1}{2}\log (2\pi e) - \frac{1}{2}[\phi_2 + \phi_3(T-t)]=:\hat{p} (t,\phi). \]
		Note that the regularizer turns out to be independent of the state $x$.
		Finally, the discount factor is $\beta = 0$.
		
		From this point on, we depart from \citet{wang2020continuous} and instead apply the methods developed in this paper to solve the problem.
		We choose the test functions for PE as the following gradients, in accordance with the most  popular $TD(0)$ algorithm:\footnote{\citet{wang2020continuous}  employ a mean--square TD error (MSTDE) algorithm to do  PE and a policy improvement theorem to update policies. However, it is shown in \cite{jia2021policy} that MSTDE only minimizes the qudratic variation of the martingale, which may not lead to the true solution of PE. As discussed earlier, other PE algorithms proposed  in \cite{jia2021policy} can also be applied.}
		\[  \frac{\partial J^{\theta}}{\partial \theta_1} (t,x;w) = t-T,\;\;
		\frac{\partial J^{\theta}}{\partial \theta_2} (t,x;w) = t^2-T^2,\;\;
		\frac{\partial J^{\theta}}{\partial \theta_3} (t,x;w) = (x-w)^2e^{-\theta_3(T-t)}(t-T) . \]
		The PE updating rule is
		\[ \theta \leftarrow \theta + \alpha_{\theta} \int_0^T \frac{\partial J^{\theta}}{\partial \theta}(t,X_t^{\bm{\pi}^{\phi}};w) \left[\dd J^{\theta}(t,X_t^{\bm{\pi}^{\phi}};w) + \gamma \hat{p}(t,\phi)\dd t \right].  \]
		For the PG part, the gradients of log-likelihood are
		\[ \frac{\partial \log \bm{\pi}^{\phi}(a|t,x;w)}{\partial \phi_1} = -\big( a + \phi_1(x-w) \big) (x-w) e^{-\phi_2 - \phi_3(T-t)},  \]
		\[ \frac{\partial \log \bm{\pi}^{\phi}(a|t,x;w)}{\partial \phi_2} = -\frac{1}{2} + \frac{\big(a + \phi_1(x-w) \big)^2}{2}e^{-\phi_2 - \phi_3(T-t)} , \]
		\[ \frac{\partial \log \bm{\pi}^{\phi}(a|t,x;w)}{\partial \phi_3} = -\frac{T-t}{2} + \frac{\big(a + \phi_1(x-w) \big)^2}{2}e^{-\phi_2 - \phi_3(T-t)} (T-t),\]
		and those  of the regularizer are
		\[ \frac{\partial \hat{p}}{\partial \phi_1}(t,\phi) = 0, \;\; \frac{\partial \hat{p}}{\partial \phi_2}(t,\phi) = -\frac{1}{2}, \;\;\frac{\partial \hat{p}}{\partial \phi_3}(t,\phi) = -\frac{T-t}{2}. \]
		Accordingly, the offline PG updating rule is
		\[\begin{aligned}
			\phi \leftarrow \phi - \alpha_{\phi}\int_0^T\bigg\{ & \frac{\partial \log \bm{\pi}^{\phi}}{\partial \phi} (a_t|t,X_t^{\bm{\pi}^{\phi}};w)  \left[\dd J^{\theta}(t,X_t^{\bm{\pi}^{\phi}};w) + \gamma \hat{p}(t,X_t^{\bm{\pi}^{\phi}},\phi)\dd t \right] \\
			& + \gamma \frac{\partial \hat{p}}{\partial \phi}(t,X_t^{\bm{\pi}^{\phi}},\phi) \dd t  \bigg\}.
		\end{aligned}  \]
		The online counterpart of this updating rule is to remove the integral ``$\int_0^T$'' in the above and use only the resulting increment to update the policy at every time step.
		
		In addition, there is the Lagrange multiplier $w$ we need to learn: we update $w$ based on the same stochastic approximation scheme  in \cite{wang2020continuous}.
		

		We present our offline and online algorithms as Algorithms \ref{algo:offline episodic mv} and \ref{algo:online episodic mv} respectively.  
		Then we replicate the simulation study of \citet{wang2020continuous} with the same basic setting: $x_0=1$, $z=1.4$, $T=1$, $\Delta t = \frac{1}{252}$. Choose temperature parameter $\gamma = 0.1$. The batch size $m=10$ for updating the Lagrange multiplier. The learning rate parameters in \citet{wang2020continuous} are set to be $\alpha_w=0.05$, and $\alpha_{\theta} = \alpha_{\phi} = 0.0005$ with decay rate $l(j) = j^{-0.51}$. In our experiment we adopt these learning rate values for the \citet{wang2020continuous} algorithm unless the algorithm does not converge, in which case we tune the initial learning rates to guarantee convergence.
		For our algorithm, we set $\alpha_w=0.05$, and $\alpha_{\theta} = \alpha_{\phi} = 0.1$ with decay rate $l(j) = j^{-0.51}$ and tune the initial learning rate when necessary. The initialization of the parameters $\theta$ and $\phi$ is set to be all 0 for both algorithms (the initialization is not discussed in \citealt{wang2020continuous}). In particular, to mimic the real scenario, we choose a reasonable size of the training sample, with length of 20 years. In each iteration, we randomly sample 128 1-year trajectories to update the rest parameters, and we train the model for $N=2\times 10^4$ iterations. We calculate the performance metrics -- the mean, variance and Sharpe ratio of the resulting terminal wealth -- of the learned policies of both methods with the training set generated from  the same distribution.\footnote{\citet{wang2020continuous} report in-sample performance of the last 2000 iterations in the training set but does not present out-of-sample test results.} 
We then repeat the experiment for 100 times and report the standard deviation of each metric.

\begin{algorithm}[htbp]
	\caption{Offline--Episodic Actor--Critic Mean--Variance Algorithm}
	\textbf{Inputs}: initial state $x_0$, horizon $T$, time step $\Delta t$, number of episodes $N$, number of time grids $K$, initial learning rates $\alpha_{\theta},\alpha_{\phi},\alpha_w$ and learning rate schedule function $l(\cdot)$ (a function of the number of episodes), and temperature parameter $\gamma$.
	
	\textbf{Required program}: a market simulator $x' = \textit{Market}_{\Delta t}(t,x,a)$ that takes current time-state pair $(t,x)$ and action $a$ as inputs and generates state $x'$ at time $t+\Delta t$.

	\textbf{Learning procedure}:
	\begin{algorithmic}
		\STATE Initialize $\theta,\phi,w$.
		\FOR{episode $j=1$ \TO $N$} \STATE{Initialize $k = 0$. Observe the initial state $x$ and store $x_{t_k} \leftarrow  x$.
			\WHILE{$k < K$} \STATE{
				Compute and store the test function $\xi_{t_k} = \frac{\partial J^{\theta}}{\partial \theta}(t_k,x_{t_k};w)$.	
				
				Generate action $a_{t_k}\sim \bm{\pi}^{\phi}(\cdot|t_k,x_{t_k})$.
				
				Apply $a_{t_k}$ to the market simulator $x = Market_{\Delta t}(t_k, x_{t_k}, a_{t_k})$, and observe the output new state $x$. Store $x_{t_{k+1}} \leftarrow x$.
				
				Update $k \leftarrow k + 1$.
			}
			\ENDWHILE	
			
			Store the terminal wealth $X_T^{(j)} \leftarrow x_{t_K}$.
			
			Compute
			\[ \Delta \theta = \sum_{i=0}^{K-1} \xi_{t_i} \big[ J^{\theta}(t_{i+1},x_{t_{i+1}};w) - J^{\theta}(t_{i},x_{t_{i}};w)  + \gamma \hat{p}(t_i,x_{t_i},\phi)\Delta t  \big], \]
			\[ \begin{aligned}
				\Delta \phi = \sum_{i=0}^{K-1} & \frac{\partial}{\partial \phi}\log\bm{\pi}^{\phi}(a_{t_i}|t_i,x_{t_i})\big[ J^{\theta}(t_{i+1},x_{t_{i+1}}) - J^{\theta}(t_{i},x_{t_{i}})+ \gamma \hat{p}(t,x_{t_i},\phi)\Delta t   \big] \\
				& + \gamma \frac{\partial \hat{p}}{\partial \phi}(t_i,x_{t_i},\phi)\Delta t .
			\end{aligned} \]
			
			Update $\theta$ (policy evaluation) by
			\[ \theta \leftarrow \theta + l(j)\alpha_{\theta} \Delta \theta .\]
			
			Update $\phi$ (policy gradient) by
			\[ \phi \leftarrow \phi - l(j)\alpha_{\phi} \Delta \phi .  \]
			
			\textbf{Update $w$ (Lagrange multiplier) every $m$ episodes:}
			\IF{$j \equiv 0 \mod m$} \STATE{
				\[ w\leftarrow w - \alpha_w \frac{1}{m}\sum_{i=j-m+1}^{j} X_T^{(i)}.\]
			}
			\ENDIF
		}
		\ENDFOR
	\end{algorithmic}
	\label{algo:offline episodic mv}
\end{algorithm}

\begin{algorithm}[htbp]
	\caption{Online--Episodic Actor--Critic Mean--Variance Algorithm}
	\textbf{Inputs}: initial state $x_0$, horizon $T$, time step $\Delta t$, number of episodes $N$, number of time grids $K$, initial learning rates $\alpha_{\theta},\alpha_{\phi},\alpha_w$ and learning rate schedule function $l(\cdot)$ (a function of the number of episodes), and temperature parameter $\gamma$.
	
	\textbf{Required program}: a market simulator $x' = \textit{Market}_{\Delta t}(t,x,a)$ that takes current time-state pair $(t,x)$ and action $a$ as inputs and generates state $x'$ at time $t+\Delta t$.

	\textbf{Learning procedure}:
	\begin{algorithmic}
		\STATE Initialize $\theta,\phi,w$.
		\FOR{episode $j=1$ \TO $N$} \STATE{Initialize $k = 0$. Observe the initial state $x$ and store $x_{t_k} \leftarrow  x$.
			\WHILE{$k < K$} \STATE{
				Compute and store the test function $\xi_{t_k} = \frac{\partial J^{\theta}}{\partial \theta}(t_k,x_{t_k};w)$.	
				
				Generate action $a_{t_k}\sim \bm{\pi}^{\phi}(\cdot|t_k,x_{t_k})$.
				
				Apply $a_{t_k}$ to the market simulator $x = Market_{\Delta t}(t_k, x_{t_k}, a_{t_k})$, and observe the output new state $x$. Store $x_{t_{k+1}} \leftarrow x$.
				
				Compute
				\[ \Delta \theta = \xi_{t_k} \big[ J^{\theta}(t_{k+1},x_{t_{k+1}};w) - J^{\theta}(t_{k},x_{t_{k}};w)  + \gamma \hat{p}(t_k,x_{t_k},\phi)\Delta t  \big], \]
				\[ \begin{aligned}
					\Delta \phi = & \frac{\partial}{\partial \phi}\log\bm{\pi}^{\phi}(a_{t_k}|t_k,x_{t_k})\big[ J^{\theta}(t_{k+1},x_{t_{k+1}}) - J^{\theta}(t_{k},x_{t_{k}})+ \gamma \hat{p}(t,x_{t_k},\phi)\Delta t   \big] \\
					& + \gamma \frac{\partial \hat{p}}{\partial \phi}(t_k,x_{t_k},\phi)\Delta t .
				\end{aligned} \]
				
				Update $\theta$ (policy evaluation) by
				\[ \theta \leftarrow \theta + l(j)\alpha_{\theta} \Delta \theta .\]
				
				Update $\phi$ (policy gradient) by
				\[ \phi \leftarrow \phi - l(j)\alpha_{\phi} \Delta \phi .  \]
				
				Update $k \leftarrow k + 1$.
			}
			\ENDWHILE	
			
			Store the terminal wealth $X_T^{(j)} \leftarrow x_{t_K}$.

			\textbf{Update $w$ (Lagrange multiplier) every $m$ episodes:}
			\IF{$j \equiv 0 \mod m$} \STATE{
				\[ w\leftarrow w - \alpha_w \frac{1}{m}\sum_{i=j-m+1}^{j} X_T^{(i)}.\]
			}
			\ENDIF
		}
		\ENDFOR
	\end{algorithmic}
	\label{algo:online episodic mv}
\end{algorithm}

Tables \ref{tab:mv wang td} and \ref{tab:mv pg offline} present the test results of the algorithm in \cite{wang2020continuous} and the offline Algorithm \ref{algo:offline episodic mv} in this paper respectively, when stock price is generated from  geometric Brownian motion under different specifications of the market parameters $\mu$ and $\sigma$. Our algorithm achieves significantly higher out-of-sample average Sharpe ratios in most scenarios. Underperformance of our strategy occurs mainly when the actual return of the stock is low ($\mu=0,\pm 0.1$). 
In those cases, our learned policy yields  larger volatility and less stable out-of-sample performance. However, although the average out-of-sample Sharpe ratios of the learned policy are lower than those of \cite{wang2020continuous}, the standard deviations of ours are large. Hence, even in those few scenarios it  is still statistically inconclusive to determine which method is better.

We further carry out tests to compare Algorithm \ref{algo:offline episodic mv} with the online Algorithm \ref{algo:online episodic mv}. In implementing Algorithm \ref{algo:online episodic mv},  we update parameters at each time step and conduct learning for 20 years, with the same simulated stock prices as in  offline learning.  The batch size is set to be $m=1$ for updating the Lagrange multiplier. We also repeat the experiment for 100 times to calculate the standard deviation of each metric. The results are presented in Table \ref{tab:mv pg online}. Compared with Table \ref{tab:mv pg offline},  offline learning outperforms  online one in terms of Sharpe ratio  in most cases. Moreover, the former is always preferred when it comes to stably reaching the target return (set to be 40\% annually  in the experiments).

With a given training data set, it is not surprising that offline learning is typically preferred because it allows us to fully use the data set by bootstrapping multiple 1-year episodes. By contrast, online learning pretends the data to come sequentially without storing past data. For example, under our online setting, the 20-year training set only contains 20 complete episodes sequentially to adjust the final terminal wealth level, unlike in the offline setting where we bootstrap multiple 1-year episodes. Therefore, offline learning uses data thoroughly and efficiently. However, other important considerations motivate or even force us to use online learning. First and foremost, data distribution may not be stationary, so offline learning may suffer from overfitting. Second, for large-scale problems, online and incremental learning is more computationally efficient in reducing storage costs and computational time. Finally, there are also computational techniques to store a certain amount of past data to boost the efficiency of online learning, such as the experience replay with off-policy learning \citep{zhang2017deeper,fedus2020revisiting}.

\begin{table}[htbp]
	\centering
	\caption{\textbf{Out-of-sample performance of algorithm proposed in \cite{wang2020continuous} for mean--variance problem when data are generated by geometric Brownian motion.} The columns ``Mean'' and ``Variance'' report the average  mean and variance, respectively,  of terminal wealth over 100 independent experiments. The column ``Sharpe ratio" reports the corresponding average Sharpe ratio ($\frac{\text{Mean} - 1}{\sqrt{\text{Variance}}}$).
		The numbers in the brackets are the standard deviations. }
	\begin{tabular}{ccccc}
		\toprule
		$\mu$    & $\sigma$ & Mean  & Variance & Sharpe ratio \\
		\midrule
		-0.5  & 0.1   & 1.4 ( 0.015 ) & 0 ( 0.00033 ) & 6.69 ( 0.096 ) \\
		-0.3  & 0.1   & 1.4 ( 0.027 ) & 0.01 ( 0.002 ) & 3.59 ( 0.064 ) \\
		-0.1  & 0.1   & 1.4 ( 0.11 ) & 0.11 ( 0.059 ) & 1.25 ( 0.02 ) \\
		0     & 0.1   & 1.04 ( 0.028 ) & 0.06 ( 0.057 ) & 0.2 ( 9.2e-05 ) \\
		0.1   & 0.1   & 1.45 ( 0.29 ) & 0.43 ( 0.55 ) & 0.81 ( 0.0057 ) \\
		0.3   & 0.1   & 1.41 ( 0.033 ) & 0.02 ( 0.0032 ) & 3.06 ( 0.046 ) \\
		0.5   & 0.1   & 1.4 ( 0.017 ) & 0 ( 0.00044 ) & 6 ( 0.087 ) \\
		-0.5  & 0.2   & 1.4 ( 0.03 ) & 0.01 ( 0.0031 ) & 3.38 ( 0.15 ) \\
		-0.3  & 0.2   & 1.41 ( 0.067 ) & 0.05 ( 0.022 ) & 1.81 ( 0.074 ) \\
		-0.1  & 0.2   & 1.44 ( 0.3 ) & 0.74 ( 1.3 ) & 0.61 ( 0.0041 ) \\
		0     & 0.2   & 1.04 ( 0.032 ) & 0.27 ( 0.27 ) & 0.1 ( 6.3e-05 ) \\
		0.1   & 0.2   & 1.25 ( 0.076 ) & 0.43 ( 0.23 ) & 0.4 ( 0.0014 ) \\
		0.3   & 0.2   & 1.42 ( 0.089 ) & 0.08 ( 0.04 ) & 1.54 ( 0.055 ) \\
		0.5   & 0.2   & 1.41 ( 0.034 ) & 0.02 ( 0.004 ) & 3.03 ( 0.13 ) \\
		-0.5  & 0.3   & 1.4 ( 0.057 ) & 0.03 ( 0.016 ) & 2.25 ( 0.16 ) \\
		-0.3  & 0.3   & 1.41 ( 0.14 ) & 0.13 ( 0.12 ) & 1.2 ( 0.057 ) \\
		-0.1  & 0.3   & 1.32 ( 0.12 ) & 0.71 ( 0.43 ) & 0.41 ( 0.0023 ) \\
		0     & 0.3   & 1.04 ( 0.031 ) & 0.55 ( 0.58 ) & 0.07 ( 3e-05 ) \\
		0.1   & 0.3   & 1.19 ( 0.11 ) & 0.67 ( 0.49 ) & 0.27 ( 0.0011 ) \\
		0.3   & 0.3   & 1.44 ( 0.14 ) & 0.22 ( 0.21 ) & 1 ( 0.018 ) \\
		0.5   & 0.3   & 1.41 ( 0.055 ) & 0.04 ( 0.016 ) & 2.01 ( 0.13 ) \\
		-0.5  & 0.4   & 1.41 ( 0.079 ) & 0.07 ( 0.041 ) & 1.67 ( 0.13 ) \\
		-0.3  & 0.4   & 1.43 ( 0.15 ) & 0.28 ( 0.23 ) & 0.86 ( 0.011 ) \\
		-0.1  & 0.4   & 1.28 ( 0.12 ) & 1.04 ( 0.76 ) & 0.3 ( 0.0016 ) \\
		0     & 0.4   & 1.04 ( 0.028 ) & 0.85 ( 0.89 ) & 0.05 ( 2.3e-05 ) \\
		0.1   & 0.4   & 1.17 ( 0.1 ) & 0.93 ( 0.82 ) & 0.2 ( 0.00069 ) \\
		0.3   & 0.4   & 1.46 ( 0.17 ) & 0.44 ( 0.43 ) & 0.74 ( 0.012 ) \\
		0.5   & 0.4   & 1.42 ( 0.082 ) & 0.09 ( 0.046 ) & 1.44 ( 0.058 ) \\
		\bottomrule
	\end{tabular}%
	\label{tab:mv wang td}%
\end{table}%

\begin{table}[htbp]
	\centering
	\caption{\textbf{Out-of-sample performance of offline learning (Algorithm \ref{algo:offline episodic mv}) for mean--variance problem when data are generated by geometric Brownian motion.} The columns ``Mean'' and ``Variance'' report the average  mean and variance, respectively,  of terminal wealth over 100 independent experiments. The column ``Sharpe ratio" reports the corresponding average Sharpe ratio ($\frac{\text{Mean} - 1}{\sqrt{\text{Variance}}}$).
		The numbers in the brackets are the standard deviations. }
	\begin{tabular}{ccccc}
		\toprule
		$\mu$    & $\sigma$ & Mean  & Variance & Sharpe ratio \\
		\midrule
		-0.5  & 0.1   & 1.4 ( 0.012 ) & 0 ( 0.00011 ) & 8.15 ( 0.06 ) \\
		-0.3  & 0.1   & 1.4 ( 0.023 ) & 0.01 ( 0.00084 ) & 4.37 ( 0.029 ) \\
		-0.1  & 0.1   & 1.41 ( 0.08 ) & 0.09 ( 0.037 ) & 1.37 ( 0.0073 ) \\
		0     & 0.1   & 1.13 ( 0.14 ) & 0.91 ( 0.49 ) & 0.12 ( 0.16 ) \\
		0.1   & 0.1   & 1.51 ( 0.27 ) & 0.47 ( 0.72 ) & 0.84 ( 0.0023 ) \\
		0.3   & 0.1   & 1.41 ( 0.028 ) & 0.01 ( 0.0015 ) & 3.71 ( 0.025 ) \\
		0.5   & 0.1   & 1.4 ( 0.014 ) & 0 ( 0.00016 ) & 7.35 ( 0.055 ) \\
		-0.5  & 0.2   & 1.4 ( 0.025 ) & 0.01 ( 0.001 ) & 3.98 ( 0.047 ) \\
		-0.3  & 0.2   & 1.4 ( 0.049 ) & 0.04 ( 0.0088 ) & 2.1 ( 0.017 ) \\
		-0.1  & 0.2   & 1.53 ( 0.27 ) & 0.93 ( 1 ) & 0.62 ( 0.0012 ) \\
		0     & 0.2   & 1.06 ( 0.15 ) & 2.51 ( 1.6 ) & 0.04 ( 0.094 ) \\
		0.1   & 0.2   & 1.44 ( 0.37 ) & 2.02 ( 1.5 ) & 0.35 ( 0.21 ) \\
		0.3   & 0.2   & 1.42 ( 0.065 ) & 0.06 ( 0.018 ) & 1.78 ( 0.012 ) \\
		0.5   & 0.2   & 1.41 ( 0.029 ) & 0.01 ( 0.0015 ) & 3.58 ( 0.041 ) \\
		-0.5  & 0.3   & 1.4 ( 0.04 ) & 0.03 ( 0.0047 ) & 2.54 ( 0.026 ) \\
		-0.3  & 0.3   & 1.41 ( 0.088 ) & 0.1 ( 0.049 ) & 1.32 ( 0.007 ) \\
		-0.1  & 0.3   & 1.43 ( 0.33 ) & 1.79 ( 1.9 ) & 0.37 ( 0.18 ) \\
		0     & 0.3   & 1.03 ( 0.12 ) & 3.46 ( 2.3 ) & 0.02 ( 0.064 ) \\
		0.1   & 0.3   & 1.26 ( 0.36 ) & 2.78 ( 2.3 ) & 0.18 ( 0.2 ) \\
		0.3   & 0.3   & 1.44 ( 0.13 ) & 0.17 ( 0.14 ) & 1.12 ( 0.012 ) \\
		0.5   & 0.3   & 1.41 ( 0.048 ) & 0.03 ( 0.0076 ) & 2.28 ( 0.02 ) \\
		-0.5  & 0.4   & 1.41 ( 0.061 ) & 0.05 ( 0.017 ) & 1.8 ( 0.01 ) \\
		-0.3  & 0.4   & 1.43 ( 0.15 ) & 0.24 ( 0.23 ) & 0.93 ( 0.014 ) \\
		-0.1  & 0.4   & 1.31 ( 0.44 ) & 3.13 ( 4.8 ) & 0.25 ( 0.17 ) \\
		0     & 0.4   & 1.02 ( 0.096 ) & 3.77 ( 2.7 ) & 0.01 ( 0.049 ) \\
		0.1   & 0.4   & 1.14 ( 0.36 ) & 3.6 ( 2.8 ) & 0.1 ( 0.18 ) \\
		0.3   & 0.4   & 1.53 ( 0.34 ) & 0.74 ( 1.3 ) & 0.73 ( 0.0024 ) \\
		0.5   & 0.4   & 1.42 ( 0.076 ) & 0.07 ( 0.032 ) & 1.6 ( 0.014 ) \\
		\bottomrule
	\end{tabular}%
	\label{tab:mv pg offline}%
\end{table}%

\begin{table}[htbp]
	\centering
	\caption{\textbf{Out-of-sample performance of online learning (Algorithm \ref{algo:online episodic mv}) for mean--variance problem when data are generated by geometric Brownian motion.} The columns ``Mean'' and ``Variance'' report the average  mean and variance, respectively,  of terminal wealth over 100 independent experiments. The column ``Sharpe ratio" reports the corresponding average Sharpe ratio ($\frac{\text{Mean} - 1}{\sqrt{\text{Variance}}}$).
		The numbers in the brackets are the standard deviations.}
	\begin{tabular}{ccccc}
		\toprule
		$\mu$    & $\sigma$& Mean  & Variance & Sharpe ratio \\
		\midrule
		-0.5  & 0.1   & 1.78 ( 0.0082 ) & 0.01 ( 3e-04 ) & 7.43 ( 0.04 ) \\
		-0.3  & 0.1   & 1.55 ( 0.0077 ) & 0.02 ( 0.00034 ) & 3.84 ( 0.027 ) \\
		-0.1  & 0.1   & 1.14 ( 0.022 ) & 0.01 ( 0.0039 ) & 1.24 ( 0.0073 ) \\
		0     & 0.1   & 1.01 ( 0.0055 ) & 0 ( 0.0016 ) & 0.12 ( 0.16 ) \\
		0.1   & 0.1   & 1.26 ( 0.052 ) & 0.09 ( 0.032 ) & 0.85 ( 0.012 ) \\
		0.3   & 0.1   & 1.83 ( 0.021 ) & 0.04 ( 0.0023 ) & 4.31 ( 0.056 ) \\
		0.5   & 0.1   & 1.92 ( 0.017 ) & 0.01 ( 0.00032 ) & 10.82 ( 0.15 ) \\
		-0.5  & 0.2   & 1.77 ( 0.015 ) & 0.04 ( 0.0023 ) & 3.65 ( 0.038 ) \\
		-0.3  & 0.2   & 1.54 ( 0.018 ) & 0.08 ( 0.0036 ) & 1.89 ( 0.025 ) \\
		-0.1  & 0.2   & 1.14 ( 0.043 ) & 0.05 ( 0.028 ) & 0.62 ( 0.006 ) \\
		0     & 0.2   & 1.01 ( 0.01 ) & 0.01 ( 0.015 ) & 0.04 ( 0.091 ) \\
		0.1   & 0.2   & 1.12 ( 0.072 ) & 0.12 ( 0.09 ) & 0.36 ( 0.19 ) \\
		0.3   & 0.2   & 1.79 ( 0.048 ) & 0.17 ( 0.019 ) & 1.94 ( 0.052 ) \\
		0.5   & 0.2   & 1.92 ( 0.033 ) & 0.04 ( 0.0032 ) & 4.8 ( 0.11 ) \\
		-0.5  & 0.3   & 1.76 ( 0.02 ) & 0.1 ( 0.0073 ) & 2.36 ( 0.035 ) \\
		-0.3  & 0.3   & 1.52 ( 0.035 ) & 0.18 ( 0.018 ) & 1.23 ( 0.021 ) \\
		-0.1  & 0.3   & 1.12 ( 0.061 ) & 0.11 ( 0.079 ) & 0.39 ( 0.14 ) \\
		0     & 0.3   & 1.01 ( 0.013 ) & 0.05 ( 0.051 ) & 0.02 ( 0.063 ) \\
		0.1   & 0.3   & 1.1 ( 0.12 ) & 2.27 ( 20 ) & 0.17 ( 0.21 ) \\
		0.3   & 0.3   & 1.44 ( 0.049 ) & 0.18 ( 0.032 ) & 1.05 ( 0.018 ) \\
		0.5   & 0.3   & 1.73 ( 0.018 ) & 0.12 ( 0.0072 ) & 2.1 ( 0.033 ) \\
		-0.5  & 0.4   & 1.74 ( 0.028 ) & 0.19 ( 0.016 ) & 1.7 ( 0.033 ) \\
		-0.3  & 0.4   & 1.48 ( 0.062 ) & 0.29 ( 0.057 ) & 0.9 ( 0.017 ) \\
		-0.1  & 0.4   & 1.11 ( 0.075 ) & 0.19 ( 0.15 ) & 0.25 ( 0.17 ) \\
		0     & 0.4   & 1.01 ( 0.016 ) & 0.11 ( 0.12 ) & 0.01 ( 0.048 ) \\
		0.1   & 0.4   & 1.04 ( 0.06 ) & 0.13 ( 0.13 ) & 0.08 ( 0.18 ) \\
		0.3   & 0.4   & 1.4 ( 0.08 ) & 0.28 ( 0.087 ) & 0.77 ( 0.015 ) \\
		0.5   & 0.4   & 1.71 ( 0.027 ) & 0.22 ( 0.015 ) & 1.52 ( 0.03 ) \\
		\bottomrule
	\end{tabular}%
	\label{tab:mv pg online}%
\end{table}%

\subsection{Ergodic Linear--Quadratic Control}
\label{sec:application lq ergodic}	
Consider the ergodic linear--quadratic (LQ) control problem where state responds to actions in a linear way
\begin{equation}
	\label{eq:lq dynamics}
	\dd X_t = (AX_t + Ba_t)\dd t + (CX_t + D a_t)\dd W_t,\ X_0 = x_0,
\end{equation}
and the goal is  to maximize the long term average payoff
\begin{equation}
	\label{eq:lq payoff}
	\liminf_{T\to \infty}\frac{1}{T}\E\left[\int_0^T r(X_t,a_t)\dd t | X_0 = x_0 \right],
\end{equation}
with $r(x,a) = -(\frac{M}{2}x^2 + Rxa + \frac{N}{2}a^2 + Px + Qa)$.

In  the entropy-regularized RL formulation, the policy is denoted by $\bm{\pi}(\cdot|x)$ and actions are generated from this policy. The corresponding goal is to maximize 
\begin{equation}
	\label{eq:lq payoff entropy}
	\liminf_{T\to \infty}\frac{1}{T}\E^{\p^W}\bigg[ \int_0^T \int_{\mathbb{R}} r(\tilde{X}_t^{\bm{\pi}},a)\bm{\pi}(a|\tilde{X}_t^{\bm{\pi}})\dd a \dd t + \gamma \mathcal{H}(\bm{\pi}\big(\cdot|\tilde{X}_t^{\bm{\pi}})\big) \dd t \Big| \tilde{X}_0^{\bm{\pi}} = x_0\bigg] ,
\end{equation}
where $\mathcal{H}$ is the differential entropy as before. Moreover, $\tilde{X}^{\bm{\pi}}$ satisfies
\begin{equation}
	\label{eq:lq dynamics randomized}
	\dd \tilde{X}_t^{\bm{\pi}} = \int_{\mathbb{R}} (A \tilde{X}_t^{\bm{\pi}} + B a)\bm{\pi}(a|\tilde{X}_t^{\bm{\pi}})\dd a \dd t  + \sqrt{\int_{\mathbb{R}}  \big( C{{}\tilde{X}^{\bm{\pi}}_t}^2 + D a\big)^2\bm{\pi}(a|\tilde{X}_t^{\bm{\pi}})\dd a} \dd W_t.
\end{equation}

Following the same line of deductions  as in \cite{wang2020reinforcement},
we can show that the optimal policy is a normal distribution whose mean is linear in the state and whose variance is a constant. Therefore we parameterize the policy by $\bm{\pi}^{\phi}(\cdot|x) = \mathcal{N}(\cdot|\phi_1 x + \phi_2, e^{\phi_3})$.
Moreover, the function $J$ is parameterized as a quadratic function $J^{\theta}(x) = \frac{1}{2}\theta_0 x^2 + \theta_1 x$ (we ignore the constant term since $J$  is  unique up to a constant) and the optimal value $V$ is an extra parameter.

This problem falls into the formulation of an ergodic task; so we directly implement Algorithm \ref{algo:ergodic incremental} in our simulation and then compare the learned parameters with the theoretically optimal ones. In addition, we compare the up-to-now average reward during the online learning process to two theoretical benchmarks. The first one is the omniscient optimal level, which is the maximum long term average reward that can be achieved by a hypothetical agent who knows completely about the environment  (i.e. the correct model and model parameters) and acts optimally (the optimal policy is a deterministic one) without needing to explore (and hence there is no entropy regularization). The second benchmark is the omniscient optimal level less the exploration cost, which is the maximum long term average reward that can be achieved by the aforementioned  hypothetical agent who  is however forced to explore under entropy regularization.\footnote{See Appendix B2 for precise definitions of these two benchmarks and detailed calculations of them.} 
Clearly, since exploration (rendering a stochastic policy) is inherent in the RL setting, our algorithm can at most achieve the second benchmark. In other words, after learning for a sufficiently long time, we can learn the correct optimal policy but can only expect  the up-to-now average reward  to approach the optimal level less the exploration cost.

To guarantee the stationarity of the controlled state process, we set $A = -1,B = C= 0$ and $D = 1$. 
Moreover, we set $x_0=0$,  $M = N=Q =2$, $R=P =1$, and  $\gamma = 0.1$. Learning rate is initialized as $\alpha_{\theta} = \alpha_{\phi} = 0.001$, and decays according to $l(t) = \frac{1}{\max\{1, \log t \}}$. All the parameters to be learned are initialized as 0 and time discretization is taken as $\Delta t = 0.01$. We repeat the experiment for 100 times.

\begin{figure}[H]
	\centering
	\begin{subfigure}{0.45\textwidth}
		\centering
		\includegraphics[width = 1\textwidth]{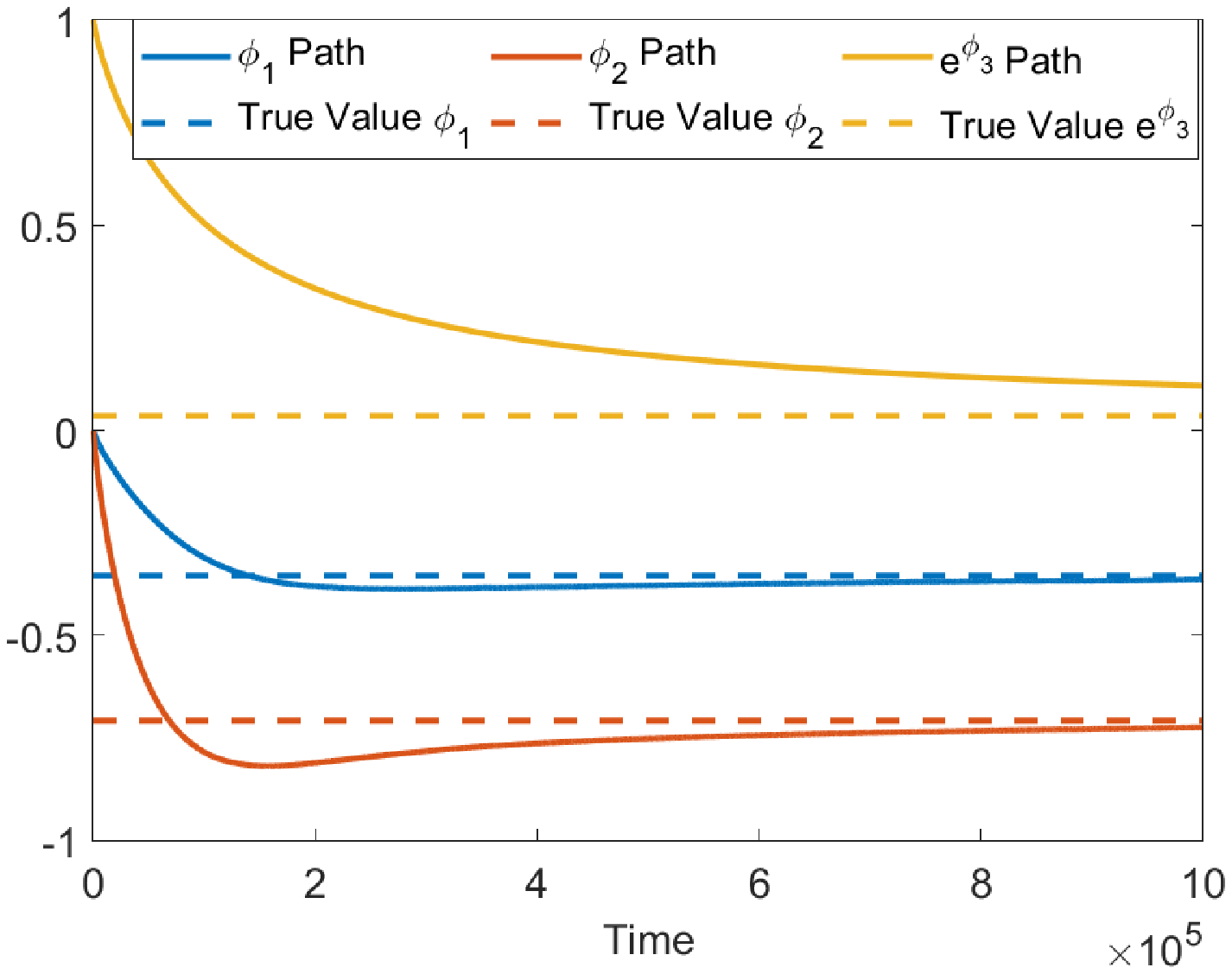}
		\caption{The learned parameters in the policy along one sample trajectory.}
	\end{subfigure}
	\begin{subfigure}{0.45\textwidth}
		\centering
		\includegraphics[width = 1\textwidth]{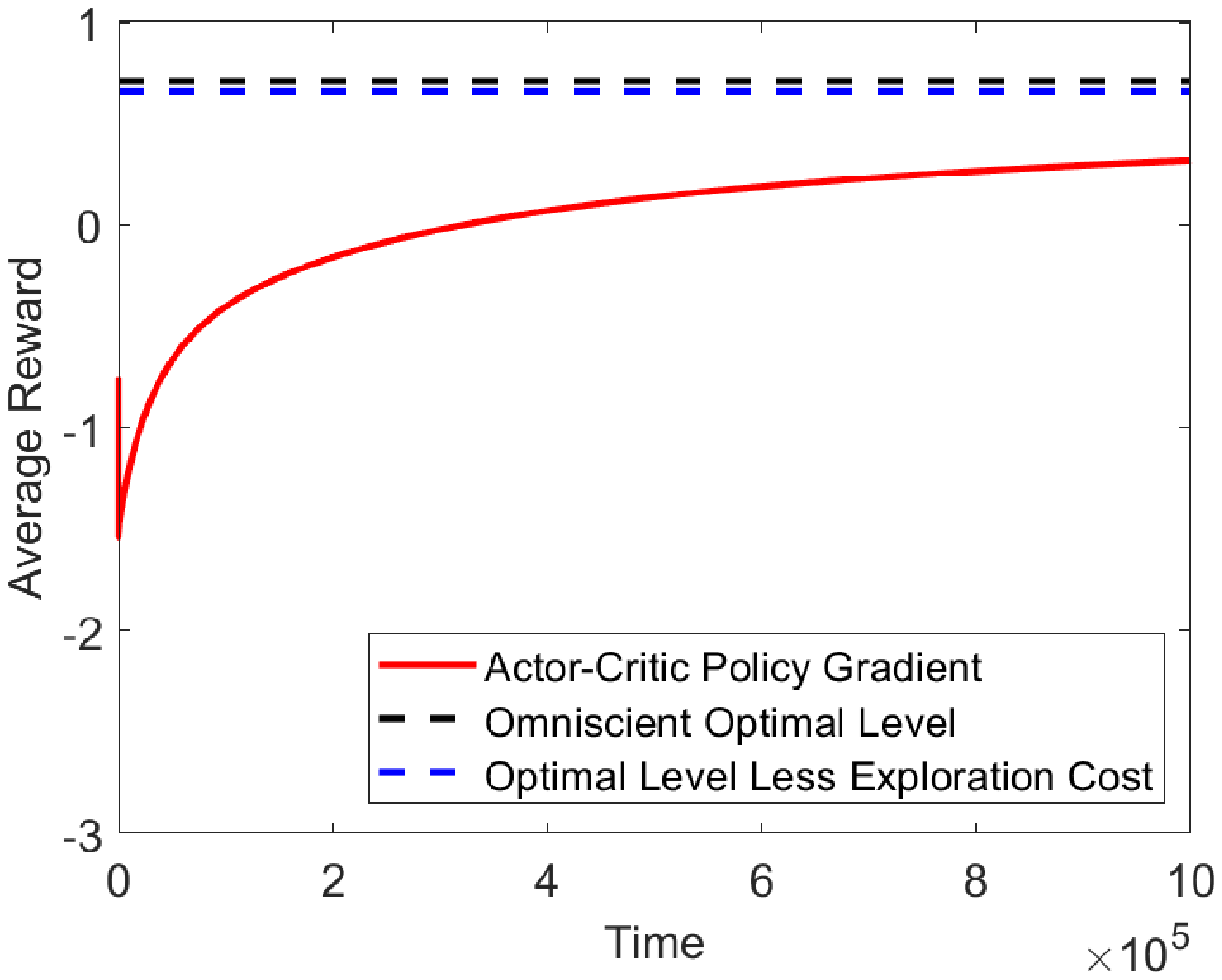}
		\caption{The average reward along one sample trajectory.}
	\end{subfigure}
	\caption{\textbf{Convergence of the learned policy and the average reward under the online learning algorithm.} A single state trajectory is generated with length $T = 10^6$ under the online AC algorithm. The left panel illustrates the convergence of the policy parameters, where the dashed horizontal lines indicate the values of the respective parameters of the theoretically optimal policy to the entropy-regularized exploratory stochastic control problem. The right panel shows the convergence of the average reward, where the two dashed horizontal lines are respectively the omniscient optimal average reward without exploration when the model parameters are known, and the omniscient optimal average reward less the exploration cost. We repeat the experiment for 100 times to calculate the standard deviations of the predicted parameters, which are represented as the shaded areas. The width of each shaded area is twice the corresponding standard deviation, which is very small compared to the scale of the vertical axis.}
	\label{fig:ergodic lq}
\end{figure}

We implement $TD(0)$ for both the PE and the PG parts of the AC algorithm, referred to as the Actor--Critic Policy Gradient algorithm in Figure \ref{fig:ergodic lq}. Namely, we choose test functions $\xi_t = \frac{\partial J^{\theta}}{\partial \theta}(X_t), \eta_t = 1, \zeta_t = 0$ in Algorithm \ref{algo:ergodic incremental}. Figure \ref{fig:ergodic lq} shows  the convergence of the learned policy parameters along with that of the average reward along a single state sample  trajectory.  Observe that the average reward  first decreases at the beginning of this particular trajectory. The reason may have been that during the initial iterations the underlying state process has not yet converged to the stationary distribution and the initial policies are still far away from the optimal one, and hence the average reward is dominated by a few ``wrong trials''. After a sufficient amount of time, however, both the policies and the average reward start to converge to the theoretically optimal values. Between the two it takes a much longer time for the average reward to approach the optimal level as we wait for  the contribution from the bad performance of the beginning  period to diminish. 

\section{Conclusion}
\label{sec:conclusion}

This paper is the final installment of a ``trilogy", the first two being  \cite{wang2020reinforcement} and \cite{jia2021policy}, that endeavors to develop a
systematic and unified theoretical foundation for RL in continuous time with continuous state space and possibly continuous action space. The previous two papers
address exploration and PE, respectively, and this paper focuses on PG. A major finding of the current paper is that PG is intimately related to PE, and thus the martingale characterization of PE established in \cite{jia2021policy} can be applied to PG. Combining the theoretical results of the three papers, we propose online and offline actor--critic algorithms for general model-free RL tasks, where we learn value functions and stochastic policies simultaneously and alternatingly.

This series of papers are characterized by conducting all the theoretical analysis within the continuous setting and discretizing time only when implementing the algorithms. The advantages of this approach, versus discretizing time right at the start and then applying existing MDP results, are articulated in \cite{doya2000reinforcement}. Moreover, more analytical tools are at our disposal in the continuous setting, including calculus, stochastic calculus, stochastic control, and differential equations. The discrete-time versions of the various algorithms devised in the three papers are indeed well known in the discrete-time RL literature; hence their convergence is well established. On the other hand, \cite{jia2021policy} prove
that any convergent time-discretized PE algorithm also converges as the mesh size goes to zero. Because PG algorithms developed in the current paper are essentially derived from the martingality for PE, the same convergence
also holds for them.


It is interesting to note that the derivation and representation of PG are not entirely analogous to that of MDPs. For example, the latter involves a state--action function (Q-function), whereas the former is essentially the expected integration of a term involving the value function. 

The study on continuous-time RL is still in its infancy, and open questions abound. These include, to name but a few, regret bound of episodic RL problems in terms of the number of episodes, interpretation of Q-function and Q-learning in the continuous setting, and dependence of the performance of AC algorithms on the temperature parameter when there is an exploration regularizer.

\section*{Acknowledgement}
Zhou gratefully acknowledges financial support through a start-up grant and the Nie Center for Intelligent Asset Management at Columbia University. 


\newpage

\appendix
\section*{Appendix A. Connections with Policy Gradient in Discrete Time}
We review the classical policy gradient approach and results for discrete-time Markov decision processes (MDPs) here and compare them with their continuous-time counterparts developed in the main text.

For simplicity, we consider a time-homogeneous MDP $X=\{X_t,t=0,1,2,\cdots\}$ with a state space $\mathcal{X}$, an action space $\mathcal{A}$, and a transition matrix $\p(X_1 = x'|X_0 = x, a_0 = a) =  p(x'|x,a)$. Both $\mathcal{X}$ and $\mathcal{A}$ are finite sets. The expected reward is $r(x,a)$ with a discount factor $\beta\in (0,1)$. The agent's total expected reward is $\E\left[\sum_{t=0}^{\infty}\beta^t r(X_t,a_t) \right]$. A (stochastic) policy is denoted by $\bm\pi^{\phi}(\cdot|x)\in \mathcal{P}(\mathcal{A})$, which is a probability density function on  $\mathcal{A}$, with a suitable parameter vector $\phi\in \mathbb{R}^{L_{\phi}}$.

Define the value function  associated with a given policy $\bm\pi^{\phi}$ by
\begin{equation}
\label{eq:value mdp}
\begin{aligned}
J(x;\bm\pi^{\phi}) = & \E\left[\sum_{t=0}^{\infty}\beta^t r(X_t^{\bm\pi^{\phi}},a^{\bm\pi^{\phi}}_t)\Big| X_0^{\bm\pi^{\phi}} = x \right] \\
= & \E\left[\sum_{t=0}^{\infty}\beta^t \E_{a_t\sim \bm\pi^{\phi}(\cdot|X_t^{\bm\pi^{\phi}})}\left[  r(X_t^{\bm\pi^{\phi}},a_t)  \right]\Big| X_0^{\bm\pi^{\phi}} = x\right].
\end{aligned}
\end{equation}

We are interested in the gradient of the value function with respect to the policy parameter $\phi$, that is, $\frac{\partial J(x; \bm\pi^{\phi})}{\partial \phi}$.
The classical policy gradient theorem (e.g., \citealt[Theorem 1]{sutton2000policy}) states that
\begin{equation}
\label{eq:pg classical}
\frac{\partial J(x; \bm\pi^{\phi})}{\partial \phi} = \sum_{x'\in \mathcal{X}}\mu^{\bm\pi^{\phi}}(x')  \sum_{a\in \mathcal{A}}\fracpartial{\bm\pi^{\phi}}{\phi}(a|x')Q(x',a;\bm\pi^{\phi}) ,
\end{equation}
where $Q(x,a;\bm\pi^{\phi}) = r(x,a) + \E[\sum_{t=1}^{\infty}\beta^t r(X_t^{\bm\pi^{\phi}},a^{\bm\pi^{\phi}}_t)\Big| X_0^{\bm\pi^{\phi}} = x]$ is the Q-function, and $\mu^{\bm\pi^{\phi}}(x') = \sum_{t=0}^{\infty}\beta^t\p(X_t^{\bm\pi^{\phi}} = x'|X_0^{\bm\pi^{\phi}} = x)$ is the (discounted) occupation time.

Define $\ell(x') = \sum_{a\in \mathcal{A}}\fracpartial{\bm\pi^{\phi}}{\phi}(a|x')Q(x',a;\bm\pi^{\phi})$, which  is a deterministic function of $x'$. Since $\sum_{a\in \mathcal{A}}\fracpartial{\bm\pi^{\phi}}{\phi}(a|x') = \frac{\partial }{\partial \phi}\sum_{a\in \mathcal{A}}\bm\pi^{\phi}(a|x') = 0$,  $\ell(x')$ can be equivalently written as
\[\begin{aligned}
\ell(x') = & \sum_{a\in \mathcal{A}}\fracpartial{\bm\pi^{\phi}}{\phi}(a|x')\left[Q(x',a;\bm\pi^{\phi}) - B(x')\right] \\
= & \sum_{a\in \mathcal{A}}\fracpartial{\log\bm\pi^{\phi}}{\phi}(a|x')\left[Q(x',a;\bm\pi^{\phi}) - B(x')\right]\bm\pi^{\phi}(a|x') \\
= & \E_{a\sim \bm\pi^{\phi}(\cdot|x')}\left[  \fracpartial{\log\bm\pi^{\phi}}{\phi}(a|x')\left[Q(x',a;\bm\pi^{\phi}) - B(x')\right] \right],
\end{aligned}  \]
for any function $B(\cdot)$, sometimes known as a baseline \citep{williams1992simple}.

On the other hand,
\[ \mu^{\bm\pi^{\phi}}(x') = \sum_{t=0}^{\infty}\beta^t\p(X_t^{\bm\pi^{\phi}} = x'|X_0^{\bm\pi^{\phi}} = x) = \E\left[ \sum_{t=0}^{\infty} \beta^t \one_{\{X_t^{\bm\pi^{\phi}}  = x'\}} \Big| X_0^{\bm\pi^{\phi}} = x \right] .\]
Therefore, \eqref{eq:pg classical} is equivalent to
\begin{equation}
\label{eq:pg classical derive}
\begin{aligned}
\frac{\partial J(x; \bm\pi^{\phi})}{\partial \phi} = & \sum_{x'\in \mathcal{X}} \E\left[ \sum_{t=0}^{\infty} \beta^t \one_{\{X_t^{\bm\pi^{\phi}}  = x'\}} \Big| X_0^{\bm\pi^{\phi}} = x \right] \ell(x') \\
= & \sum_{t=0}^{\infty} \beta^t \sum_{x'\in \mathcal{X}}  \E\left[ \one_{\{X_t^{\bm\pi^{\phi}}  = x'\}} \ell(x') \Big| X_0^{\bm\pi^{\phi}} = x \right] \\
= & \sum_{t=0}^{\infty} \beta^t  \E\left[\sum_{x'\in \mathcal{X}}  \one_{\{X_t^{\bm\pi^{\phi}}  = x'\}} \ell(X_t^{\bm\pi^{\phi}}) \Big| X_0^{\bm\pi^{\phi}} = x \right] \\
=&  \E\left[\sum_{t=0}^{\infty} \beta^t   \ell(X_t^{\bm\pi^{\phi}}) \Big| X_0^{\bm\pi^{\phi}} = x \right]  \\
= & \E\left[\sum_{t=0}^{\infty} \beta^t  \E_{a\sim \bm\pi^{\phi}(\cdot|X_t^{\bm\pi^{\phi}})}\left[  \fracpartial{\log\bm\pi^{\phi}}{\phi}(a|X_t^{\bm\pi^{\phi}})\left[Q(X_t^{\bm\pi^{\phi}},a;\bm\pi^{\phi}) - B(X_t^{\bm\pi^{\phi}})\right] \right]
\Big| X_0^{\bm\pi^{\phi}} = x \right]\\
= & \E\left[\sum_{t=0}^{\infty} \beta^t   \fracpartial{\log\bm\pi^{\phi}}{\phi}(a_t^{\bm\pi^{\phi}}|X_t^{\bm\pi^{\phi}})\left[Q(X_t^{\bm\pi^{\phi}},a_t^{\bm\pi^{\phi}};\bm\pi^{\phi}) - B(X_t^{\bm\pi^{\phi}})\right] \Big| X_0^{\bm\pi^{\phi}} = x \right]\\
= & \E\left[\sum_{t=0}^{\infty} \beta^t   \fracpartial{\log\bm\pi^{\phi}}{\phi}(a_t^{\bm\pi^{\phi}}|X_t^{\bm\pi^{\phi}})\left[Q(X_t^{\bm\pi^{\phi}},a_t^{\bm\pi^{\phi}};\bm\pi^{\phi}) - J(X_t^{\bm\pi^{\phi}};\bm\pi^{\phi})\right] \Big| X_0^{\bm\pi^{\phi}} = x \right].
\end{aligned}
\end{equation}

If we choose the baseline to be the value function $B(\cdot) = J(\cdot;\bm\pi^{\phi})$, then  \eqref{eq:pg classical derive} gives the representation of policy gradient in the advantage actor--critic approach \citep{mnih2016asynchronous}.

If we are to extend the above derivation to the continuous-time setting, then an essential question is what the Q-function should be in continuous time. This question has been extensively studied in a recent paper \cite{jia2022q} from which we realize that (using the notations in this paper with the discount factor  $e^{-\beta t}$)
\[\begin{aligned}
& Q(X_t^{\bm\pi^{\phi}},a_t^{\bm\pi^{\phi}};\bm\pi^{\phi}) - J(X_t^{\bm\pi^{\phi}};\bm\pi^{\phi})  \\
\approx & \mathcal{L}^{a_t^{\bm\pi^{\phi}}} J(X_t^{\bm\pi^{\phi}};\bm\pi^{\phi}) \Delta t + r(X_t^{\bm\pi^{\phi}},a_t^{\bm\pi^{\phi}})\Delta t  - \beta J(X_t^{\bm\pi^{\phi}};\bm\pi^{\phi}) \Delta t \\
\approx & \dd J(X_t^{\bm\pi^{\phi}};\bm\pi^{\phi})+ r(X_t^{\bm\pi^{\phi}},a_t^{\bm\pi^{\phi}})\dd t  - \beta J(X_t^{\bm\pi^{\phi}};\bm\pi^{\phi}) \dd t + \fracpartial{J}{x}(X_t^{\bm\pi^{\phi}};\bm\pi^{\phi})^\top \sigma(X_t^{\bm\pi^{\phi}}) \dd W_t .
\end{aligned}  \]
Therefore, \eqref{eq:pg classical derive} becomes
\[\begin{aligned}
\frac{\partial J(x; \bm\pi^{\phi})}{\partial \phi} = \E\bigg[\int_{0}^{\infty} & e^{-\beta t}  \fracpartial{\log\bm\pi^{\phi}}{\phi}(a_t^{\bm\pi^{\phi}}|X_t^{\bm\pi^{\phi}})\big[  \dd J(X_t^{\bm\pi^{\phi}};\bm\pi^{\phi})+ r(X_t^{\bm\pi^{\phi}},a_t^{\bm\pi^{\phi}})\dd t  \\
&- \beta J(X_t^{\bm\pi^{\phi}};\bm\pi^{\phi}) \dd t + \fracpartial{J}{x}(X_t^{\bm\pi^{\phi}};\bm\pi^{\phi})^\top \sigma(X_t^{\bm\pi^{\phi}}) \dd W_t \big] \Big| X_0^{\bm\pi^{\phi}} = x \bigg] .
\end{aligned}\]

Note that  there is no policy regularizer in the current discussion; so the above coincides with the expression of the policy gradient \eqref{eq:policy gradient expectation} with $\gamma = 0$.
\section*{Appendix B. Theoretical Results Employed in Simulation Experiments}
For the reader's convenience, we summarize the theoretical results employed in the two simulation studies in Section \ref{sec:applications}. Their proofs are   similar to those of the analogous results in \cite{wang2020continuous} and \cite{wang2020reinforcement} respectively.
\subsection*{Appendix B1. Mean-Variance Portfolio Selection}
Let the true model be given as \eqref{eq:model stock} and one aims to solve \eqref{eq:mv}. An omniscient agent's optimal policy is a deterministic one, given by $a_t^* = -\frac{\mu-r}{\sigma^2}(x_t - w^*)$, where $w^* = \frac{z\exp\{ \frac{(\mu-r)^2}{\sigma^2} T\} - x_0}{\exp\{ \frac{(\mu-r)^2}{\sigma^2} T\} - 1}$. Given this policy, the discounted wealth process \eqref{eq:wealth process} becomes
\[ \dd x_t^* = -\frac{(\mu-r)^2}{\sigma^2}(x_t^*-w^*)\dd t - \frac{\mu-r}{\sigma}(x_t^*-w^*) \dd W_t . \]
Hence $x_t^*-w$ is a geometric Brownian motion. We can compute $\E[x_T^*] = z$, and
\[\operatorname{Var}(x_T^*) = (x_0 - w^*)^2 \exp\{-2\frac{(\mu-r)^2}{\sigma^2} T  \}\left(\exp\{\frac{(\mu-r)^2}{\sigma^2} T  \} - 1\right) = \frac{(x_0 - z)^2}{\left( \exp\{\frac{(\mu-r)^2}{\sigma^2} T  \} - 1 \right)^2} . \]

If the agent knows the true model but is forced to take a stochastic policy subject to the entropy regularizer, then the optimal policy is
\[ \bm\pi^*(\cdot|t,x) \sim \mathcal{N}\left( -\frac{\mu-r}{\sigma^2}(x - w^*), \frac{\gamma}{2\sigma^2}\exp\{ \frac{(\mu-r)^2}{\sigma^2} (T-t)\} \right).\]
Under this policy, the dynamics of $\tilde{X}^{\bm\pi^*}_t$ is
\[ \dd \tilde{X}^{\bm\pi^*}_t = -\frac{(\mu-r)^2}{\sigma^2}(\tilde{X}^{\bm\pi^*}_t-w^*)\dd t + \sqrt{\frac{(\mu-r)^2}{\sigma^2}(\tilde{X}^{\bm\pi^*}_t-w^*)^2  + \frac{\gamma}{2}\exp\{ \frac{(\mu-r)^2}{\sigma^2}(T-t) \}}\dd W_t . \]
With the same value of $w^*$, one can show that  $\E^{\p^W}[\tilde{X}_T^{\bm\pi^*}] = z$, and
\[ \operatorname{Var}^{\p^W}( \tilde{X}_T^{\bm\pi^*} ) = \exp\{ \frac{-(\mu-r)^2}{\sigma^2} T \}(x_0-w^*)^1 + \frac{\gamma}{2}T - (z-w^*)^2 = \frac{(x_0 - z)^2}{\left( \exp\{\frac{(\mu-r)^2}{\sigma^2} T  \} - 1 \right)^2}  + \frac{\gamma}{2}T. \]

\subsection*{Appendix B2. Ergodic Linear-Quadratic Control}
Two benchmarks, ``omniscient optimal level" and ``omniscient optimal level less  exploration cost", are used in Section \ref{sec:application lq ergodic} for comparison. We introduce their formal definitions here.

\begin{definition}
The omniscient optimal level is the maximum value of \eqref{eq:lq payoff} subject to \eqref{eq:lq dynamics} when all the model coefficients $(A,B,C,D,M,R,N,P,Q)$ are known to the agent. The omniscient optimal level less  exploration cost is defined as $$\liminf_{T\to \infty}\frac{1}{T}\E^{\p^W}\bigg[ \int_0^T \int_{\mathbb{R}} r(\tilde{X}_t^{\bm{\pi}^*},a)\bm{\pi}(a|\tilde{X}_t^{\bm{\pi}^*})\dd a \dd t \Big| \tilde{X}_0^{\bm{\pi}^*} = x_0\bigg], $$ where $\bm{\pi}^*$ is the optimal policy to \eqref{eq:lq payoff entropy} with the entropy regularizer and subject to \eqref{eq:lq dynamics randomized} when all the model coefficients are known to the agent.
\end{definition}

We compute these two values using the Hamilton-Jacobi-Bellman (HJB) equation approach.

Let the true model be given by \eqref{eq:lq dynamics} and one aims to maximize the long-term average reward \eqref{eq:lq payoff}.
Consider the associated HJB equation:
\[\begin{aligned}
0 = & \sup_{a}[ \mathcal{L}^a \varphi(x) + r(x,a) - V] \\
=&  \sup_{a}\left[(Ax + Ba)\varphi'(x) + \frac{1}{2}(Cx + Da)^2\varphi''(x) - (\frac{M}{2}x^2 + Rxa + \frac{N}{2}a^2 + Px + Qa)  -V \right] . 	
\end{aligned} \]
Conjecturing  $\varphi(x) = \frac{1}{2}k_2 x^2 + k_1 x$ and plugging it to the HJB equation, we get the first-order condition $a^* = \frac{[k_2(B+CD)-R]x + k_1B - Q}{N -k_2 D^2}$, assuming $N - k_2D^2 > 0$. The HJB equation now becomes
\[ 0 = \frac{1}{2}[k_2(2A+C^2) - M]x^2 + (k_1A - P)x - V + \frac{1}{2} \frac{\left\{ [k_2(B+CD)-R]x + k_1B - Q \right\}^2}{N - k_2D^2} .\]
This leads to three algebraic equations by matching the coefficients of $x^2$, $x$ and the constant term:
\begin{equation}
\label{eq:lq algebraic equations}
\left\{ \begin{aligned}
& k_2(2A+C^2) - M + \frac{[k_2(B+CD)-R]^2}{N-k_2 D^2}  = 0,\\
& k_1A - P + \frac{[k_2(B+CD)-R](k_1B - Q)}{N-k_2D^2} = 0,\\
& V = \frac{(k_1 B-Q)^2}{2(N - k_2D^2)}.
\end{aligned} \right.  
\end{equation}
Note that \eqref{eq:lq algebraic equations} coincides with the system of equations in footnote 12 and Theorem 9 in \citet{wang2020reinforcement} when the discount factor is 0.

Solving these algebraic equation gives the omniscient optimal reward and the corresponding optimal policy $a^* = \frac{[k_2(B+CD)-R]x + k_1B - Q}{N -k_2 D^2}$.

If the agent knows the true model but still adopts stochastic policies with a entropy regularizer, then the optimal policy is given by
\[ \bm\pi^*(\cdot|x)\sim \mathcal{N}\left(\frac{[k_2(B+CD)-R]x + k_1B - Q}{N -k_2 D^2}  , \frac{\gamma}{N - k_2D^2}\right) , \]
where $k_2,k_1$ are determined by \eqref{eq:lq algebraic equations}.
This optimal solution is identical to that in \citet[Theorem 4]{wang2020reinforcement} when the discount factor is 0.

Under this stochastic policy, the state dynamics become
\[\begin{aligned}
\dd \tilde{X}_t^{\bm{\pi}^*} = & \left( A \tilde{X}_t^{\bm{\pi}^*} + B \frac{[k_2(B+CD)-R]\tilde{X}_t^{\bm{\pi}^*} + k_1B - Q}{N -k_2 D^2} \right)\dd t  \\
& + \Bigg( C^2\tilde{X}_t^{{\bm{\pi}^*}^2} + CD\tilde{X}^{\bm{\pi}^*}_t \frac{[k_2(B+CD)-R]\tilde{X}_t^{\bm{\pi}^*} + k_1B - Q}{N -k_2 D^2} \\
& +  D^2 \Big[\big( \frac{[k_2(B+CD)-R]\tilde{X}_t^{\bm{\pi}^*} + k_1B - Q}{N -k_2 D^2}\big)^2 + \frac{\gamma}{N - k_2D^2}\Big] \Bigg)^{1/2}\dd W_t.
\end{aligned} \]

To calculate the long-term average value
$\tilde V=\liminf_{T\to \infty}\frac{1}{T}\E^{\p^W}\bigg[ \int_0^T \int_{\mathbb{R}} r(\tilde{X}_t^{\bm{\pi}^*},a)\bm{\pi}^*(a|\tilde{X}_t^{\bm{\pi}^*})\dd a \dd t  \bigg] $, consider the corresponding HJB equation
\[\int_{\mathbb{R}} [ (Ax + Ba)\tilde{\varphi}'(x) + \frac{1}{2}(Cx + Da)^2 \tilde{\varphi}''(x) - (\frac{M}{2}x^2 + Rxa + \frac{N}{2}a^2 + Px + Qa)  ] \bm\pi^*(a|x)\dd a - \tilde{V}  = 0.  \]
Starting with an ansatz $\tilde{\varphi}(x) = \frac{1}{2}\tilde{k}_2 x^2 + \tilde{k}_1 x$ and going through the same calculations as above  we obtain three equations
\[\begin{aligned}
& (A + B\frac{k_2(B+CD)-R}{N-k_2D^2}) \tilde{k}_2 + \frac{1}{2} \bigg( C^2 + 2CD\frac{k_2(B+CD)-R}{N-k_2D^2} + D^2 \frac{[k_2(B+CD)-R]^2}{(N-k_2D^2)^2} \bigg)\tilde{k}_2 \\
& - \frac{M}{2} - R\frac{k_2(B+CD)-R}{N-k_2D^2} - \frac{N}{2}\frac{[k_2(B+CD)-R]^2}{(N-k_2D^2)^2} = 0, \\
& (A + B\frac{k_2(B+CD)-R}{N-k_2D^2}) \tilde{k}_1 + B\frac{k_1B - Q}{N-k_2D^2}\tilde{k}_2 \\
& + \frac{1}{2}\Bigg(  2CD\frac{k_1B-Q}{N-k_2D^2} + 2D^2\frac{[k_2(B+CD)-R](k_1B-Q)}{(N-k_2D^2)^2} \Bigg)\tilde{k}_2\\
& -R \frac{k_1B-Q}{N-k_2D^2} - N\frac{[k_2(B+CD)-R](k_1B-Q)}{(N-k_2D^2)^2} -P - \frac{Q[k_2(B+CD)-R]}{N-k_2D^2} = 0, \\
& \tilde{k}_1 B \frac{k_1B-Q}{N-k_2D^2} + \frac{1}{2}\Bigg( D^2 \frac{(k_1B-Q)^2}{(N-k_2D^2)^2} + D^2\frac{\gamma}{N-k_2D^2}  \Bigg)\tilde{k}_2 \\
& - \frac{N}{2}\frac{(k_1B-Q)^2}{(N-k_2D^2)^2} - \frac{N}{2}\frac{\gamma}{N-k_2D^2} - Q\frac{k_1B-Q}{N-k_2D^2} = \tilde{V}.
\end{aligned}\]
The solutions to the above equations are $\tilde{k}_2 = k_2$, $\tilde{k}_1 = k_1$. Hence
\[\begin{aligned}
\tilde{V} = & k_1 B\frac{k_1B-Q}{N-k_2D^2} + \frac{k_2D^2}{2}\Bigg(  \frac{(k_1B-Q)^2}{(N-k_2D^2)^2} + \frac{\gamma}{N-k_2D^2}  \Bigg) - \frac{N}{2}\Bigg(\frac{(k_1B-Q)^2}{(N-k_2D^2)^2}+\frac{\gamma}{N-k_2D^2}\Bigg) \\
& - Q\frac{k_1B-Q}{N-k_2D^2} \\
=& \frac{(k_1B-Q)^2}{2(N-k_2D^2)} - \frac{\gamma}{2} = V - \frac{\gamma}{2}  .
\end{aligned} \]
By definition, $\tilde{V}$ is also the omniscient optimal level less  exploration cost. The difference, $\frac{\gamma}{2}$,  between $\tilde{V}$ and $V$ is hence the exploration cost due to randomization. Note that a parallel result when there is a discount factor is Theorem 10 in \citet{wang2020reinforcement}.
\section*{Appendix C. Proofs of Statements}
In all the proofs we use generic notations $C_1,C_2,\cdots$  to denote constants that are independent of  other variables involved such as $t,x,a$. A same such  notation may show up in different  places but  does  not necessarily have the same value.

\subsection*{Proof of Lemma \ref{lemma:relaxed sde solution}}

We start by examining $\tilde{b}\big(t,x,\bm{\pi}(\cdot|t,x)\big)$. Note that
\[\begin{aligned}
& \tilde{b}\big(t,x,\bm{\pi}(\cdot|t,x)\big) - \tilde{b}\big(t,x',\bm{\pi}(\cdot|t,x')\big)\\
= & \int_{\mathcal{A}} [b(t,x,a)-b(t,0,a)][\bm{\pi}(a|t,x) - \bm{\pi}(a|t,x')]\dd a + \int_{\mathcal{A}} b(t,0,a)[\bm{\pi}(a|t,x) - \bm{\pi}(a|t,x')]\dd a\\
&  + \int_{\mathcal{A}} [b(t,x,a) - b(t,x',a)]\bm{\pi}(a|t,x')\dd a .
\end{aligned}\]
Hence
\[\begin{aligned}
& \Big| \tilde{b}\big(t,x,\bm{\pi}(\cdot|t,x)\big) - \tilde{b}\big(t,x',\bm{\pi}(\cdot|t,x')\big)  \Big|\\
\leq & \int_{\mathcal{A}} |b(t,x,a)-b(t,0,a)||\bm{\pi}(a|t,x) - \bm{\pi}(a|t,x')|\dd a + \int_{\mathcal{A}} |b(t,0,a)||\bm{\pi}(a|t,x) - \bm{\pi}(a|t,x')|\dd a\\
&  + \int_{\mathcal{A}} |b(t,x,a) - b(t,x',a)|\bm{\pi}(a|t,x')\dd a \\
\leq & (C_1|x| + C_2)\int_{\mathcal{A}}|\bm{\pi}(a|t,x) - \bm{\pi}(a|t,x')|\dd a + C_1|x-x'| \\
\leq & C_1|x-x'| + (C_1|x| + C_2)C_3|x-x'| .
\end{aligned}\]
Moreover, note that
\[ |\tilde{b}\big(t,x,\bm{\pi}(\cdot|t,x)\big)| \leq \int_{\mathcal{A}} |b(t,x,a)|\bm{\pi}(a|t,x)\dd a \leq \int_{\mathcal{A}} (C_1|x|+C_2)\bm{\pi}(a|t,x) \dd a = C_1|x|+C_2.\]

Similarly, we can show that $\tilde{\sigma}\big(t,x,\bm{\pi}(\cdot|t,x)\big)$ is locally Lipschitz continuous and has linear growth in $x$. The unique existence of the strong solution to \eqref{eq:model relaxed} then follows from the standard SDE theory.

%

Next, the SDE \eqref{eq:model relaxed} yields
\[ \tilde{X}_s^{\bm{\pi}} = x + \int_t^s \tilde{b}\big( \tau,\tilde{X}_{\tau}^{\bm{\pi}}, \bm{\pi}(\cdot|\tau, \tilde{X}_{\tau}^{\bm{\pi}})\big) \dd \tau +  \int_t^s\tilde{\sigma}\big( \tau,\tilde{X}_{\tau}^{\bm{\pi}}, \bm{\pi}(\cdot|\tau, \tilde{X}_{\tau}^{\bm{\pi}})\big) \dd W_{\tau} .\]
Based on the proved growth condition on $\tilde{b},\tilde{\sigma}$, Cauchy–Schwarz inequality, and Burkholder-Davis-Gundy inequalities, we obtain
\[\begin{aligned}
& \E\bigg[\max_{t\leq s \leq T'}|\tilde{X}_s^{\bm{\pi}}|^{\mu}\Big|\tilde{X}_t^{\bm{\pi}}  = x\bigg] \\
\leq & C_1 \E\bigg[ |x|^{\mu} +  \max_{t\leq s \leq T'}\Big| \int_t^s \tilde{b}\big( \tau,\tilde{X}_{\tau}^{\bm{\pi}}, \bm{\pi}(\cdot|\tau, \tilde{X}_{\tau}^{\bm{\pi}})\big) \dd \tau\Big|^{\mu} \\
& + \max_{t\leq s \leq T'}\Big| \int_t^s \tilde{\sigma}\big( \tau,\tilde{X}_{\tau}^{\bm{\pi}}, \bm{\pi}(\cdot|\tau, \tilde{X}_{\tau}^{\bm{\pi}})\big) \dd W_{\tau}\Big|^{\mu} \Big| \tilde{X}_t^{\bm{\pi}}  = x \bigg]\\
\leq & C_1\E\bigg[ |x|^{\mu} + C_2 \int_t^{T'} (1 + \max_{t\leq s \leq \tau}|\tilde{X}_{s}^{\bm{\pi}}|)^{\mu} \dd \tau +C_3 \big(\int_t^{T'} (1 + \max_{t\leq s \leq \tau}|\tilde{X}_{s}^{\bm{\pi}}|)^2\dd \tau \big)^{\mu/2}\Big|\tilde{X}_t^{\bm{\pi}}  = x \bigg] \\
\leq & C_4(1 + |x|^{\mu}) + C_5 \int_t^{T'} \E\bigg[ \max_{t\leq s \leq \tau}|\tilde{X}_{s}^{\bm{\pi}}|^{\mu} \Big|\tilde{X}_t^{\bm{\pi}}  = x\bigg] \dd \tau .
\end{aligned}\]
Applying Gronwall's inequality to $\E\bigg[\max_{t\leq s \leq T'}|\tilde{X}_s^{\bm{\pi}}|^{\mu}\Big|\tilde{X}_t^{\bm{\pi}}  = x\bigg]$ as a function of $T'$, we obtain the second desired result of the lemma. The final result is evident.

\subsection*{Proof of Lemma \ref{lemma:f-k pde}}
Set $\tilde{v}(t,x) = e^{-\beta t} v(t,x)$. Then $\tilde{v}(T,x) = e^{-\beta T}h(x)$,  and \eqref{eq:pde characterization} implies
\begin{equation}
\label{eq:pde characterization transformed}
\int_{\mathcal{A}} \big[ \mathcal{L}^{a} \tilde{v}(t,x) + e^{-\beta t} r\big(t,x,a\big) + \gamma e^{-\beta t} p\big(t,x,a,\bm{\pi}(\cdot|t,x) \big)  \big] \bm{\pi}(a|t,x)\dd a = 0 .
\end{equation}
Similarly,  consider $\tilde{J}(t,x;\bm\pi) = e^{-\beta t} J(t,x;\bm{\pi})$. Then \eqref{eq:objective relaxed} yields
\[
\begin{aligned}
\tilde{J}(t,x;\bm\pi) = & \E^{\p^W}\bigg[\int_t^T \int_{\mathcal{A}} [ e^{-\beta s} r(s,\tilde{X}_s^{\bm{\pi}},a) + \gamma e^{-\beta s} p\big(s,\tilde{X}_s^{\bm{\pi}},a,\bm{\pi}(\cdot|s,\tilde{X}_s^{\bm{\pi}}) \big) ]\bm{\pi}(a|s,\tilde{X}_{s}^{\bm{\pi}})\dd a \dd s\\
& + e^{-\beta T}h(\tilde{X}_T^{\bm{\pi}})\Big|\tilde{X}_t^{\bm{\pi}} = x \bigg].
\end{aligned}
\]
So it suffices to prove the (viscosity) solution to \eqref{eq:pde characterization transformed}, $\tilde{v}$, coincides with $\tilde{J}(\cdot,\cdot;\bm\pi)$.
The proof now follows from applying \citet[Corollary 3.3]{beck2021nonlinear} to the SDE \eqref{eq:model relaxed}: under Assumption \ref{ass:dynamic} along with Definition  \ref{ass:admissible}, Lemma \ref{lemma:relaxed sde solution} verifies the sufficient conditions in \citet{beck2021nonlinear}.
\subsection*{Proof of Theorem \ref{prop:pe martingale}}

Using  the same discounting transformation as in the proof of Lemma \ref{lemma:f-k pde}, the first statement of Theorem \ref{prop:pe martingale} follows directly from \citet[Proposition 1]{jia2021policy} along with the Markov property of the solution to the SDE \eqref{eq:model relaxed}.

For the second statement, according to \citet[Proposition 4]{jia2021policy}, we have the following  martingale orthogonality condition for $\tilde{X}^{\bm{\pi}}$:
\[
\begin{aligned}
\E^{\p^W}\bigg[ & \int_0^T \xi_t \big[\dd J(t,\tilde{X}^{\bm{\pi}}_t;\bm{\pi}) - \beta J(t,\tilde{X}^{\bm{\pi}}_t;\bm{\pi}) \dd t \\
& + \int_{\mathcal{A}} [r(t,\tilde{X}^{\bm{\pi}}_t,a) + \gamma p\big(t,\tilde{X}^{\bm{\pi}}_t,a,\bm{\pi}(\cdot|t,\tilde{X}^{\bm{\pi}}_t) \big)] \bm{\pi}(a|t,\tilde{X}^{\bm{\pi}}_t)\dd a \dd t \big] \Big|\tilde{X}^{\bm{\pi}}_0 = x \bigg] = 0
\end{aligned}
\]
for all $\xi\in L^2_{\f^{\tilde{X}^{\bm{\pi}}}}\big( [0,T]; J(\cdot,\tilde{X}^{\bm{\pi}}_{\cdot};\bm{\pi})  \big)$. Now, any $\xi\in L^2_{\f^{\tilde{X}^{\bm{\pi}}}}\big( [0,T]; J(\cdot,\tilde{X}^{\bm{\pi}}_{\cdot};\bm{\pi})  \big)$ corresponds to a measurable functional
$\bm{\xi}:
[0,T]\times C([0,T];\mathbb{R}^d)\mapsto \mathbb{R}$ such that
$\xi_t = \bm{\xi}(t,\tilde{X}_{t\land\cdot}^{\bm{\pi}})$.
%
%
However,
$\tilde{X}^{\bm{\pi}}$ and $X^{\bm{\pi}}$ have the same distribution and $a_t^{\bm{\pi}} \sim \bm{\pi}(\cdot|t, \tilde{X}^{\bm{\pi}}_t)$; hence
\[\begin{aligned}
& \E^{\p^W}\bigg[  \int_0^T \bm{\xi}(t,\tilde{X}_{t\land\cdot}^{\bm{\pi}})  \big[\dd J(t,\tilde{X}^{\bm{\pi}}_t;\bm{\pi}) - \beta J(t,\tilde{X}^{\bm{\pi}}_t;\bm{\pi}) \dd t \big] \Big|\tilde{X}^{\bm{\pi}}_0 = x \bigg] \\
= & \E^{\p}\bigg[  \int_0^T \bm{\xi}(t,X_{t\land\cdot}^{\bm{\pi}}) \big[\dd J(t,X^{\bm{\pi}}_t;\bm{\pi}) - \beta J(t,X^{\bm{\pi}}_t;\bm{\pi}) \dd t \big]  \Big|X^{\bm{\pi}}_0 = x \bigg],
\end{aligned}  \]
and
\[\begin{aligned}
& \E^{\p^W}\bigg[  \int_0^T \bm{\xi}(t,\tilde{X}_{t\land\cdot}^{\bm{\pi}})  \int_{\mathcal{A}} [r(t,\tilde{X}^{\bm{\pi}}_t,a) + \gamma p\big(t,\tilde{X}^{\bm{\pi}}_t,a,\bm{\pi}(\cdot|t,\tilde{X}^{\bm{\pi}}_t) \big)] \bm{\pi}(a|t,\tilde{X}^{\bm{\pi}}_t)\dd a \dd t  \Big|\tilde{X}^{\bm{\pi}}_0 = x \bigg] \\
= & \E^{\p}\bigg[  \int_0^T \bm{\xi}(t,X_{t\land\cdot}^{\bm{\pi}}) [r(t,X^{\bm{\pi}}_t,a_t^{\bm{\pi}}) + \gamma p\big(t,X^{\bm{\pi}}_t,a_t^{\bm{\pi}},\bm{\pi}(\cdot|t,X^{\bm{\pi}}_t) \big)]  \dd t \Big|X^{\bm{\pi}}_0 = x \bigg].
\end{aligned}   \]

Combining the above two equations leads to  \eqref{eq:martingale orthogonal}.

\subsection*{Proof of Theorem \ref{thm:pg}}

It suffices to prove \eqref{eq:gradient representation expectation integrated a} equals \eqref{eq:policy gradient expectation}.

Fix $t$. Define a sequence of stopping times $\tau_n = \inf\{s\geq t:|X_s^{\bm{\pi}^{\phi}}| \geq n \}$.
Applying It\^o's lemma to $J(s, X_s^{\bm{\pi}^{\phi}})$, we obtain:
\[ \begin{aligned}
& \int_t^{T \wedge \tau_n}  e^{-\beta (s-t)} \bigg\{ \big[ \frac{\partial}{\partial \phi}\log \bm{\pi}^{\phi}(a_s^{\bm{\pi}^{\phi}}|s,X_s^{\bm{\pi}^{\phi}}) \big]  \times \big[  \dd J(s, X_s^{\bm{\pi}^{\phi}};\bm{\pi}^{\phi}) \\
& +  [  r(s,X_s^{\bm{\pi}^{\phi}},a_s^{\bm{\pi}^{\phi}}) + \gamma p\big(s,X_s^{\bm{\pi}^{\phi}},a_s^{\bm{\pi}^{\phi}},\bm{\pi}^{\phi}(\cdot|s,X_s^{\bm{\pi}^{\phi}})  \big) - \beta J(s,X_s^{\bm{\pi}^{\phi}};\bm{\pi}^{\phi})  ]     \dd s  \big]  \\
& + \gamma q (s,X_s^{\bm{\pi}^{\phi}},a_s^{\bm{\pi}^{\phi}},\phi) \dd s\bigg\}\\
= & \int_t^{T \wedge \tau_n}  e^{-\beta (s-t)} \bigg\{  \big[ \frac{\partial}{\partial \phi}\log \bm{\pi}^{\phi}(a_s^{\bm{\pi}^{\phi}}|s,X_s^{\bm{\pi}^{\phi}}) \big]  \times \big\{ \big[  \mathcal{L}^{a_s} J(s, X_s^{\bm{\pi}^{\phi}};\bm{\pi}^{\phi})  +    r(s,X_s^{\bm{\pi}^{\phi}},a_s^{\bm{\pi}^{\phi}}) \\
& + \gamma p\big(s,X_s^{\bm{\pi}^{\phi}},a_s^{\bm{\pi}^{\phi}},\bm{\pi}^{\phi}(\cdot|s,X_s^{\bm{\pi}^{\phi}})  \big) - \beta J(s,X_s^{\bm{\pi}^{\phi}};\bm{\pi}^{\phi})  \big]  \dd s \\
& + \frac{\partial J}{\partial x}(s,X_s^{\bm{\pi}^{\phi}};\bm{\pi}^{\phi})^\top  \sigma(s,X_s^{\bm{\pi}^{\phi}},a_s^{\bm{\pi}^{\phi}})  \dd W_s \big\}  + \gamma q (s,X_s^{\bm{\pi}^{\phi}},a_s^{\bm{\pi}^{\phi}},\phi) \dd s \bigg\}.
\end{aligned} \]
Taking expectation yields
\[\begin{aligned}
& \E^{\p}\bigg[\int_t^{T \wedge \tau_n} e^{-\beta (s-t)}  \bigg\{ \big[ \frac{\partial}{\partial \phi}\log \bm{\pi}^{\phi}(a_s^{\bm{\pi}^{\phi}}|s,X_s^{\bm{\pi}^{\phi}}) \big]  \times \big[  \dd J(s, X_s^{\bm{\pi}^{\phi}};\bm{\pi}^{\phi}) \\
& +  [  r(s,X_s^{\bm{\pi}^{\phi}},a_s^{\bm{\pi}^{\phi}}) + \gamma p\big(s,X_s^{\bm{\pi}^{\phi}},a_s^{\bm{\pi}^{\phi}},\bm{\pi}^{\phi}(\cdot|s,X_s^{\bm{\pi}^{\phi}})  \big) - \beta J(s,X_s^{\bm{\pi}^{\phi}};\bm{\pi}^{\phi})  ]     \dd s  \big]  \\
& + \gamma q (s,X_s^{\bm{\pi}^{\phi}},a_s^{\bm{\pi}^{\phi}},\phi) \dd s\bigg\} \Big|X_t^{\bm{\pi}^{\phi}} = x  \bigg] \\
= & \E^{\p}\bigg[\int_t^{T \wedge \tau_n}  e^{-\beta (s-t)}  \bigg\{  \big[ \frac{\partial}{\partial \phi}\log \bm{\pi}^{\phi}(a_s^{\bm{\pi}^{\phi}}|s,X_s^{\bm{\pi}^{\phi}}) \big]  \times \big\{ \big[  \mathcal{L}^{a_s^{\bm{\pi}^{\phi}}} J(s, X_s^{\bm{\pi}^{\phi}};\bm{\pi}^{\phi})  +    r(s,X_s^{\bm{\pi}^{\phi}},a_s^{\bm{\pi}^{\phi}}) \\
& + \gamma p\big(s,X_s^{\bm{\pi}^{\phi}},a_s^{\bm{\pi}^{\phi}},\bm{\pi}^{\phi}(\cdot|s,X_s^{\bm{\pi}^{\phi}})  \big) - \beta J(s,X_s^{\bm{\pi}^{\phi}};\bm{\pi}^{\phi})  \big]  \dd s \big\} + \gamma q (s,X_s^{\bm{\pi}^{\phi}},a_s^{\bm{\pi}^{\phi}},\phi) \dd s \bigg\} \Big|X_t^{\bm{\pi}^{\phi}} = x  \bigg] \\
& +  \E^{\p}\bigg[\int_t^{T \wedge \tau_n}  e^{-\beta (s-t)}   \frac{\partial}{\partial \phi}\log \bm{\pi}^{\phi}(a_s^{\bm{\pi}^{\phi}}|s,X_s^{\bm{\pi}^{\phi}}) \frac{\partial J}{\partial x}(s,X_s^{\bm{\pi}^{\phi}};\bm{\pi}^{\phi})^\top  \sigma(s,X_s^{\bm{\pi}^{\phi}},a_s)  \dd W_s  \Big|X_t^{\bm{\pi}^{\phi}} = x  \bigg] \\
= & \E^{\p}\bigg[\int_t^{T \wedge \tau_n}  e^{-\beta (s-t)}  \check{r}(s,X_s^{\bm{\pi}^{\phi}},a_s^{\bm{\pi}^{\phi}};\phi) \dd s \Big|X_t^{\bm{\pi}^{\phi}} = x  \bigg] .
\end{aligned}\]
The second term above vanishes because when $t\leq s \leq T\wedge \tau_n$, it follows from  Assumptions \ref{ass:dynamic} that
\[ \begin{aligned}
& \Big| \int_{\mathcal{A}}  \big[\frac{\partial}{\partial \phi}\log \bm{\pi}^{\phi}(a|s,\tilde{X}_s^{\bm{\pi}^{\phi}}) \big] \times \frac{\partial J}{\partial x}(s,\tilde{X}_s^{\bm{\pi}^{\phi}};\bm{\pi}^{\phi})^\top  \sigma(s,\tilde{X}_s^{\bm{\pi}^{\phi}},a) \bm{\pi}^{\phi}(a|s,\tilde{X}_s^{\bm{\pi}^{\phi}}) \dd a   \Big|^2  \\
\leq & \int_{\mathcal{A}}  \Big| \frac{\partial}{\partial \phi}\log \bm{\pi}^{\phi}(a|s,\tilde{X}_s^{\bm{\pi}^{\phi}})  \frac{\partial J}{\partial x}(s,\tilde{X}_s^{\bm{\pi}^{\phi}};\bm{\pi}^{\phi})^\top  \sigma(s,\tilde{X}_s^{\bm{\pi}^{\phi}},a) \Big|^2 \bm{\pi}^{\phi}(a|s,\tilde{X}_s^{\bm{\pi}^{\phi}}) \dd a \\
\leq & C\max_{|x|\leq n,0\leq t\leq T}| \frac{\partial }{\partial x}J(t,x;\bm{\pi}^{\phi})| \int_{\mathcal{A}}  \Big| \frac{\partial}{\partial \phi}\log \bm{\pi}^{\phi}(a|s,\tilde{X}_s^{\bm{\pi}^{\phi}})\Big|^2 (1 + |\tilde{X}_s^{\bm{\pi}^{\phi}}|)^2 \bm{\pi}^{\phi}(a|s,\tilde{X}_s^{\bm{\pi}^{\phi}}) \dd a \\
\leq & C(1+n)^2\max_{|x|\leq n,0\leq t\leq T}| \frac{\partial }{\partial x}J(t,x;\bm{\pi}^{\phi})| \int_{\mathcal{A}}  \Big| \frac{\partial}{\partial \phi}\log \bm{\pi}^{\phi}(a|s,\tilde{X}_s^{\bm{\pi}^{\phi}})\Big|^2  \bm{\pi}^{\phi}(a|s,\tilde{X}_s^{\bm{\pi}^{\phi}}) \dd a,
\end{aligned}\]
which is bounded by a function of $n$ due to Assumption \ref{ass:log likelihood pg}.
Thus,
\[ \begin{aligned}
&	\E^{\p}\bigg[\int_t^{T \wedge \tau_n}  e^{-\beta (s-t)}  \big[ \frac{\partial}{\partial \phi}\log \bm{\pi}^{\phi}(a_s^{\bm{\pi}^{\phi}}|s,X_s^{\bm{\pi}^{\phi}}) \big] \frac{\partial J}{\partial x}(s,X_s^{\bm{\pi}^{\phi}};\bm{\pi}^{\phi})^\top  \sigma(s,X_s^{\bm{\pi}^{\phi}},a_s^{\bm{\pi}^{\phi}})  \dd W_s  \Big|X_t^{\bm{\pi}^{\phi}} = x  \bigg] \\
= & \E^{\p^W}\bigg[\int_t^{T \wedge \tau_n}  e^{-\beta (s-t)}  \int_{\mathcal{A}} \big[ \frac{\partial}{\partial \phi}\log \bm{\pi}^{\phi}(a|s,\tilde{X}_s^{\bm{\pi}^{\phi}}) \big] \\
& \times \frac{\partial J}{\partial x}(s,\tilde{X}_s^{\bm{\pi}^{\phi}};\bm{\pi}^{\phi})^\top  \sigma(s,\tilde{X}_s^{\bm{\pi}^{\phi}},a) \bm{\pi}^{\phi}(a|s,\tilde{X}_s^{\bm{\pi}^{\phi}}) \dd a \dd W_s  \Big|\tilde{X}_t^{\bm{\pi}^{\phi}} = x  \bigg]=0.
\end{aligned} \]

Lastly, note
\[\begin{aligned}
& \E^{\p}\bigg[\int_t^{T \wedge \tau_n}  e^{-\beta (s-t)}  \check{r}(s,X_s^{\bm{\pi}^{\phi}},a_s^{\bm{\pi}^{\phi}};\phi) \dd s \Big|X_t^{\bm{\pi}^{\phi}} = x  \bigg] \\
& = \E^{\p^W}\bigg[\int_t^{T \wedge \tau_n}  e^{-\beta (s-t)} \int_{\mathcal{A}} \check{r}(s,\tilde{X}_s^{\bm{\pi}^{\phi}},a;\phi)\bm{\pi}^{\phi}(a|s, \tilde{X}_s^{\bm{\pi}^{\phi}})\dd a \dd s \Big|\tilde{X}_t^{\bm{\pi}^{\phi}} = x  \bigg].
\end{aligned} \]
By Assumption \ref{ass:log likelihood pg} and Lemma \ref{lemma:relaxed sde solution}, we get
\[\begin{aligned}
& \E^{\p^W}\bigg[\int_{\mathcal{A}} |\check{r}(s,\tilde{X}_s^{\bm{\pi}^{\phi}},a;\phi)|\bm{\pi}^{\phi}(a|s, \tilde{X}_s^{\bm{\pi}^{\phi}})\dd a \Big|\tilde{X}_t^{\bm{\pi}^{\phi}} = x  \bigg] \\
\leq & C_1\E^{\p^W}\bigg[ 1+|\tilde{X}_t^{\bm{\pi}^{\phi}}|^{\mu} \Big|\tilde{X}_t^{\bm{\pi}^{\phi}} = x  \bigg] \leq C_2 (1+|x|^{\mu}) .
\end{aligned}  \]
Hence by the dominance convergence theorem, we conclude that as $n\to \infty$,
\[\begin{aligned}
& \E^{\p^W}\bigg[\int_t^{T \wedge \tau_n}  e^{-\beta (s-t)} \int_{\mathcal{A}} \check{r}(s,\tilde{X}_s^{\bm{\pi}^{\phi}},a;\phi)\bm{\pi}^{\phi}(a|s, \tilde{X}_s^{\bm{\pi}^{\phi}})\dd a \dd s \Big|\tilde{X}_t^{\bm{\pi}^{\phi}} = x  \bigg] \\
\to & \E^{\p^W}\bigg[\int_t^{T }  e^{-\beta (s-t)} \int_{\mathcal{A}} \check{r}(s,\tilde{X}_s^{\bm{\pi}^{\phi}},a;\phi)\bm{\pi}^{\phi}(a|s, \tilde{X}_s^{\bm{\pi}^{\phi}})\dd a \dd s \Big|\tilde{X}_t^{\bm{\pi}^{\phi}} = x  \bigg] .
\end{aligned}  \]
This proves the desired result.

\subsection*{Proof of Theorem \ref{prop:pg optimal policy condition}}

The proof is similar to the proof of Theorem \ref{prop:pe martingale} and that in \citet[Proposition 4]{jia2021policy} by noticing $\bm{\pi}^*$ satisfies \eqref{eq:optimal value function foc}.

First, it suffices to consider the case when $\zeta = 0$ because of Theorem \ref{prop:pe martingale}. For $\eta\in L^2_{\f^{X^{\bm{\pi}^{\phi^*}}}}\big( [0,T] , J(\cdot,X_{\cdot}^{\bm{\pi}^{\phi^*}};\bm{\pi}^{\phi^*}) \big)$, we write $\eta_s = \bm{\eta}(s, X^{\bm{\pi}^{\phi^*}}_{s\wedge \cdot})$. Then the right hand side of \eqref{eq:optimal policy condition} can be written as
\[
\begin{aligned}
& \E^{\p}\Bigg[\int_0^T \bm{\eta}(s, X^{\bm{\pi}^{\phi^*}}_{s\wedge \cdot}) \bigg\{  \big[ \frac{\partial}{\partial \phi}\log \bm{\pi}^{\phi^*}(a_s^{\bm{\pi}^{\phi^*}}|s,X_s^{\bm{\pi}^{\phi^*}}) \big]  \big[  \dd J(s, X_s^{\bm{\pi}};\bm{\pi}^{\phi^*})  \\
& +  [ r(s,X_s^{\bm{\pi}^{\phi^*}},a_s^{\bm{\pi}^{\phi^*}}) + \gamma p\big( s,X_s^{\bm{\pi}^{\phi^*}},a_s^{\bm{\pi}^{\phi^*}},\bm{\pi}^{\phi^*}(\cdot|s, X_s^{\bm{\pi}^{\phi^*}}) \big) - \beta J(s,X_s^{\bm{\pi}^{\phi^*}};\bm{\pi}^{\phi^*})  ]      \dd s  \big] \\
&  +\gamma q(s, X_s^{\bm{\pi}^{\phi^*}},a_s^{\bm{\pi}^{\phi^*}},\phi^*) \dd s\bigg\} \Big| X_0^{\bm{\pi}^{\phi^*}} = x  \Bigg]\\
= & \int_0^T  \E^{\p}\Bigg[  \bm{\eta}(s, X^{\bm{\pi}^{\phi^*}}_{s\wedge \cdot}) \bigg\{  \big[ \frac{\partial}{\partial \phi}\log \bm{\pi}^{\phi^*}(a_s^{\bm{\pi}^{\phi^*}}|s,X_s^{\bm{\pi}^{\phi^*}}) \big]  \big[  \mathcal{L}^{a_s^{\bm{\pi}^{\phi^*}}} J(s, X_s^{\bm{\pi}};\bm{\pi}^{\phi^*})  \\
& +  r(s,X_s^{\bm{\pi}^{\phi^*}},a_s^{\bm{\pi}^{\phi^*}}) + \gamma p\big( s,X_s^{\bm{\pi}^{\phi^*}},a_s^{\bm{\pi}^{\phi^*}},\bm{\pi}^{\phi^*}(\cdot|s, X_s^{\bm{\pi}^{\phi^*}}) \big) - \beta J(s,X_s^{\bm{\pi}^{\phi^*}};\bm{\pi}^{\phi^*})    \big] \\
&  +\gamma q(s, X_s^{\bm{\pi}^{\phi^*}},a_s^{\bm{\pi}^{\phi^*}},\phi^*) \bigg\}  \Big| X_0^{\bm{\pi}^{\phi^*}} = x  \Bigg] \dd s \\
= & \int_0^T  \E^{\p^W}\Bigg[  \bm{\eta}(s, \tilde{X}^{\bm{\pi}^{\phi^*}}_{s\wedge \cdot}) \int_{{\cal A}} \bigg\{  \big[ \frac{\partial}{\partial \phi}\log \bm{\pi}^{\phi^*}(a|s,\tilde{X}_s^{\bm{\pi}^{\phi^*}}) \big]  \big[  \mathcal{L}^{a} J(s, \tilde{X}_s^{\bm{\pi}};\bm{\pi}^{\phi^*})  \\
& +  r(s,\tilde{X}_s^{\bm{\pi}^{\phi^*}},a) + \gamma p\big( s,\tilde{X}_s^{\bm{\pi}^{\phi^*}},a,\bm{\pi}^{\phi^*}(\cdot|s, \tilde{X}_s^{\bm{\pi}^{\phi^*}}) \big) - \beta J(s,\tilde{X}_s^{\bm{\pi}^{\phi^*}};\bm{\pi}^{\phi^*})    \big] \\
&  +\gamma q(s, \tilde{X}_s^{\bm{\pi}^{\phi^*}},a,\phi^*) \bigg\} \bm{\pi}^{\phi^*}(a|s,\tilde{X}_s^{\bm{\pi}^{\phi^*}}) \dd a  \Big|\tilde{X}_0^{\bm{\pi}^{\phi^*}} = x  \Bigg] \dd s  = 0,\\
\end{aligned}
\]
where 
the last equality follows from \eqref{eq:optimal value function foc}.

\subsection*{Proof of Lemma \ref{lemma:ergodic feynmann-kac}}
Apply It\^o's lemma to $J(\tilde{X}_s^{\bm{\pi}};\bm{\pi})$ on $s\in [t,T]$ to obtain
\[\begin{aligned}
&\E^{\p^W}\bigg[ J(\tilde{X}_T^{\bm{\pi}};\bm{\pi}) \Big| \tilde{X}_t^{\bm{\pi}} = x \bigg] - J(x;\bm{\pi}) \\
= & \E^{\p^W}\bigg[\int_t^T \int_{\mathcal{A}} \mathcal{L}^a J(\tilde{X}_s^{\bm{\pi}};\bm{\pi}) \bm{\pi}(a|\tilde{X}_s^{\bm{\pi}}) \dd a \dd s \Big| \tilde{X}_t^{\bm{\pi}} = x\bigg] \\
= & \E^{\p^W}\bigg[ \int_t^T\int_{\mathcal{A}} -r(\tilde{X}_s^{\bm{\pi}}, a) - \gamma p\big(\tilde{X}_s^{\bm{\pi}}, a, \bm{\pi}(\cdot|\tilde{X}_s^{\bm{\pi}}) \big) \bm{\pi}(a|\tilde{X}_s^{\bm{\pi}}) \dd a \dd s\Big| \tilde{X}_t^{\bm{\pi}} = x \bigg] + V(\bm{\pi})(T-t) .
\end{aligned}  \]
Therefore,
\[ \begin{aligned}
& \frac{1}{T}\E^{\p}\bigg[\int_t^T \big[ r(X_s^{\bm{\pi}}, a_s^{\bm{\pi}}) + \gamma p\big(X_s^{\bm{\pi}}, a_s^{\bm{\pi}}, \bm{\pi}(\cdot|X_s^{\bm{\pi}}) \big) \big]\dd s \Big| X_t^{\bm{\pi}} = x\bigg] \\
= &\frac{1}{T}\E^{\p^W}\bigg[\int_t^T\int_{\mathcal{A}} \big[ r(\tilde{X}_s^{\bm{\pi}}, a) + \gamma p\big(\tilde{X}_s^{\bm{\pi}}, a, \bm{\pi}(\cdot|\tilde{X}_s^{\bm{\pi}}) \big)\big] \bm{\pi}(a|\tilde{X}_s^{\bm{\pi}}) \dd a \dd s\Big| \tilde{X}_t^{\bm{\pi}} = x \bigg]\\
= & V(\bm{\pi})\frac{T-t}{T} + \frac{1}{T}J(x;\bm{\pi}) - \frac{1}{T}\E^{\p^W}\bigg[ J(\tilde{X}_T^{\bm{\pi}};\bm{\pi}) \Big| \tilde{X}_t^{\bm{\pi}} = x \bigg].
\end{aligned}  \]

By a similar localization argument as in the proof of Theorem \ref{thm:pg}, we can show that $\limsup_{T\to \infty}\E^{\p^W}\bigg[ J(\tilde{X}_T^{\bm{\pi}};\bm{\pi}) \Big| \tilde{X}_t^{\bm{\pi}} = x \bigg] $ is finite and independent of  $x$. Taking limit $T\to \infty$ on both sides of the above yields \eqref{con1}.

The above analysis also implies
\[ J(\tilde{X}_t^{\bm{\pi}};\bm{\pi}) + \int_0^t \int_{\mathcal{A}} [ r(\tilde{X}_s^{\bm{\pi}}, a) + \gamma p\big(\tilde{X}_s^{\bm{\pi}}, a, \bm{\pi}(\cdot|\tilde{X}_s^{\bm{\pi}}) \big) - V(\bm{\pi}) ] \bm{\pi}(a|\tilde{X}_s^{\bm{\pi}}) \dd a \dd s  \]
is an $(\f^{\tilde{X^{\bm{\pi}}}},\p^W)$-martingale. For the same reason as in the proof of Theorem \ref{prop:pe martingale}, we arrive at the second desired conclusion of the lemma.

\vskip 0.2in
\bibliography{reference}

\end{document}